%% file: acl_latex.tex
\documentclass[11pt]{article}

\usepackage[final]{acl}

\usepackage{times}
\usepackage{latexsym}
\usepackage{amsmath}
\usepackage{amsfonts}
\usepackage[T1]{fontenc}

\usepackage[utf8]{inputenc}

\usepackage{microtype}

\usepackage{inconsolata}

\usepackage{graphicx}

\usepackage{algorithm}
\usepackage{algpseudocode}
\algrenewcommand\algorithmiccomment[1]{\hfill$\triangleright$ #1}

\usepackage{etoolbox}
\usepackage{titlesec}

\pretocmd{\appendix}{%
  \renewcommand{\thesection}{\Alph{section}}

  \titleformat{\section}
    {\normalfont\large\bfseries}
    {Appendix \thesection}
    {1em}
    {}
}{}{}

\usepackage{booktabs}
\usepackage{makecell}
\usepackage{pifont} 
\newcommand{\cmark}{\ding{51}} 

\usepackage{tabularx}
\usepackage[nolist,nohyperlinks]{acronym}
\usepackage{cleveref}
\usepackage{comment}
\usepackage[most]{tcolorbox}
\definecolor{customblue}{rgb}{0.27, 0.36, 0.55}
\usepackage{subcaption}
\usepackage{multirow}
\usepackage[dvipsnames,table]{xcolor}
\usepackage{xstring}
\usepackage{pgfmath}
\usepackage{colortbl}
\usepackage{stfloats}

\definecolor{highlightcolor}{RGB}{255,77,77}

\newcommand{\basevqacell}[1]{\cellcolor{gray!10}\textcolor{darkgray}{$#1$}}

\newcommand{\basevqa}[1]{{\color{darkgray}#1}}

\newcommand{\colornum}[1]{%
    \pgfmathtruncatemacro{\val}{#1}%
    \pgfmathsetmacro{\t}{(\val-1)/47}%
    \ifnum\val<25
        \pgfmathsetmacro{\r}{0 + 0.5*(\val-1)/23}%
        \pgfmathsetmacro{\g}{0.5}%
        \pgfmathsetmacro{\b}{0 + 0.5*(\val-1)/23}%
    \else
        \pgfmathsetmacro{\r}{0.5 + 0.1*(\val-24)/24}%
        \pgfmathsetmacro{\g}{0.5 - 0.5*(\val-24)/24}%
        \pgfmathsetmacro{\b}{0.5 - 0.5*(\val-24)/24}%
    \fi
    \definecolor{tmpcolor}{rgb}{\r,\g,\b}%
    \textcolor{tmpcolor}{$\mathbf{#1}$}%
}

\newcommand{\signed}[1]{%
  \IfBeginWith{#1}{\pm}{%
    \textcolor{gray}{#1}%
  }{%
    \IfBeginWith{#1}{-}{%
      \textcolor{BrickRed}{#1}%
    }{%
      \textcolor{OliveGreen}{#1}%
    }%
  }%
}

\definecolor{cellgray}{gray}{0.93}

\newcommand{\GrayBoldNum}[1]{%
    \cellcolor{cellgray}
    \bfseries #1
}

\newcommand{\GrayNum}[1]{%
    \cellcolor{cellgray}
    #1
}

\begin{acronym}
    \acro{MLLM}[MLLM]{Multimodal Large Language Model}
    \acro{LLM}[LLM]{Large Language Model}
    \acro{VQA}[VQA]{Visual Question Answering}
    \acro{MC-VQA}[MC-VQA]{multiple-choice VQA}
    \acro{PIA}[PIA]{position-invariant accuracy}
    \acro{PriDe}[PriDe]{Debiasing with Prior estimation}
    \acro{FiTo}[FiTo]{First Token estimation}
    \acro{LMM}[LMM]{linear mixed model}
    \acro{CP-LN}[CP-LN]{Cloze prompt (length normalized)}
    \acro{pp}[pp]{percentage points}
\end{acronym}

\usepackage{siunitx}

\definecolor{mygray}{gray}{0.9}

\global

\tcbset{
  vqalayoutbox/.style={
    width=\linewidth,
    colback=white,
    colframe=black,
    colbacktitle=black,
    coltitle=white,
    fonttitle=\bfseries,
    boxrule=0.6pt,
    arc=1.2mm,
    left=6pt,right=6pt,top=6pt,bottom=6pt,
    title style={left=4pt,right=4pt,top=2pt,bottom=2pt},
    before skip=6pt,
    after skip=6pt,
  }
}
\newtcolorbox{VQABox}[2][]{%
  vqalayoutbox,
  title={#2},
  #1
}

\newcommand{\parag}[1]{\noindent\textbf{#1.}\hspace{1.8mm}}

\title{Unexplored flaws in multiple-choice VQA make benchmarking unreliable}

\author{Fabio Rosenthal\,$^{1,2}$ \quad Sebastian Schmidt\,$^{1}$ \quad Thorsten Bagdonat\,$^{2}$ \\ \textbf{Thorsten Graf\,}$^{2}$ \quad \textbf{Stephan Günnemann\,}$^{1}$ \quad \textbf{Leo Schwinn\,}$^{1,3,4}$ \\ $^{1}$\;Technical University of Munich, $^{2}$\;Volkswagen AG, $^{3}$\;Helmholtz AI, $^{4}$\;ELLIS \\ \texttt{Correspondence: f.rosenthal@tum.de}}

\crefname{figure}{Fig.}{Figs.}
\crefname{table}{Tab.}{Tabs.}
\crefname{section}{Sec.}{Secs.}
\crefname{subsection}{Sec.}{Secs.}
\crefname{subsubsection}{Sec.}{Secs.}
\crefname{equation}{Eq.}{Eqs.}
\crefname{appendix}{App.}{Apps.}

\begin{document}
\maketitle

\begin{abstract}
    Previous works identify sensitivity to option order as a key issue in \ac{MC-VQA} evaluation and propose protocols to mitigate this effect.
    We show that such mitigation is insufficient to ensure the validity of \ac{MC-VQA} as a reliable benchmark for \acp{MLLM}: performance remains highly sensitive to \emph{semantically neutral} prompt format choices that are not controlled by current benchmarks.
    In a large-scale study spanning \textbf{seven} \acp{MLLM} and \textbf{five} \ac{MC-VQA} datasets, we find frequent rank reversals even under order-invariant evaluation.
    These reversals arise when we systematically vary option ID sets, delimiters, and separators, yielding $48$ semantically equivalent prompt formats.
    Mechanistic analyses trace this instability to low-level language modeling effects: tokenizer-induced fusion or removal of option ID tokens introduces corrupted option ID tokens into the input sequence, while the choice of option ID sets directly affects the reliability of attention patterns for option selection.
    Accordingly, \ac{MC-VQA} rankings correlate weakly with open-ended evaluation, indicating that \ac{MC-VQA} reflects option-selection dynamics in addition to multimodal reasoning.
    These findings identify prompt formatting as a major, previously under-controlled confounder in \ac{MC-VQA} benchmarking and motivate evaluation protocols that explicitly control prompt format sensitivity.
\end{abstract}

\section{Introduction}
\label{sec:intro}

\acp{MLLM} demonstrate strong perceptual and reasoning capabilities on image-text interleaved inputs \cite{bai_qwen_25_vl_2025}, making them applicable across domains such as high-resolution imaging \cite{zhong_2025_focus}, medical imaging \cite{matos_medicalvqa_2025}, and autonomous driving \cite{marcu_lingovqa_2025}.
Benchmarking aims to evaluate these capabilities through standardized proxy tasks \cite{lamparth_2024_aibenchmark, eriksson_2025_aievaluation}.
Ideally, benchmarks enable reliable predictions about model performance on future real‑world tasks \cite{singh_leaderboardillusion_2025}.
However, benchmarks inherently provide only an imperfect proxy for true \ac{MLLM} capabilities, as the open-ended tasks encountered in deployment are not directly measurable \cite{longjohn_2024_betterbenchmark}.
Therefore, benchmarks are only meaningful insofar as their outcomes reflect underlying model capabilities, rather than being driven by superficial variations in the evaluation setup.

\noindent Common \ac{MLLM} benchmarks are \acf{MC-VQA}, in which \acp{MLLM} must select an answer from a set of options \cite{schwenk_aokvqa_2022, liu_mmbench_2024, yue_mmmu_2024}, and thereby simplifies answer evaluation compared to open-ended generation \cite{robinson_leveraging_2023, molfese_rightanswer_2025}.
Previous works show that the option order influences multiple-choice performance of (M)LLMs, even when the underlying task remains unchanged \cite{zheng_llmsnotrobust_2024, pezeshkpour_llmsensitivity_2024}.
This may undermine the reliability of benchmarking results on \ac{MC-VQA} and limit its validity as a proxy for evaluating \acp{MLLM}.
However, these works propose methods to mitigate the effects of option order.

\begin{figure}[t]
    \centering
    \includegraphics[clip, trim=0cm 27.9cm 8.2cm 0cm, width=\linewidth]{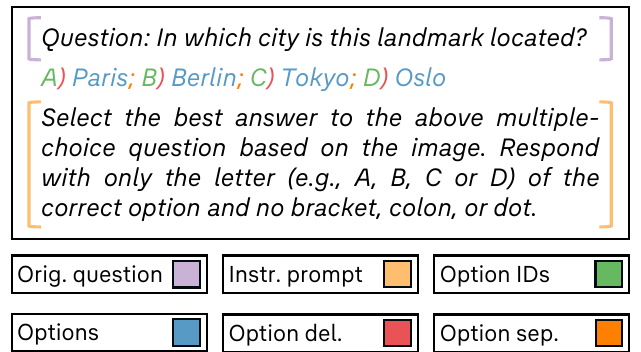}
    \vspace{-13pt}
    \caption{\textbf{Prompt template in \ac{MC-VQA}.} A prompt in \ac{MC-VQA} consists of the \textit{question}, the \textit{instruction}, and \textit{options}, with their respective indexing \textit{option IDs}.}
    \label{fig:prompt_template}
    \vspace{-15pt}
\end{figure}

\noindent Yet, these efforts focus exclusively on option ordering, overlooking other fundamental aspects of prompt design.
Beyond ordering, multiple format factors influence the options representation in the prompt \cite{wei_unveiling_2024, sanz-guerrero-etal-2025-mindgap}.
Typically, \textit{option IDs} index each option, and the prompt instructs the \ac{MLLM} to output the corresponding ID rather than the option \cite{robinson_leveraging_2023, liu_mmbench_2024}.
The \textit{option delimiter} hyphenates each option ID and the respective option, and the \textit{option separator} splits different option ID-option pairs in the prompt (see \cref{fig:prompt_template}).
Note that there is no standardized format for these elements, and prompt formats vary across datasets.
Therefore, the influence of prompt format decisions on evaluation outcomes has not been systematically studied.

\begin{figure}[t]
        \centering
        \includegraphics[clip, trim=1.2cm 1.1cm 1.1cm 1.1cm,width=\linewidth]{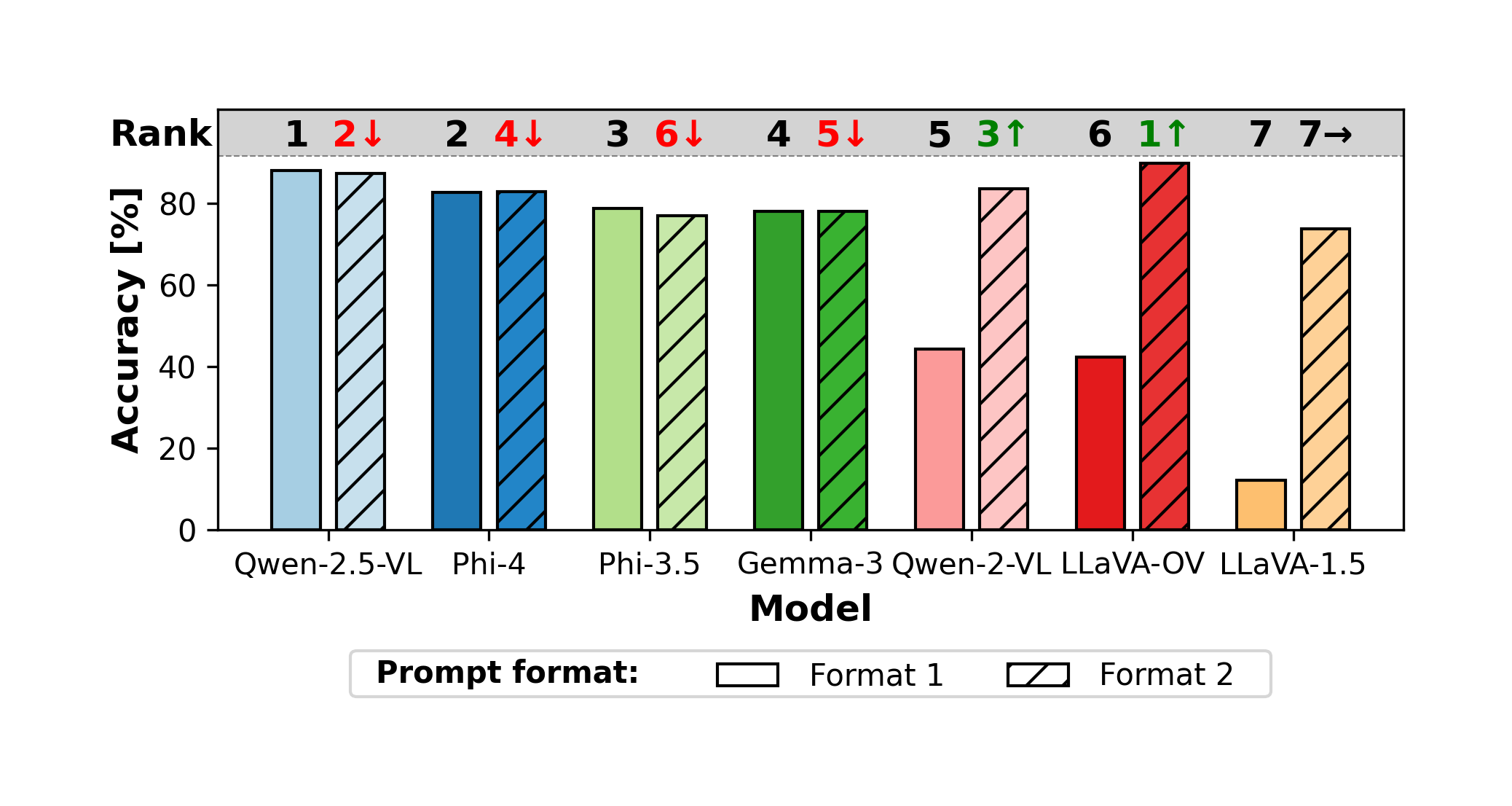}
        \vspace{-18pt}
        \caption{\textbf{Benchmark unreliability in \ac{MC-VQA}.} 
        We compare \ac{MLLM} benchmarking on A-OKVQA using two prompt formats in the worst-case scenario. 
        \textit{Format 1} uses lowercase option IDs, double brackets as delimiters, and line breaks as separators, whereas \textit{Format 2} uses uppercase option IDs, dots as delimiters, and commas as separators.}
        \label{fig:benchmark_difference}
        \vspace{-7pt}
\end{figure}

\noindent In this paper, we show that semantically neutral prompt format choices substantially affect \ac{MC-VQA} evaluation, calling into question its reliability as a benchmark.
For instance, LLaVA-OV \cite{li_llava_ov_2025} shifts from the second-lowest to the top rank under a changed prompt format (see \cref{fig:benchmark_difference}), with an accuracy difference of nearly $50$ \ac{pp}.
Similarly, LLaVA-1.5 \cite{liu_llava_15_2024} and Qwen-2-VL \cite{yang_qwen_2_vl_2024} exhibit considerable accuracy differences across formats.\footnote{These bias effects were isolated from the known position bias through a circular evaluation scheme.}
Across seven MLLMs, five datasets, and 48 prompt formats, we observe large accuracy variations and frequent rank reversals, even under order-invariant evaluation.
These effects persist for larger and proprietary models.

\noindent We trace this instability to low-level language modeling effects rather than changes in visual or semantic reasoning.
These effects introduce variability in how option ID tokens are represented and selected, leading to inconsistent model predictions across prompt formats.
We further show that \acp{MLLM} are systematically aware of certain prompt formats that yield better performance.
Finally, the weak alignment with open-ended evaluation suggests that \ac{MC-VQA} reflects both option-selection dynamics and reasoning capabilities.

\noindent Our main contributions are: 
(1) We show in a large-scale study that small structural prompt format changes lead to benchmark rank variations,
(2) We identify mechanistic sources of format-dependent instability, including tokenization failures, generalization behavior of option ID sets beyond training exposure, and the selection of option ID set influences attention patterns in \acp{MLLM}, 
(3) We show that \acp{MLLM} exhibit implicit awareness of prompt format effectiveness, systematically favoring certain formats at rates far exceeding random selection, and
(4) We demonstrate that \ac{MC-VQA} substantially overestimates multimodal reasoning ability and correlates weakly with open-ended evaluation, calling into question its reliability as a benchmark and proxy for downstream performance.

\section{Preliminary}
\label{sec:background}

This section provides the necessary background on \acp{MLLM}, the \ac{MC-VQA} task, and the benchmarking scope used in this work.

\subsection{Multimodal Large Language Models}
\label{subsect:mllms}

\acfp{MLLM} consist of three components: a \ac{LLM}, a vision encoder, and a modality projector \cite{li_blip2_2023, liu_llava_15_2024}.
The vision encoder processes the input image to produce visual features represented as tokens $\mathbf{X}_{\text{v}}$ \cite{li_blip2_2023, liu_llava_2023}.
A modality projector maps $\mathbf{X}_{\text{v}}$ into the \ac{LLM} embedding space, yielding $\mathbf{X}_{\text{v, proj}}$ \cite{liu_llava_2023, liu_llava_15_2024}.
A model-specific tokenizer converts the text input into textual tokens $\mathbf{X}_\mathrm{t}$ \cite{li_blip2_2023, liu_llava_2023}.
Finally, the \ac{LLM} receives a concatenation of the visual and text tokens \cite{abdin_phi4_2024}:
\begin{equation}
    \mathbf{X}_{\mathrm{in}} = [\, \mathbf{X}_\mathrm{t} \,;\, \mathbf{X}_{\mathrm{v, proj}} \,]
\end{equation}

\noindent The \ac{LLM} generates its output autoregressively, predicting one token at a time conditioned on the prompt and previously generated tokens \cite{vaswani_attention_2017, liu_llava_2023}.
In \ac{MC-VQA}, this procedure typically reduces to generating a short option identifier (e.g., \texttt{'A'} or \texttt{'a'}) that already appears in the prompt \cite{antol_vqa_2015, vqav2_goyal_2017}.
This decoding setup makes multiple-choice evaluation particularly sensitive to how option IDs are tokenized and represented in the input.
Small prompt format changes that alter the tokenization of option IDs may therefore affect the generated tokens, even when the underlying reasoning remains unchanged.

\subsection{Visual Question Answering}
\label{subsect:vqa}

Visual Question Answering (VQA) evaluates the multimodal reasoning ability of \acp{MLLM}, which must answer questions about a given image \cite{schwenk_aokvqa_2022, yue_mmmu_2024}.
\ac{MC-VQA} simplifies correctness assertion by restricting the output space to a fixed set of option IDs \cite{robinson_leveraging_2023}.
Formally, each sample consists of an image, a question $q$, and a set of answer options $\mathcal{O} =\{o_j\}_{j=1}^{n}$ \cite{antol_vqa_2015, vqav2_goyal_2017}.
An option ID $i_j$ from a predefined ID set $\mathcal{I} =\{i_j\}_{j=1}^{n}$ indexes each option $o_j$ in the prompt.
We define a prompt format as $F = (\mathcal{I}, d, s)$, where $d$ denotes the delimiter separating an option ID from its option text, and $s$ denotes the separator between different option ID-option pairs.
Further, the permutation $\pi$ specifies the order in which options appear.
Given a fixed instruction prompt $\texttt{In}$ that requests the model to output an option ID, the resulting multiple-choice prompt is \cite{liu_mmbench_2024, zhang_mmerealworld_2025}:
\[ T(q,F,\pi) = q \;\Vert\; \operatorname{Join}_{s} \bigl((i_{\pi(k)}d\,o_{\pi(k)})_{k=1}^{n}\bigr) \;\Vert\; \texttt{In}. \]

\noindent The \ac{MLLM} generates an output token sequence that ideally matches one of the option IDs and maps to the corresponding GT answer~\cite{duan_vlmevalkit_2024, polo_promptllms_2024}.
This setup tightly couples model predictions to the tokenization and representation of option IDs, because the output  of the model is a discrete ID that appears in the prompt. 
In this work, we analyze how variations in the prompt format $F$ influence model behavior in \ac{MC-VQA}, while controlling for the effect of option order $\pi$.

\subsection{Benchmarking}
Benchmarking evaluates a set of models $\{\mathcal{M}_1, \dots, \mathcal{M}_K\}$ on datasets and ranks them according to task accuracy:
$\mathcal{M}_{(1)} \geq \mathcal{M}_{(2)} \geq \cdots \geq \mathcal{M}_{(K)}$, where $\mathcal{M}_{(1)}$ is the highest-scoring model~\cite{srivastava2023beyond,liang2023holistic}.
These rankings aim to reflect global model capability and therefore should not depend on arbitrary evaluation choices, such as the prompt format, that may alter rankings \cite{alzahrani_2024_benchmarks}.
For \acp{MLLM}, benchmarking frequently relies on \ac{MC-VQA} \cite{bai_qwen_25_vl_2025,abdin_phi4_2024}.
In this work, we analyze how these rankings change under variations in prompt format, holding the underlying task and evaluation protocol fixed.

\section{Related Work}
\label{sect:related_work}
We review work on prompt sensitivity in (M)LLMs, biases in multiple-choice evaluation, and the reliability of benchmark-based evaluation.

\parag{Prompt sensitivity in (M)LLMs}
(M)LLMs are sensitive to prompt formulation, where small changes in phrasing or structure can lead to different outputs~\cite{he_promptformat_2024, li_modelingpromptsvlms_2025}.
Prior work studies this phenomenon in a variety of settings, including few-shot learning, instruction following, and zero-shot evaluation~\cite{zhuo_prosa_2024, scalar_llmsens_2024, sun_2024_zeroshotrobust, mizrahi_2024_multipromptllm}.
These results show that model behavior can depend strongly on surface-level prompt variations, raising concerns about evaluation consistency.
However, most of this work focuses on free-form generation, while less attention has been paid to structured settings with semantically neutral option IDs, such as multiple-choice evaluation.

\parag{Biases in multiple-choice evaluation}
In multiple-choice settings, prior work identifies systematic biases related to option order, where models prefer certain positions even when the underlying task is unchanged~\cite{zheng_llmsnotrobust_2024, pezeshkpour_llmsensitivity_2024}.
Mitigation strategies include post-hoc calibration and set-invariant training objectives~\cite{egressy_setllm_2025, chen_pearl_2025}, while work in multimodal settings remains limited~\cite{tan_ordermatters_2024}.
Beyond order effects, recent evidence suggests that multiple-choice tasks can overestimate (M)LLM capabilities~\cite{raman2025testexploiter, liu2025revel}, indicating that high accuracy may not translate to strong performance in open-ended settings.
Further, it was shown that varying option ID sets has influence on \ac{LLM} benchmarking outcomes \cite{yang_2025_option}.
However, existing work primarily focuses on option permutations or model calibration, and does not systematically study how semantically neutral prompt format choices influence evaluation.

\parag{Benchmark reliability}
Recent work questions the reliability of benchmark-based evaluation, showing that model rankings can change under small variations in evaluation design, such as prompt phrasing, formatting, or response ordering~\cite{zhuo_prosa_2024, singh_leaderboardillusion_2025, he_promptformat_2024, wei_unveiling_2024, sanz-guerrero-etal-2025-mindgap}.
These findings suggest that evaluation results often depend on incidental design choices rather than intrinsic model capability~\cite{zheng_llmsnotrobust_2024}.
In this work, we focus on \ac{MC-VQA} and show that prompt format choices (option ID sets, delimiters, and separators) constitute a previously under-explored source of instability.
In contrast to prior work, we provide both a systematic analysis and mechanistic evidence linking evaluation instability to tokenization and decoding behavior.

\section{Benchmark Instability under Prompt Format Variations}
\label{sect:benchmark_instability}

\begin{figure*}[!t]
    \centering
    \includegraphics[width=\linewidth]{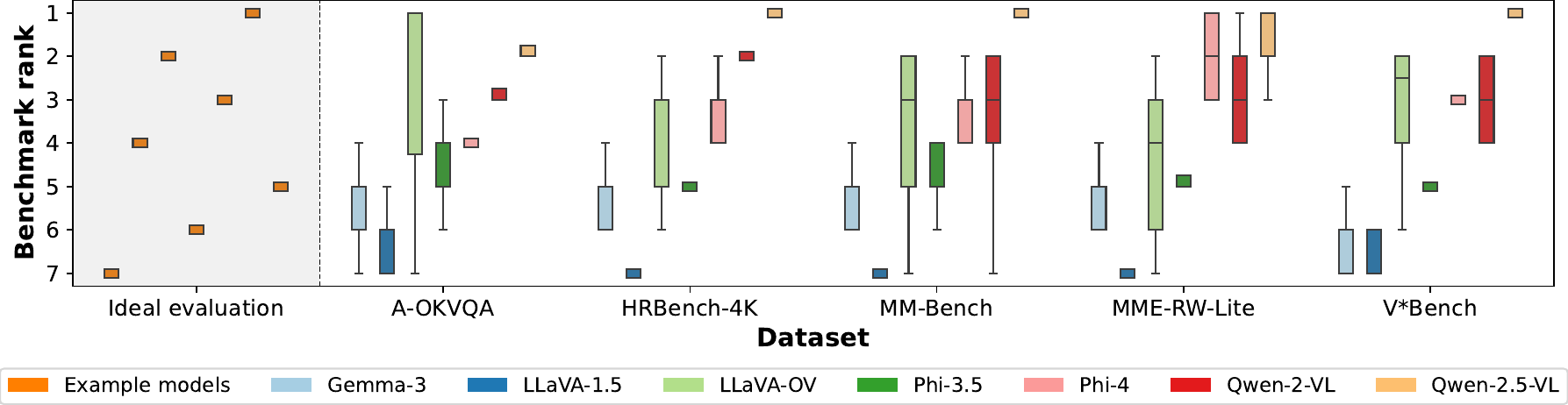}
    \vspace{-15pt}
    \caption{\textbf{Benchmarking results across five \ac{MC-VQA} datasets.} The figure shows the distribution of ranks for seven \acp{MLLM} across $48$ prompt format variants. The ranks vary substantially across datasets, indicating high sensitivity to prompt format. Compared to an ideal evaluation (left), where ranks remain constant, real-world evaluations exhibit significant rank variability: \textit{highlighting that \ac{MC-VQA} benchmarking is unreliable}.}
    \label{fig:benchmarking_results_boxplot}
\end{figure*}

Previous studies on biases in multiple-choice evaluation primarily focus on the effect of option order~\cite{zheng_llmsnotrobust_2024, pezeshkpour_llmsensitivity_2024, tan_ordermatters_2024}, while other sources of prompt variation remain largely unexplored.
To address this gap, we conduct a systematic study of how semantically neutral prompt format choices influence \ac{MC-VQA} evaluation.
We evaluate model performance across $48$ prompt formats, varying option IDs, delimiters, and separators, while controlling for option order through circular evaluation.
\cref{fig:benchmarking_results_boxplot} shows the resulting rank distributions of models across five \ac{MC-VQA} datasets.

\subsection{Experimental Framework}
\label{subsect:implementation_details}

Here, we outline the experimental framework, including the models, datasets, implementation details, and evaluation methodology used in our study.
We provide more extensive details in \cref{app:experimental_setup}.

\parag{\acp{MLLM}}
We use popular \acp{MLLM} covering different architectures: Gemma-3 \cite{kamath_gemma_3_2025}, LLaVA-1.5 \cite{liu_llava_15_2024}, LLaVA-OV \cite{li_llava_ov_2025}, Phi-3.5 \cite{abdin_phi35_2024}, Phi-4 \cite{abdin_phi4_2024}, Qwen-2-VL \cite{yang_qwen_2_vl_2024} and Qwen-2.5-VL \cite{bai_qwen_25_vl_2025}.

\parag{Datasets}
We use a highly diverse selection of datasets covering different domains and task difficulties, containing high-resolution datasets HRBench-4K \cite{wang_hrbench_2025} and V*Bench \cite{wu_vstar_2024}, general knowledge dataset A-OKVQA \cite{schwenk_aokvqa_2022}, and multi-domain benchmark datasets MME-RealWorld-Lite \cite{zhang_mmerealworld_2025} and MMBench \cite{liu_mmbench_2024}.
To reduce poor instruction-following of models, we use an optimized instruction prompt for all datasets.

\parag{Evaluation Setting}
As is common with \ac{MC-VQA} datasets, we evaluate all models in a \textit{zero-shot setting} \cite{li_blip_2022, li_blip2_2023, liu_llava_15_2024}.
Additionally, we enforce fully deterministic outputs across all models by disabling sampling, ensuring consistent comparisons on the same datasets for different prompt formats.
Further, we set the number of generated output tokens to match the maximum required to output all option IDs from the utilized option ID set (see \cref{tab:decoding_token_length_option_ids}).

\begin{table}[h]
    \centering
    \caption{\textbf{Prompt format variations.} We systematically vary the prompt format across three factors: (1) option ID set, (2) option delimiter, and (3) option separator.}
    \label{tab:prompt_format_variations}
    \vspace{-5pt}
    \resizebox{0.9\linewidth}{!}{
    \begin{tabular}{l l l}
        \toprule
        \textbf{Factor} & \textbf{Format variant} & \textbf{Example repr.} \\
        \midrule
        \multirow{4}{*}{\textbf{\makecell[l]{Option\\ID set}}} & Uppercase & \texttt{A{\color{gray}/}B{\color{gray}/}C{\color{gray}/}D} \\
        & Lowercase & \texttt{a{\color{gray}/}b{\color{gray}/}c{\color{gray}/}d} \\
        & Numbers & \texttt{1{\color{gray}/}2{\color{gray}/}3{\color{gray}/}4}  \\
        & Roman numbers & \texttt{I{\color{gray}/}II{\color{gray}/}III{\color{gray}/}IV}  \\
        \midrule
        \multirow{4}{*}{\textbf{\makecell[l]{Option\\delimiter}}} & Colon & \texttt{{\color{gray}\{opt\}}:} \\
        & Dot & \texttt{{\color{gray}\{opt\}}.}  \\
        & Bracket & \texttt{{\color{gray}\{opt\}})}  \\
        & Double brackets & \texttt{({\color{gray}\{opt\}})}  \\
        \midrule
        \multirow{3}{*}{\textbf{\makecell[l]{Option\\separator}}} & Line break & \texttt{{\color{gray}\{opt\}}\textbackslash n {\color{gray}\{opt\}}} \\
        & Comma & \texttt{{\color{gray}{\color{gray}\{opt\}}}, {\color{gray}{\color{gray}\{opt\}}}} \\
        & Semicolon & \texttt{{\color{gray}\{opt\}}; {\color{gray}\{opt\}}} \\
        \bottomrule
    \end{tabular}%
    }
\end{table}

\parag{Prompt Format Variations}
We systematically vary three components of the prompt format: (1) the option ID set, (2) the delimiter between ID and option text, and (3) the separator between options (see \cref{tab:prompt_format_variations}).
Combining four option ID sets, four delimiters, and three separators results in $48$ distinct prompt formats.\footnote{While computationally expensive, this exhaustive setup serves as a high-fidelity reference for analyzing evaluation instability rather than a practical evaluation protocol.}

\subsection{Rank Instability Across Prompt Formats}
\label{subsect:rank_instability}

We next analyze how model rankings vary across semantically equivalent prompt formats.
\cref{fig:benchmarking_results_boxplot} shows the rank distributions of all models across $48$ prompt formats per dataset.
Model rankings vary substantially across prompt formats and do not remain constant, in contrast to the \textit{ideal evaluation}\footnote{An ideal \ac{MC-VQA} evaluation yields rankings that are largely invariant to changes in prompt format.} shown in \cref{fig:benchmarking_results_boxplot}.
On several datasets, rank distributions overlap strongly and frequently invert, such that no single model maintains a consistent position.
This instability persists despite controlling for option-order effects through circular evaluation.

\noindent On A-OKVQA and MME-RealWorld-Lite, rank variability is high, with no model exhibiting a stable rank.
For example, LLaVA-OV ranges from top to near-bottom ranks on A-OKVQA depending on the prompt format.
On MMBench and V*Bench, the top-ranked model remains stable, but lower-ranked models exhibit substantial fluctuations.
In contrast, HRBench-4K shows the most stable rankings across prompt formats.

\noindent Additional analyses show that these effects are not explained by instruction-following failures or low-confidence examples, and persist under existing bias mitigation methods (see \cref{app:additional_results}).
Additionally, we find that circular evaluation effectively mitigates option order preference.
We further confirm that variations across all prompt format variations cause statistically significant accuracy changes (see \cref{app:significance_analysis}).
Overall, these results show that \ac{MC-VQA} rankings are sensitive to neutral prompt format variations, even under controlled evaluation settings.

\subsection{Model Scaling Does Not Eliminate Prompt Sensitivity}
The instability observed in \cref{subsect:rank_instability} raises the question whether larger models exhibit more robust behavior under prompt format variations.
To examine this, we extend our analysis on V*Bench to larger variants of Gemma-3 and Qwen2.5-VL, as well as proprietary models from the GPT-5.2 family.

\noindent Our analysis shows that increasing model size does not consistently reduce prompt sensitivity.
Qwen2.5-VL-32B exhibits reduced variation compared to its 7B counterpart, with an accuracy deviation of $2.6$ \ac{pp} compared to $7.1$ \ac{pp}.
However, the smaller Qwen2.5-VL-7B model outperforms the 32B variant under certain prompt formats, indicating that prompt design can outweigh gains from model scaling.
In contrast, Gemma-3-27B shows increased sensitivity, with $5.9$ \ac{pp} deviation compared to $3.9$ \ac{pp} for the smaller model.
Variants of GPT-5.2 similarly display substantial prompt-induced variability in \ac{MC-VQA}.
Overall, these results show that increasing model scale does not guarantee stable evaluation behavior under prompt format variations.

\begin{table}[hbtp]
    \centering
    \caption{\textbf{Sensitivity of larger \acp{MLLM} on V*Bench.} Results show that increased model size does not consistently reduce accuracy variance across prompt formats. Further, we report which model size within each family achieves the highest accuracy for each prompt format.}
    \vspace{-5pt}
    \label{tab:results_larger_models}
    \resizebox{\linewidth}{!}{
    \renewcommand{\arraystretch}{0.95}
    \begin{tabular}{l c c c c}
        \toprule
        \textbf{Model} & \textbf{\makecell{Avg. \\acc. $[\%]$}} & \textbf{\makecell{Min. \\acc. $[\%]$}} & \textbf{\makecell{Max. \\acc. $[\%]$}} & \textbf{\makecell{Higher acc. per\\prompt format}}\\
        \midrule
        Qwen2.5-VL-7B & $77.44$ & $72.65$ & $81.71$ & $7$ out of $48$\\ 
        Qwen2.5-VL-32B & $80.52$ & $78.85$ & $81.54$ & $41$ out of $48$\\
        \midrule
        \midrule
        Gemma-3-4B & $34.41$ & $32.31$ & $36.25$ & $0$ out of $48$\\ 
        Gemma-3-27B & $45.09$ & $41.78$ & $47.65$ & $48$ out of $48$\\
        \midrule
        \midrule
        GPT-5.2-nano & $59.70$ & $57.21$ & $61.41$ & $0$ out of $48$\\
        GPT-5.2-mini & $70.05$ & $67.61$ & $71.81$ & $48$ out of $48$\\
        \bottomrule
    \end{tabular}%
    }
\end{table}

\section{Mechanistic Sources of Instability}
\label{sect:mechanistic_sources}

\cref{sect:benchmark_instability} shows that semantically neutral prompt format choices induce rank reversals in \ac{MC-VQA}.
We now ask why these changes affect model predictions so strongly.
Our hypothesis is that prompt format sensitivity arises from low-level language-model behavior rather than changes in reasoning.
In particular, we trace the observed benchmark instability to three interacting factors: tokenization of option IDs, the ability of models to generalize beyond familiar option ID sets, and attention patterns to option IDs during answer generation.

\subsection{Tokenization Effects on Option IDs}
\label{subsect:tokenization_effects}

One potential source of instability is how tokenizers represent option IDs in the input sequence.
Across prompt formats, we observe two recurring patterns: (1) the option ID-delimiter pair tokenizes into more than two tokens, and (2) the tokenizer fails to preserve the option ID as a standalone token.
Both effects alter the input representation in ways that are unrelated to the task semantics and may reduce evaluation stability.
Therefore, we analyze whether atypical tokenization under varying prompt formats causes the observed benchmark instability.
We provide more details in \cref{app:mechanistic_tokenization}.

\parag{Token count of option ID-delimiter}
First, we analyze how many tokens each tokenizer assigns to the option ID-delimiter pair across prompt formats.
While some models exhibit accuracy differences when these tokenized pairs expand beyond two tokens, the overall effect is minor and inconsistent across \acp{MLLM}.
For more details, see \cref{app:tokenization_input}.

\begin{figure}[!b]
    \centering
    \includegraphics[width=0.95\linewidth]{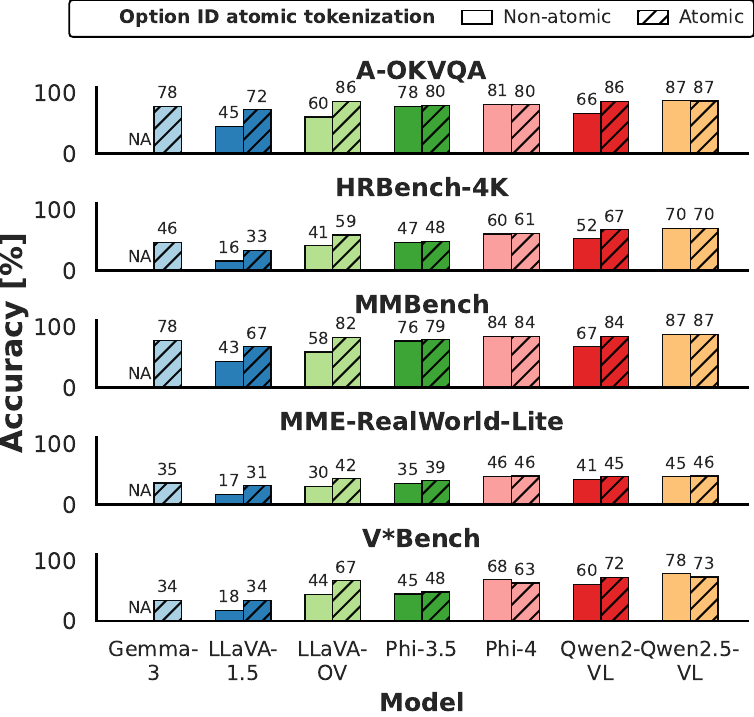}
    \vspace{-8pt}
    \caption{\textbf{Effect of atomic and non-atomic option ID tokenization.}
    We group prompt formats based on whether they preserve or remove the bare option ID token.
    Models whose tokenizers fail to preserve standalone option IDs often show accuracy decrease when the bare option ID token is absent.
    This result links prompt format sensitivity directly to tokenization rather than to changes in task semantics.}
    \label{fig:bareoption}
\end{figure}

\parag{Option ID atomic tokenization}
Next, we examine whether the choice of option ID set and delimiter causes the tokenizer to fuse them into a corrupted token, preventing the option ID from appearing as a standalone token.
For example, a tokenizer may split \texttt{"(A)"} into \texttt{"(A"} and \texttt{")"}, such that the token \texttt{"A"} never appears.
Therefore, we first identify which option ID-delimiter combinations induce this tokenization pattern and then group prompt formats according to whether the pattern occurs for any option ID.
Interestingly, this phenomenon does not occur for the Gemma-3, which consistently preserves atomic option IDs under all evaluated formats.
Correspondingly, Gemma‑3 exhibits the lowest accuracy variation in our study.
In contrast, the absence of a bare option ID token has a pronounced impact for the other models.
We observe a significant accuracy decrease for LLaVA‑1.5, LLaVA‑OV, and Qwen2‑VL in cases where the bare option ID token is missing, indicating a strong dependence on explicit option ID representation in the input.
In contrast, accuracy is stable for Phi‑4 and Qwen2.5‑VL when the bare option ID token is not present, suggesting that these models rely less on isolated option ID tokens and can operate on fused representations.
A key finding of this analysis is that a well-designed tokenizer (e.g., \texttt{GemmaTokenizer}) enables consistent encoding behavior for multiple‑choice prompts independent of prompt format choices.
This supports the claim that robust tokenizer design can substantially mitigate the impact of prompt format variations.
The results of this analysis are essential as the widely used \ac{MC-VQA} benchmarks MMMU and MMMU‑Pro rely on the \texttt{"(A)"} option format \cite{yue_mmmu_2024, yue_2025_mmmupro}.

\begin{table}[hbtp]
    \centering
    \caption{\textbf{Impact of non-atomic option ID tokenization on embeddings.}
    We compare the cosine similarity between fused and original option ID tokens. Models that are sensitive to missing bare tokens exhibit low representational similarity, indicating that fused tokens fail to preserve ID identity and lead to accuracy deviations.}
    \vspace{-7pt}
    \label{tab:cosine_sim_token}
    \resizebox{0.95\linewidth}{!}{
    \begin{tabular}{l c c c}
        \bottomrule
        \textbf{Model} & \textbf{\makecell{Negative \\ effect?}} & \textbf{\makecell{Average \\cos. sim.}} & \textbf{\makecell{Avg. acc. \\deviation $[\%]$}} \\
        \midrule
        Phi-3.5 & \cmark & $0.0110$ & $-5.0$\\
        LLaVA-1.5 & \cmark & $0.0341$ & $-43.4$\\
        Qwen2-VL & \cmark & $0.2617$ & $-18.3$\\
        LLaVA-OV & \cmark & $0.2672$ & $-30.6$\\
        \midrule
        Qwen2.5-VL & $-$ & $0.2390$ & $+0.9$\\
        Phi-4 & $-$ & $0.6139$ & $+4.2$\\
        \midrule
        Gemma-3 & $-$ & $-$ & $-$\\
        \bottomrule
    \end{tabular}%
    }
\end{table}

\noindent To further characterize this effect, we compare the representations of bare option ID tokens and their fused counterparts in input embedding space.
Models that show a negative accuracy effect from missing bare tokens consistently exhibit low cosine similarity between the corrupted and bare option‑ID embeddings (see \cref{tab:cosine_sim_token}).
This indicates that the fused tokens occupy a distant region of embedding space and cannot reliably substitute for the standalone option ID.
For models without a negative effect, we observe that Phi‑4 displays markedly higher cosine similarity, suggesting that option‑ID identity is better preserved despite fusion, whereas Qwen2.5‑VL shows low similarity while remaining unaffected.
For models with low cosine similarity already at the embedding, correct multiple‑choice behavior requires the model to internally remap corrupted option ID tokens to their intended identities during decoding.
This additional representational correction might increase sensitivity to prompt formatting and prior work has shown that such early‑stage tokenization differences can substantially affect multiple‑choice evaluation outcomes, even when task semantics remain unchanged \cite{mielke2021between, sanz-guerrero-etal-2025-mindgap}.
More generally, analyses of tokenizer choice show that tokenizer design can substantially affect downstream model robustness and behavior \cite{altintas2025toksuite}.

\subsection{Option ID Performance Cannot Be Explained Solely by Training Exposure}
\label{subsect:training_exposure_option_IDs}

Tokenization explains part of the observed instability, but it does not explain why some option ID sets consistently perform better even when all IDs tokenize cleanly.
We next test whether this preference is simply a consequence of training exposure.

\begin{figure}[b]
    \centering
    \includegraphics[width=\linewidth]{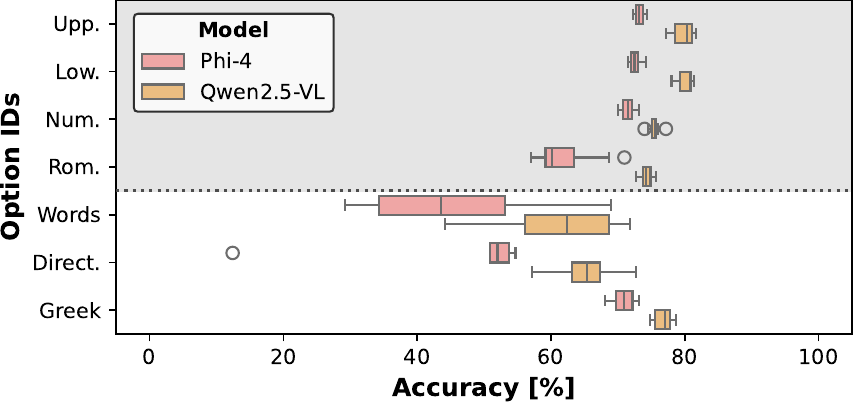}
    \vspace{-20pt}
    \caption{\textbf{Accuracy for alternative option IDs.}
    We evaluate additional option ID sets that differ in training exposure and semantic grounding.
    Greek-letter IDs perform comparably to letter-based IDs and outperform numbers and Roman numerals.
    It shows that ID preferences cannot be explained by training exposure alone.}
    \label{fig:option_id_sets_accuracy}
\end{figure}

\noindent A hypothesis might be that option ID sensitivity reflects training exposure: models may simply perform better on frequently seen IDs (e.g., uppercase letters) than on less common ones.
To probe this, we evaluate Phi-4 and Qwen2.5-VL on V*Bench using additional option ID sets that are unlikely to occur in multiple-choice tasks during training.
We introduce three sets with increasing semantic grounding: (i) Greek letter words, (ii) directional words, and (iii) concrete nouns (see~\cref{tab:additional_option_ids}).
We mitigate output-length effects on accuracy deviations by ensuring that each option ID maps to a single token in both models.
Across both models, Greek-letter IDs outperform numbers and Roman numerals and are comparable to letter-based option IDs (see~\cref{fig:option_id_sets_accuracy}).
This pattern suggests that option ID performance is not solely a consequence of exposure during training.
Instead, \acp{MLLM} can generalize multiple-choice selection to other IDs.

\subsection{Attention Heads Can Be More Discriminative for Certain Option IDs}
\label{subsect:copy_attention_heads}

\begin{table}[!b]
    \centering
    \caption{\textbf{Attention margin to option IDs correlates with \ac{MC-VQA} accuracy.} We compare the cumulative attention margin of attention heads with \ac{MC-VQA} accuracy across option ID sets. Higher margins consistently correlate with higher accuracy, indicating that multiple-choice selection depends on decisive attention. Greek IDs ($\dagger$) are included for an extended analysis.}
    \vspace{-5pt}
    \label{tab:copy_attention_margin}
    \resizebox{0.9\linewidth}{!}{
    \begin{tabular}{lcc}
        \toprule
        \multicolumn{3}{c}{A-OKVQA} \\
        \midrule
        \textbf{Option IDs} & \textbf{\makecell{Attention\\Margin (Avg.)}} & \textbf{\makecell{\ac{MC-VQA} Acc.\\(Avg.) [\%] $\uparrow$}} \\
        \midrule
        Lowercase & 0.609 & 87.51 \\
        Uppercase & 0.596 & 87.64 \\
        Numbers & 0.526 & 85.73 \\
        Roman & 0.509 & 86.19 \\
        \midrule
        \textbf{Pearson $\rho$} & \multicolumn{2}{c}{\textbf{0.930}} \\
        
        \midrule
        \multicolumn{3}{c}{V*Bench} \\
        \midrule
        \textbf{Option IDs} & \textbf{\makecell{Attention\\Margin (Avg.)}} & \textbf{\makecell{\ac{MC-VQA} Acc.\\(Avg.) [\%] $\uparrow$}} \\
        \midrule
        Lowercase & 0.651 & 80.10 \\
        Uppercase & 0.648 & 79.58 \\
        Greek$^{\dagger}$ & 0.620 & 76.69 \\
        Numbers & 0.543 & 75.45 \\
        Roman & 0.499 & 74.34 \\
        \midrule
        \textbf{Pearson $\rho$} & \multicolumn{2}{c}{\textbf{0.934}} \\
        \bottomrule
    \end{tabular}
    }
\end{table}

\cref{subsect:training_exposure_option_IDs} shows that some option ID sets yield higher \ac{MC-VQA} accuracy independent of training exposure.
We hypothesize that these differences arise from how decisively models attend to candidate option IDs during decoding.

\noindent In (M)LLMs, attention heads can specialize in distinct behaviors, e.g., select-and-copy behavior \cite{olsson_inductionheads_2022, bietti_memory_2023, dangelo_selectiveheads_2025}.
In multiple-choice settings, this mechanism is relevant because the model only needs to reproduce an option ID rather than generate a free-form answer \cite{tulchinskii_listening_2024}.
We analyze this behavior by tracking attention to option ID tokens across varying ID sets on A-OKVQA and V*Bench.
To isolate the effect of ID set variation, we fix the option delimiter and separator.
We measure the decisiveness of option selection using the margin between the top-1 and top-2 attention weights to option IDs.
Details are provided in \cref{app:copy_attention_heads}.

\noindent We identify a set of attention heads on V*Bench that consistently exhibit strong attention to option IDs and function as copy-style heads. To test whether this behavior generalizes beyond the dataset used for attention head selection, we apply the same heads to A-OKVQA.

\noindent Across both datasets, option ID sets with higher \ac{MC-VQA} accuracy exhibit larger cumulative attention margins across these heads (see \cref{tab:copy_attention_margin}).
For V*Bench, we additionally include Greek option IDs (see \cref{subsect:training_exposure_option_IDs}) to test whether the attention patterns generalize to uncommon ID sets.
\ac{MC-VQA} accuracy correlates strongly with the copy-attention margin across ID sets ($\rho = 0.930$ on A-OKVQA and $\rho = 0.934$ on V*Bench including Greek option IDs).
This result indicates that the effect is not specific to a single dataset.
Instead, \ac{MC-VQA} accuracy depends in part on how decisively the model selects and reliably reproduces an option ID during generation.
These findings show that prompt-format sensitivity arises not only from tokenization effects, but also from decoding behavior.
The same copy-style attention heads remain predictive of option ID performance across datasets, indicating that prompt formatting influences the reliability of the option-selection mechanism itself.

\subsection{MLLMs Can Reason About Favorable Prompt Formats For Multiple-Choice}
\label{subsec:models_infer_good_prompt_factors}

Previously, we showed that model-specific encoding and decoding behavior causes format sensitivity.
We now test whether \acp{MLLM} can predict which prompt formats are more favorable for their own performance.
Therefore, we present each model with the design space in \cref{tab:prompt_format_variations} and ask it to select a preferred configuration for each formatting factor.
Additional details are provided in \cref{app:models_infer_good_prompt_factors}.
We find that all \acp{MLLM}, except Phi-4, select similar or identical prompt formats, and that the selected formats consistently outperform random guessing\footnote{Randomly selecting one of the $48$ prompt formats yields an expected prompt-format rank of $24$.}, achieving average ranks between $5.4$ and $14.4$ across datasets.
We observe similar behavior for the additional option ID sets introduced in \cref{subsect:training_exposure_option_IDs} (see \cref{tab:model_selected_prompt_formats_2}).
These results suggest that prompt format preferences are structured and partially aligned with model-specific processing.
However, they do not eliminate the need to control for prompt format effects in evaluation.

\begin{table}[h]
    \centering
    \caption{\textbf{Model-selected prompt formats.}
    Each model selects a prompt format from \cref{tab:prompt_format_variations}.
    Most models choose format with lower average ranks, which suggests that prompt format preferences are recoverable from \acp{MLLM}.
    Lower average rank indicates better performance.
    We further report the relative accuracy difference between each selected prompt format and the best-performing prompt format across all datasets.}
    \vspace{-5pt}
    \label{tab:model_selected_prompt_formats}
    \resizebox{\linewidth}{!}{
    \begin{tabular}{l c c c c c}
        \toprule
        \textbf{Model} & \textbf{Option IDs} & \textbf{Delimiter} & \textbf{Separator} & \textbf{\makecell{Avg.\\ Rank}} & \textbf{\makecell{Rel. gap\\to optimal}}\\
        \midrule
        Gemma-3    & Uppercase & Colon   & Linebreak & \colornum{14.4} & $-1.61$ \\
        LLaVA-1.5  & Uppercase & Colon   & Linebreak & \colornum{7.0} & $-3.12$  \\
        LLaVA-OV   & Uppercase & Bracket & Linebreak & \colornum{5.4} & $-0.79$  \\
        Phi-3.5    & Uppercase & Colon   & Linebreak & \colornum{9.6} & $-1.42$  \\
        Phi-4      & Numbers   & Dot     & Comma     & \colornum{34.0} & $-4.21$ \\
        Qwen2-VL   & Uppercase & Colon   & Linebreak & \colornum{11.6} & $-1.08$ \\
        Qwen2.5-VL & Uppercase & Colon   & Linebreak & \colornum{7.4} & $-0.92$  \\
        \bottomrule
    \end{tabular}%
    }
\end{table}

\subsection{Multiple-Choice VQA May Not Fully Reflect Open-Ended MLLM Performance}
\label{subsect:mc_vqa_weak_proxy}

\begin{figure}[b]
    \centering
    \includegraphics[width=\linewidth]{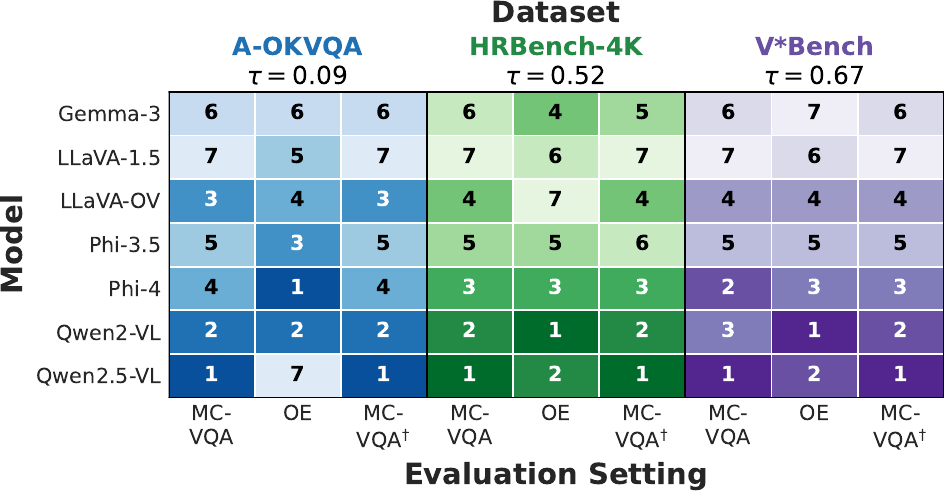}
    \vspace{-18pt}
    \caption{\textbf{Open-ended and grounded multiple-choice evaluation.}
    We compare \ac{MC-VQA}, open-ended (OE), and grounded multiple-choice (MC-VQA$^{\dagger}$).
    Kendall’s $\tau$ indicates weak agreement between MC-VQA and OE, and MC-VQA$^{\dagger}$ does not recover the original ordering.}
    \label{fig:open_ended_evaluation}
\end{figure}

The previous analyses show that \ac{MC-VQA} results depend strongly on how models encode and decode option IDs.
We now test the broader implication of this result: whether \ac{MC-VQA} remains a reliable proxy for open-ended \ac{MLLM} performance.

\noindent Therefore, we compare multiple-choice and open-ended evaluation (see~\cref{fig:open_ended_evaluation}).
For open-ended evaluation, we remove answer options and prompt models to generate free-form responses (see~\cref{app:conversion_open_ended}).
We assess correctness using an LLM-as-a-judge framework (see~\cref{app:correctness_assessment}).
Across all datasets, rankings correlate weakly to moderately between the two settings: Kendall’s~$\tau$ is $0.09$ on A‑OKVQA, $0.52$ on HRBench‑4K, and $0.67$ on V*Bench.

\noindent To further analyze option selection effects, we perform \emph{grounded multiple-choice} evaluation by providing each model its open-ended answer (see~\cref{app:grounded_mc_vqa}).
This setup provides context while preserving the multiple-choice setting.
Grounding improves \ac{MC-VQA} accuracy for most models compared to open-ended, but does not consistently restore original rankings.
More importantly, it does not eliminate prompt sensitivity across prompt format variations (see \cref{tab:full_results_grounded_mc_vqa}).
This indicates that multiple-choice selection remains influenced even when high-level reasoning signals are available.

\noindent Our results suggest that \ac{MC-VQA} captures both reasoning ability and option selection behavior. The agreement between multiple-choice and open-ended rankings varies substantially across datasets, ranging from minor correlation on A-OKVQA ($\tau=0.09$) to moderate correlation on HRBench-4K ($\tau=0.52$) and V*Bench ($\tau=0.67$).
Consequently, \ac{MC-VQA} should be interpreted as a distinct evaluation setting rather than as a direct measure of open-ended \ac{MLLM} performance.

\input{tables/simple_mitigation_approach}

\subsection{Practical Recommendations}
Based on our findings, we want to provide several practical recommendations for \ac{MC-VQA} evaluation.
First, \ac{MC-VQA} dataset creators should explicitly report the choice of the prompt format (i.e., option IDs, delimiter and separator) for the evaluation on the respective dataset.
Second, benchmark results should ideally be complemented with open-ended evaluation, since \ac{MC-VQA} reflects both reasoning and option-selection behavior.
Third, practitioners should verify that prompt formats preserve standalone option ID tokens, as non-atomic tokenization can cause substantial accuracy degradation for some models.
Finally, we evaluate a simple mitigation strategy that selects a model-specific prompt format on a small validation set and reuses it during evaluation (see \cref{tab:mitigation}).
This approach reduces extreme prompt-format failures and can improve performance relative to standard formats.
However, gains are inconsistent across datasets and models, and the method still requires evaluating multiple prompt formats on a validation set.
We therefore view validation-based prompt format selection as a practical mitigation rather than a complete solution to prompt-format sensitivity.

\section{Conclusion}
\label{sect:conclusion}

We show that multiple-choice VQA is sensitive to semantically neutral prompt format choices.
Variations in option IDs, delimiters, and separators induce large accuracy shifts and frequent rank reversals, even under order-invariant evaluation.
We trace this instability to low-level language-model behavior: tokenization can fuse option ID tokens under certain prompt formats, and different option ID sets activate specialized attention heads to varying degrees.
Interestingly, the performance differences across option ID sets are not solely explained by training exposure.
Further, we show that models can also identify prompt formats that improve performance.
Finally, the weak to moderate correlation with open-ended evaluation indicates that multiple-choice accuracy is not a reliable proxy for general \ac{MLLM} capability.

\section*{Limitations}
\label{sect:limitations}

Our work has the following limitations:
(i) \emph{Evaluation setting:} We focus on zero-shot \ac{MC-VQA} with deterministic decoding to enable controlled comparisons. Results may differ under few-shot prompting, or stochastic decoding, where sampling may interact differently with prompt format.
(ii) \emph{Evaluation methodology:} Our open-ended evaluation relies on an LLM-as-a-judge framework, which may introduce bias. We do not systematically validate this protocol across all datasets. In grounded \ac{MC-VQA}, models also receive unfiltered generated answers, which may introduce noise when incorrect responses are provided. Further, \acp{LLM} from the same model family as some of the tested \acp{MLLM} are used to generate the open-ended answer and for the LLM-as-a-judge correctness assertion.
(iii) \emph{Mitigation:} We show that models can partially reason effective prompt formats, revealing that prompt format preferences may be structured and accessible to the models. Further, selecting prompts on a small validation set can reduce extreme performance drops, indicating limited cross-dataset generalization. However, it does not consistently improve performance, and remains computationally expensive due to the large number of prompt formats to evaluate. These findings provide a concrete direction for developing more robust evaluation protocols in future work.

\section*{Acknowledgments}
\label{sect:acknowledgments}
During the preparation of this paper, the first author's father passed away. The authors would like to honor his memory and express profound gratitude for his support, encouragement, and belief in the value of education and scientific inquiry.

\section*{Disclaimer}
The results, opinions, and conclusions expressed in this publication are not necessarily those of Volkswagen Aktiengesellschaft.

\bibliography{custom}

\input{secs/appendix}

\end{document}

%% file: tables/simple_mitigation_approach.tex
\begin{table*}[t]
    \centering
    \caption{\textbf{Prompt-format transfer as a simple bias mitigation strategy.} We select the best prompt format on a small validation benchmark (HRBench-4K or V*Bench) per \ac{MLLM} and apply the same format to all other benchmarks. While this strategy often reduces severe accuracy drops, the selected format does not transfer reliably across datasets and fails to consistently improve accuracy. Values denote the accuracy difference in \ac{pp} relative to each dataset's standard format. $^\dagger$ denotes the benchmark used for prompt-format selection.}
    \label{tab:mitigation}
    \vspace{-6pt}
    \begingroup
    \small
    \renewcommand{\arraystretch}{1}
    \setlength{\tabcolsep}{3pt}
    \resizebox{0.95\textwidth}{!}{%
    \begin{tabular}{l *{4}{>{\centering\arraybackslash}m{1.5cm}} c *{4}{>{\centering\arraybackslash}m{1.5cm}} }
        \toprule
        & \multicolumn{9}{c}{\textbf{Accuracy difference to the dataset-specific standard prompt format [pp.]}} \\[2pt]
        & \multicolumn{4}{c}{{Prompt format selection dataset: HRBench-4K$^\dagger$}} & & \multicolumn{4}{c}{{Prompt format selection dataset: V*Bench$^\dagger$}} \\
        \cmidrule(lr){2-5} \cmidrule(lr){7-10}
        \textbf{Model} & A-OKVQA & MMBench & MME-RW & {V*Bench} & & {A-OKVQA} & {HRBench} & {MMBench} & {MME-RW} \\
    
        \midrule
        
        Gemma-3 & {\color{Bittersweet}\num{-0.57}} & {\color{OliveGreen}\num{0.14}} & {\color{Bittersweet}\num{-0.80}} & {\color{Bittersweet}\num{-0.17}} & & {\color{Bittersweet}\num{-0.68}} & {\color{OliveGreen}\num{0.63}} & {\color{OliveGreen}\num{0.39}} & {\color{Bittersweet}\num{-0.92}} \\
        LLaVA-1.5 & {\color{OliveGreen}\num{65.17}} & \num{0.00} & {\color{Bittersweet}\num{-1.13}} & {\color{Bittersweet}\num{-1.01}} & & {\color{OliveGreen}\num{59.89}} & {\color{Bittersweet}\num{-3.63}} & {\color{Bittersweet}\num{-3.26}} & {\color{Bittersweet}\num{-3.42}} \\
        LLaVA-OV & {\color{OliveGreen}\num{48.21}} & {\color{OliveGreen}\num{0.33}} & {\color{Bittersweet}\num{-0.27}} & \num{0.00} & & {\color{OliveGreen}\num{48.56}} & {\color{OliveGreen}\num{0.25}} & {\color{OliveGreen}\num{0.70}} & {\color{OliveGreen}\num{0.25}} \\
        Phi-3.5 & {\color{OliveGreen}\num{2.38}} & {\color{OliveGreen}\num{2.73}} & \num{0.00} & {\color{Bittersweet}\num{-1.01}} & & {\color{OliveGreen}\num{2.34}} & \num{0.00} & {\color{OliveGreen}\num{2.87}} & {\color{OliveGreen}\num{0.89}} \\
        Phi-4 & {\color{Bittersweet}\num{-0.07}} & {\color{OliveGreen}\num{0.80}} & {\color{Bittersweet}\num{-0.28}} & {\color{Bittersweet}\num{-0.50}} & & {\color{OliveGreen}\num{0.02}} & \num{0.00} & {\color{OliveGreen}\num{1.03}} & {\color{OliveGreen}\num{0.07}} \\
        Qwen-2-VL & {\color{OliveGreen}\num{40.96}} & {\color{OliveGreen}\num{0.74}} & {\color{OliveGreen}\num{1.42}} & {\color{Bittersweet}\num{-2.68}} & & {\color{OliveGreen}\num{41.09}} & {\color{Bittersweet}\num{-0.13}} & {\color{OliveGreen}\num{1.62}} & {\color{OliveGreen}\num{0.91}} \\
        Qwen-2.5-VL & {\color{Bittersweet}\num{-0.15}} & \num{0.00} & {\color{OliveGreen}\num{0.97}} & {\color{Bittersweet}\num{-1.51}} & & \num{0.00} & \num{0.00} & {\color{Bittersweet}\num{-0.21}} & {\color{Bittersweet}\num{-0.38}} \\
        \midrule
        \textbf{Avg.} & {\color{OliveGreen}\textbf{\num{22.28}}} & {\color{OliveGreen}\textbf{\num{0.68}}} & {\color{Bittersweet}\textbf{\num{-0.01}}} & {\color{Bittersweet}\textbf{\num{-0.98}}} & & {\color{OliveGreen}\textbf{\num{21.60}}} & {\color{Bittersweet}\textbf{\num{-0.41}}} & {\color{OliveGreen}\textbf{\num{0.45}}} & {\color{Bittersweet}\textbf{\num{-0.37}}} \\
        \bottomrule
    \end{tabular}%
    }
    \endgroup
\end{table*}

%% file: secs/appendix.tex
\clearpage
\appendix

\renewcommand{\thesection}{\Alph{section}}

\crefformat{appendix}{App.~#2#1#3}
\Crefformat{appendix}{App.~#2#1#3}

\section{Experimental Setup}
\label{app:experimental_setup}

In this section, we provide additional technical details on the following components: instruction prompts (\cref{app:instruction_prompts}), evaluation settings (\cref{app:evaluation_settings}), metrics (\cref{app:metrics}), models and datasets (\cref{app:models_datasets}), the computation report (\cref{app:computation_report}), and bias mitigation methods (\cref{app:bias_mitigation_methods}).

\subsection{Instruction Prompts}
\label{app:instruction_prompts}

\input{tables/instruction_prompts}

As shown in \cref{fig:prompt_template}, instruction prompts are essential for prompting in \ac{MC-VQA} tasks.
They guide \acp{MLLM} to output only the corresponding option ID for the selected option \cite{duan_vlmevalkit_2024}.
In practice, many datasets use simple instruction prompts \cite{wu_vstar_2024, wang_hrbench_2025, schwenk_aokvqa_2022, zhang_mmerealworld_2025, duan_vlmevalkit_2024}; see the \textit{Reference} instruction prompt from V*Bench in \cref{tab:instruction_prompts}.
Vague instruction prompts, such as failing to specify that the \ac{MLLM} should return the option ID rather than the option text, can lead to poor instruction-following of \acp{MLLM}.
This can lead to poor instruction-following and consequently lower accuracy, not due to reasoning errors but simply because of inadequate prompt design.
Therefore, we use an optimized instruction prompt that clearly specifies what the \acp{MLLM} should generate and what it should avoid in all experiments of our comprehensive analysis.
Additionally, the prompt includes the exact option IDs that the models are expected to generate (see examples in \cref{tab:instruction_prompts}).

\subsection{Evaluation Settings}
\label{app:evaluation_settings}

We evaluate all models in a fully deterministic setting by disabling sampling.
Specifically, we set \texttt{do\_sample=False} and $k=0$ during generation for each \ac{MLLM}.
This ensures that each model produces identical outputs for identical inputs.
Before evaluating an \ac{MLLM} on a given prompt format, we iterate over all option IDs required by the dataset.
We tokenize each option ID and set the number of generated tokens to the maximum required to generate all option IDs.
Furthermore, we evaluate the seven \acp{MLLM} across all \ac{MC-VQA} datasets in a zero-shot setting, i.e. without providing any in-context learning examples, as is standard for this task \cite{duan_vlmevalkit_2024}.
Further, we use Flash-Attention for all model evaluations besides the attention head analysis.

\subsection{Metrics}
\label{app:metrics}

\parag{Accuracy}
The most commonly used metric for evaluating the reasoning ability of \acp{MLLM} in \ac{VQA} is \textit{accuracy} \cite{vqav2_goyal_2017, antol_vqa_2015}.
The accuracy is defined as the ratio of correctly answered questions to the total number of questions in the evaluation set \cite{vqav2_goyal_2017, antol_vqa_2015}.
The correctness assessment is performed directly on the model’s generated output sequence without applying any answer extraction mechanism (e.g., regular expressions).
An answer is considered correct only if it exactly matches the ground-truth label.
For instance, if the label is \texttt{"A"}, then \texttt{"A"} is correct, whereas \texttt{"A."} or \texttt{"A)"} are not.

\parag{Coverage}
To assess instruction-following capability, we introduce the metric \textit{coverage}. Coverage is defined as the ratio of answers that belong to the valid option ID set to the total number of questions in the evaluation set:
\begin{equation}
    \text{coverage} = \frac{s_i}{n}
\end{equation}
\noindent where \(s_i\) denotes the number of answers within the option ID set and \(n\) is the total number of questions. A higher coverage value indicates better instruction-following capability.

\subsection{Models and Datasets}
\label{app:models_datasets}

\parag{Models}
We evaluate a set of seven widely used \acp{MLLM} (see \cref{tab:model_overview}), each selected to have a comparable \ac{LLM} parameter size.
We consider three categories of models: (i) those that process a single downsized global view of the input image (LLaVA-1.5), (ii) those that combine a global view with additional high-resolution local crops (Gemma-3, LLaVA-OV, Phi-3.5, and Phi-4), and (iii) those that operate on a high-resolution global view of the input image (Qwen-2-VL and Qwen-2.5-VL).

\input{tables/model_overview}

\parag{Datasets}
For our comprehensive analysis, we employ five \ac{MC-VQA} datasets that span a range of task difficulties and domains.
Additional details about these datasets are provided in \cref{tab:dataset_overview}.

\input{tables/datasets}

\subsection{Computation Report}
\label{app:computation_report}

We provide an overview of the computational complexity involved in our comprehensive analysis of prompt format-induced biases in \ac{MC-VQA}, including the number of questions and inference time, in \cref{tab:computation_report}.

\input{tables/computation_report}

\subsection{Bias Mitigation Methods}
\label{app:bias_mitigation_methods}

We assess the effectiveness of several bias mitigation techniques in reducing prompt format-induced biases.
All methods are reimplemented according to the specifications provided in their papers.

\parag{Circular evaluation}
The circular evaluation scheme shifts the options in the prompt by one position \cite{liu_mmbench_2024} and thereby increases the number of questions - e.g., for four options it enlarges the dataset by a factor of four.
This mitigates the option position bias.
While the approach requires no retraining or calibration, it is computationally more expensive.

\parag{\Acf{PIA}}
\ac{PIA} is a post-hoc metric that mitigates the influence of the option position bias by weighting the per-option accuracy by the number of times an \ac{MLLM} selects a specific option \cite{tan_ordermatters_2024}:

\begin{equation}
    \text{PIA} = \frac{1}{M} \sum^M_{i=1} \frac{C_i}{Pr_i} \frac{C_i}{N}
\end{equation}

\noindent where $M$ is the number of options, $C_i$ is the number of times option $i$ was correctly selected, $Pr_i$ is the number of times option $i$ was selected in total and $N$ is the number of questions \cite{tan_ordermatters_2024}.
This method does not require a calibration; however, it is prone to variations in the prompt format, as these changes may influence the selection frequency of \acp{MLLM}.
Furthermore, it changes the accuracy calculation, which has two disadvantages: (i) this measure is less intuitive and not comparable to standard accuracy, and (ii) it can skew the result, e.g., when a model performs overall well but it does not match the dataset's positional distribution or when a model answers one option only rarely.

\parag{\Acf{PriDe}}
\ac{PriDe} estimates the prior for each option ID based on calibration data containing circularly shifted multiple-choice questions \cite{zheng_llmsnotrobust_2024}.
This prior is then used to debias the predictions of an \ac{MLLM}, adjusting the probability of each option ID according to its corresponding prior:
\begin{equation}
    P_{\text{debiased}}(o_i \mid q, x) \propto \frac{P_{\text{observed}}(d_i \mid q, x)}{P_{\text{prior}}(d_i)}
\end{equation}
\noindent where $o_i$ denotes the $i$-th option, $q$ is the question, $x$ represents the default ordering of option IDs, and $d_i$ is the $i$-th option ID \cite{zheng_llmsnotrobust_2024}.
The answer is then selected as the option ID with the highest debiased probability \cite{zheng_llmsnotrobust_2024}.
For our evaluations with \ac{PriDe}, we estimate this prior using the 4-option questions from V*Bench.
These priors are only valid for questions with the same number of options as in the calibration data; therefore, they cannot be applied to datasets with varying numbers of options.

\parag{\acf{CP-LN}}
An alternative approach to mitigating biases in \acp{MLLM} involves modifying the answer generation process.
Rather than presenting all options to the model together with a predefined prompt format and instruction, this method calculates the perplexity of the question for each individual option.
Perplexity serves as a metric for \acp{MLLM}, indicating how unexpected it is for the model to generate an output sequence $\mathbf{y}$ given the input token sequence $\mathbf{X}_{\text{in}}$ and the previously generated tokens $y_{<i}$ \cite{bengio_neuralmodel_2003, brown_llmsfewshot_2020, wei_cogreg_2021}:
\begin{equation}
    \text{PPL}(\mathbf{X}_{\text{in}})^\mathbf{y} = \left( \prod_i^n P(y_i|\mathbf{X}_{\text{in}},  y_{<i}) \right) ^{-\frac{1}{n}}
\end{equation}
\noindent where $n$ is the number of tokens in the sequence.
A lower perplexity value indicates a more likely prediction of the output token sequence than a higher one.
Unlike the sum of log-probabilities, perplexity is normalized and thus less sensitive to the sequence length $n$.  
Given the set of perplexities $\text{PPL}^o(X)$ computed by the \ac{MLLM} for input $\mathbf{X}_{\text{in}}$ conditioned on each option $o$, we select the predicted answer $\hat{Y}$ as the option with the lowest perplexity among all options $O$:

\begin{equation}
    \hat{Y} = \arg\min_{o \in O} \big(\text{PPL}^o(\mathbf{X}_{\text{in}})\big)
\end{equation}

\noindent This approach is closely related to the methods used in \cite{robinson_leveraging_2023, brown_llmsfewshot_2020}. It has been shown to be less effective than the standard multiple-choice evaluation scheme in terms of accuracy \cite{robinson_leveraging_2023}, while its impact on bias mitigation remains unclear.

\clearpage
\section{Significance Analysis}
\label{app:significance_analysis}

We test whether prompt formats have a statistically significant effect on accuracy and, if so, which factors drive this effect.

\subsection{Linear Mixed Models}
\label{app:linear_mixed_models}

For the significance analysis of prompt format variations, we employ a \acf{LMM} to assess the impact of prompt format factors such as \textit{option ID set}, \textit{option delimiter}, and \textit{option separator} on accuracy.
Our comprehensive study spans multiple performance levels, including diverse datasets and \acp{MLLM}.
Standard linear models (e.g., linear regression) are unsuitable for this non-independent data due to their independence assumption \cite{jiang_linear_2021}.
In contrast, LMMs handle such dependencies by modeling the target variable $\mathbf{y}$ as:

\begin{equation}
    \mathbf{y} = X\boldsymbol{\beta} + Z\mathbf{u} + \boldsymbol{\epsilon}
\end{equation}

\noindent where $\mathbf{y}$ is the vector of observations, $\boldsymbol{\beta}$ the fixed effects, and $\mathbf{u}$ the random effects \cite{jiang_linear_2021}.
The design matrices $X$ and $Z$ capture the relationships between $\mathbf{y}$ and $\beta$, and between $\mathbf{y}$ and $u$, respectively \cite{jiang_linear_2021}.
\noindent In our application, we want to model the accuracy $\text{Acc}_{ijk}$, therefore, our modeling design of the \ac{LMM} is:
\begin{equation}
    \begin{aligned}
        \text{Acc}_{ijk} =\;& \beta_0 + \beta_1\text{OptionID}_i + \beta_2\text{OptionsDelim}_j\\
        &+ \beta_3\text{OptionsSep}_k + u_{g} + \epsilon_{ijk}
    \end{aligned}
\end{equation}
\noindent where $\beta_0$ is the intercept, $\beta_1, \beta_2, \beta_3$ are fixed-effect coefficients for the prompt format factors, $u_{g}$ is the random effect for group $g$ (model-dataset), and $\epsilon_{ijk}$ is the residual error.
This specification estimates the effects of prompt format variations on accuracy while accounting for non-independence across different models and datasets.

\noindent Specifically, we fit a \ac{LMM} using \texttt{statsmodels} in Python to estimate the influence of prompt format variations on accuracy.

\subsection{Significance Testing Details}
\label{app:significance_testing_details}

\begin{figure}[hbtp]
    \centering
    \includegraphics[width=\linewidth]{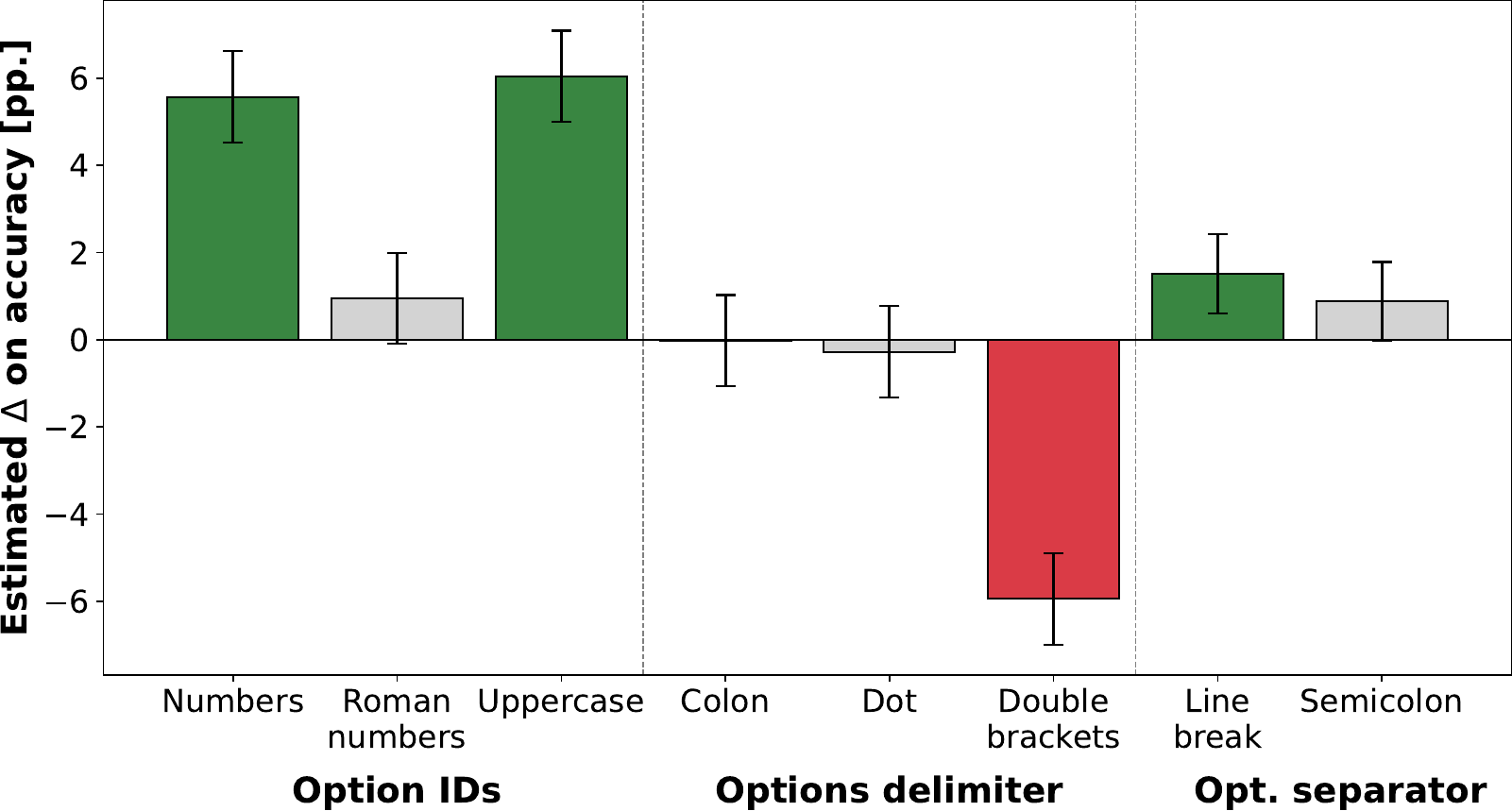}
    \vspace{-15pt}
    \caption{
        \textbf{Prompt formatting choices can have significant effects on \ac{MLLM} accuracy, even after controlling for model and dataset differences.}
        Bars show the effect of each prompt format factor, with error bars indicating $95\%$ confidence intervals. Significant effects are colored by their direction, and non-significant effects are gray. 
        All effects are interpreted relative to the base configuration (\textit{option IDs: lowercase}, \textit{option delimiter: bracket}, \textit{option separator: comma}).
    }
    \label{fig:significance_prompt_format}
\end{figure}

We employ a \acf{LMM} to investigate which biases have a statistical influence on the performance of \acp{MLLM}.
\acp{LMM} model fixed effects (e.g., prompt format factors) and random effects (e.g., combined group of model and dataset), thereby yielding less biased and more generalizable results than traditional linear models (e.g., linear regression) \cite{jiang_linear_2021}.
We control for performance differences across \acp{MLLM} and datasets by incorporating random effects, acknowledging differences in model capabilities and dataset difficulty.
The results of the \ac{LMM}, shown in \cref{fig:significance_prompt_format}, reveal the estimated accuracy change in \ac{pp} when varying each prompt format factor from a baseline configuration, which in our case consists of \textit{lowercase} (option IDs), \textit{bracket} (option delimiter), and \textit{comma} (option separator). 
We find significant effects for all three prompt format factors.
Changing the option IDs yields positive effects for \textit{Uppercase} ($+6$ \ac{pp}) and \textit{Numbers} ($+5$ \ac{pp}).
Changing the delimiter to \textit{Double brackets} results in a negative effect ($-6$ \ac{pp}), while using a \textit{Line break} as separator yields a small positive effect ($+1$ \ac{pp}).

\clearpage
\section{Additional Results}
\label{app:additional_results}

This section provides additional results supporting the instability results in \cref{sect:benchmark_instability}.
We provide detailed results on instruction-following analysis in \cref{app:instruction_following_analysis}, option selection patterns in \cref{app:option_selection_patterns}, bias mitigation results in \cref{app:bias_mitigation_results}, and the full benchmark results in \cref{app:full_benchmark_results}.

\subsection{Instruction-Following Analysis}
\label{app:instruction_following_analysis}

\begin{figure*}[b]
    \centering
    \includegraphics[width=0.95\textwidth]{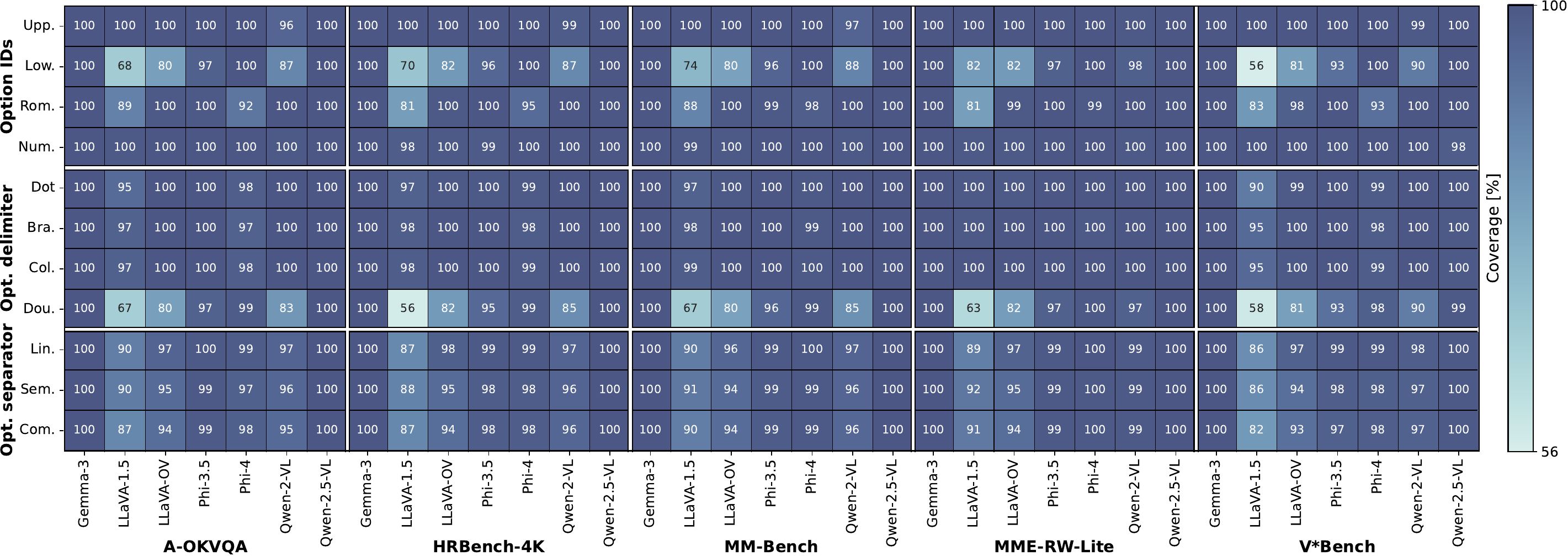}
    \vspace{-5pt}
    \caption{\textbf{Coverage results of \acp{MLLM} on \ac{MC-VQA} datasets.} The results are averaged over 12 evaluations for \textit{option IDs} and \textit{option delimiter}, and 16 for \textit{option separator}. While models such as Gemma-3, LLaVA-OV, Phi-3.5, Phi-4, Qwen2-VL, and Qwen2.5-VL consistently achieve coverage above $75\%$, indicating stable instruction-following capabilities, only LLaVA-1.5 exhibits minor drops under certain prompt formats, revealing strong sensitivity to formatting variations.}
    \label{fig:coverage_heatmap}
\end{figure*}

As previously noted, a key factor affecting \ac{MLLM} accuracy in \ac{MC-VQA} tasks is the instruction-following capability of the models.
We measure this as the \textit{coverage} of given answers (see \cref{app:metrics}).
Low coverage reflects poor instruction-following, limiting accuracy because only part of the provided answers appear in the option ID set.
We report the coverage results across all datasets in \cref{fig:coverage_heatmap}.
The coverage trends are consistent across datasets, with minimal deviation.
Most \acp{MLLM} maintain near-perfect coverage.
Gemma-3 and Qwen-2.5-VL achieve nearly $100\%$ coverage across all datasets and prompt formats.
In contrast, LLaVA-1.5 and LLaVA-OV show slight degradation, with LLaVA-1.5 dropping to a minimum of $56\%$ for lowercase option IDs on V*Bench.
Overall, these results indicate that instruction-following limitations play only a minor role in the observed accuracy deviations across prompt format variations.

\subsection{Option Selection Patterns}
\label{app:option_selection_patterns}

\begin{figure*}[hbtp]
    \centering
    \includegraphics[width=0.95\textwidth]{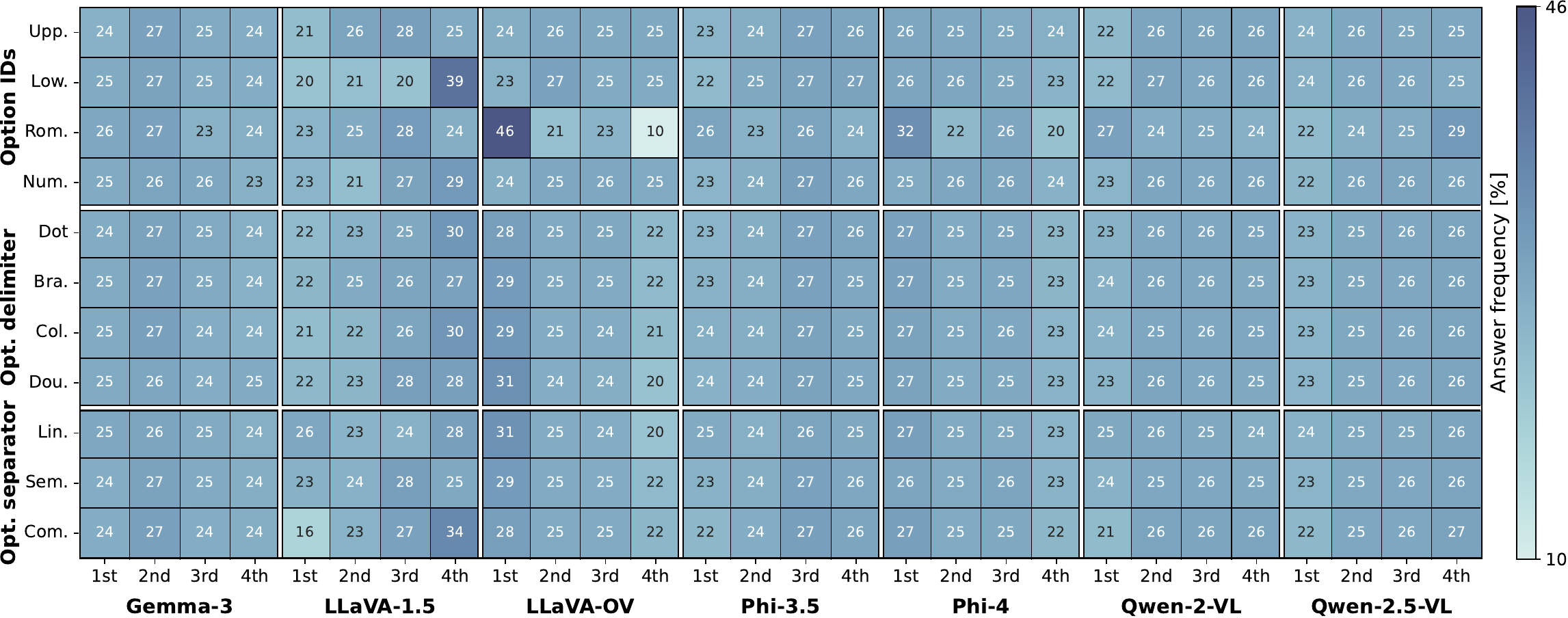}
    \vspace{-8pt}
    \caption{\textbf{Answer behavior of \acp{MLLM} on the A-OKVQA dataset.} We report the average answering frequency per option position for all seven \acp{MLLM} across the prompt format variations. The results are averaged over 12 evaluations for \textit{option IDs} and \textit{option delimiter}, and 16 for \textit{option separator}. Option position bias is mitigated via a circular evaluation scheme.}
    \label{fig:answer_freq_aokvqa}
\end{figure*}

\begin{figure*}[hbtp]
    \centering
    \includegraphics[width=0.95\textwidth]{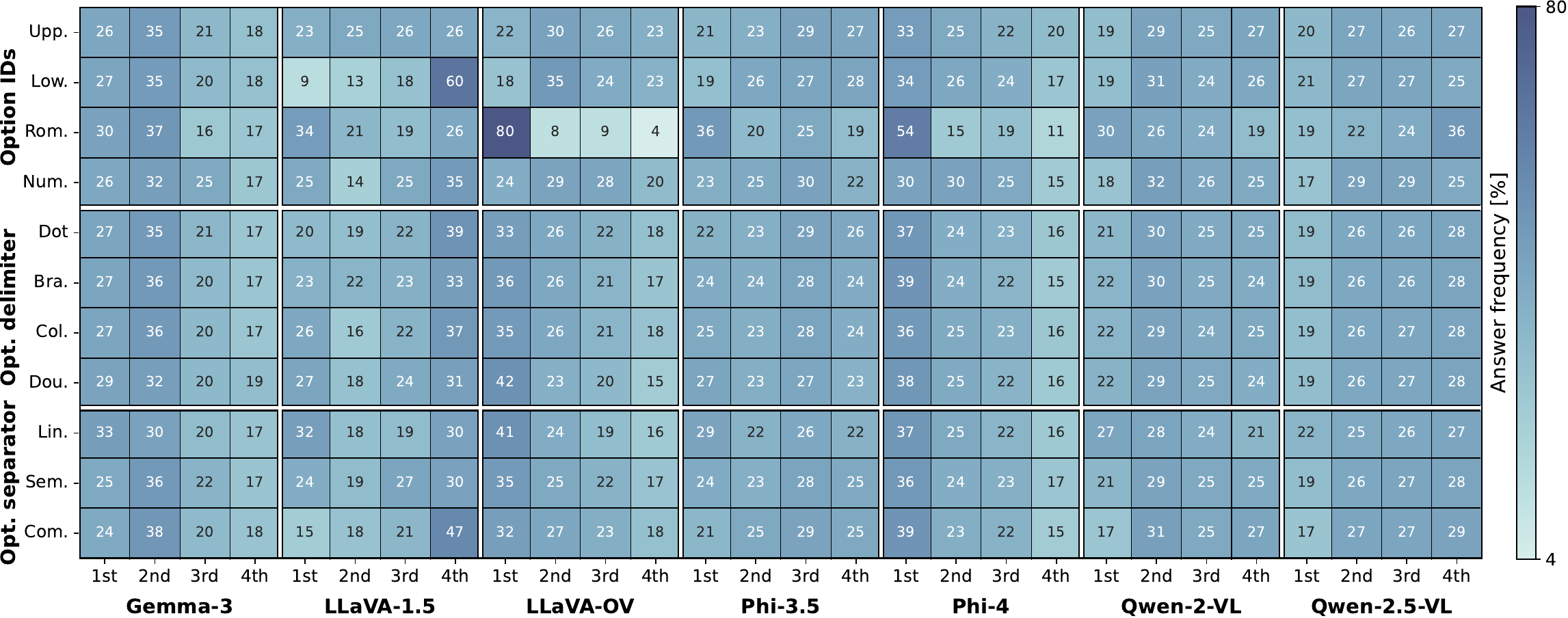}
    \vspace{-8pt}
    \caption{\textbf{Answer behavior of \acp{MLLM} on the HRBench-4K dataset.} We report the average answering frequency per option position for all seven \acp{MLLM} across the prompt format variations. The results are averaged over 12 evaluations for \textit{option IDs} and \textit{option delimiter}, and 16 for \textit{option separator}. Option position bias is mitigated via a circular evaluation scheme.}
    \label{fig:answer_freq_hrbench}
\end{figure*}

\begin{figure*}[hbtp]
    \centering
    \includegraphics[width=0.95\textwidth]{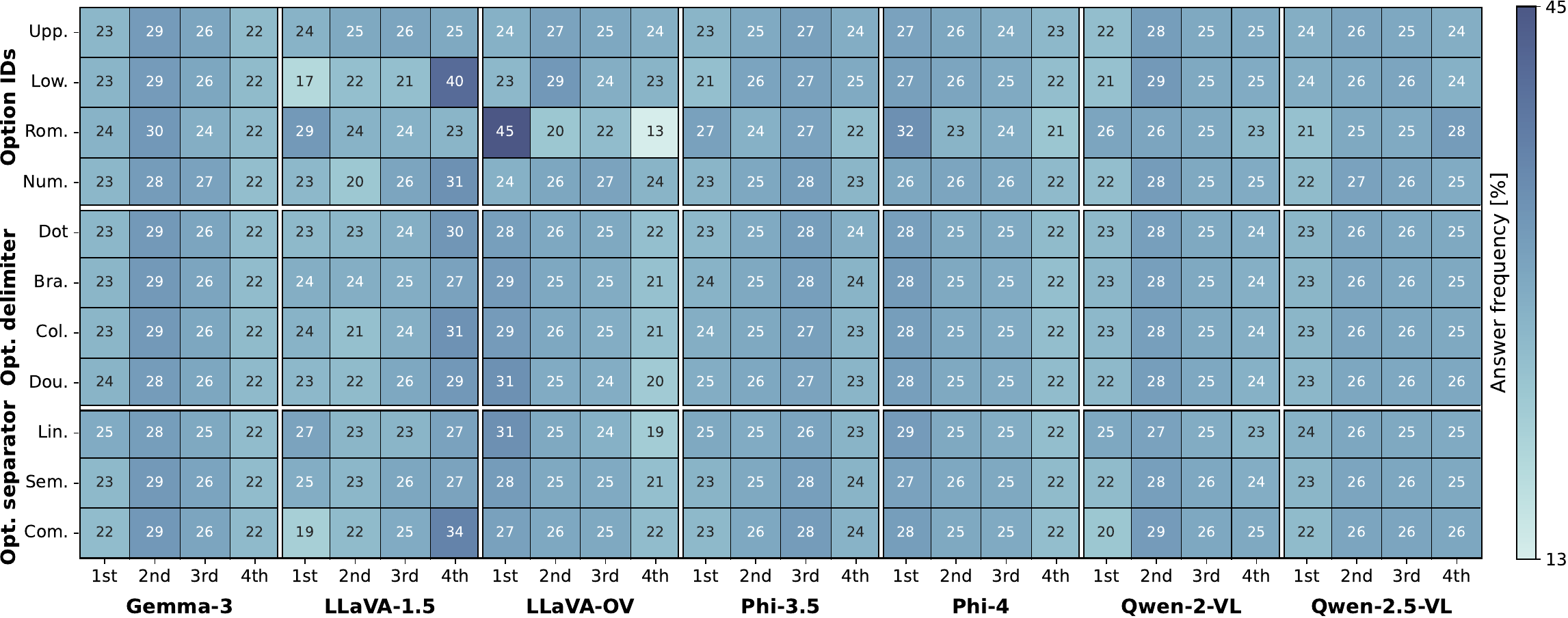}
    \vspace{-8pt}
    \caption{\textbf{Answer behavior of \acp{MLLM} on the MM-Bench dataset.} We report the average answering frequency per option position for all seven \acp{MLLM} across the prompt format variations. The results are averaged over 12 evaluations for \textit{option IDs} and \textit{option delimiter}, and 16 for \textit{option separator}. Option position bias is mitigated via a circular evaluation scheme.}
    \label{fig:answer_freq_mmbench}
\end{figure*}

\begin{figure*}[hbtp]
    \centering
    \includegraphics[width=0.95\textwidth]{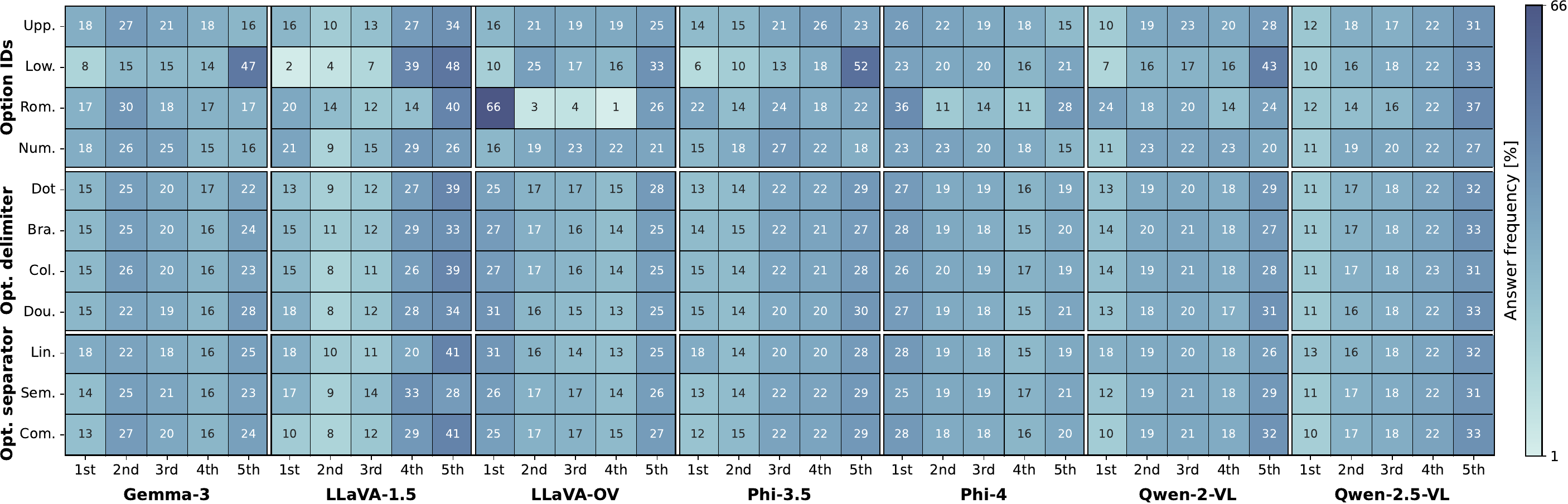}
    \vspace{-8pt}
    \caption{\textbf{Answer behavior of \acp{MLLM} on the MME-RealWorld-Lite dataset.} We report the average answering frequency per option position for all seven \acp{MLLM} across the prompt format variations. The results are averaged over 12 evaluations for \textit{option IDs} and \textit{option delimiter}, and 16 for \textit{option separator}. Option position bias is mitigated via a circular evaluation scheme.}
    \label{fig:answer_freq_mmerwlite}
\end{figure*}

\begin{figure*}[hbtp]
    \centering
    \includegraphics[width=0.95\textwidth]{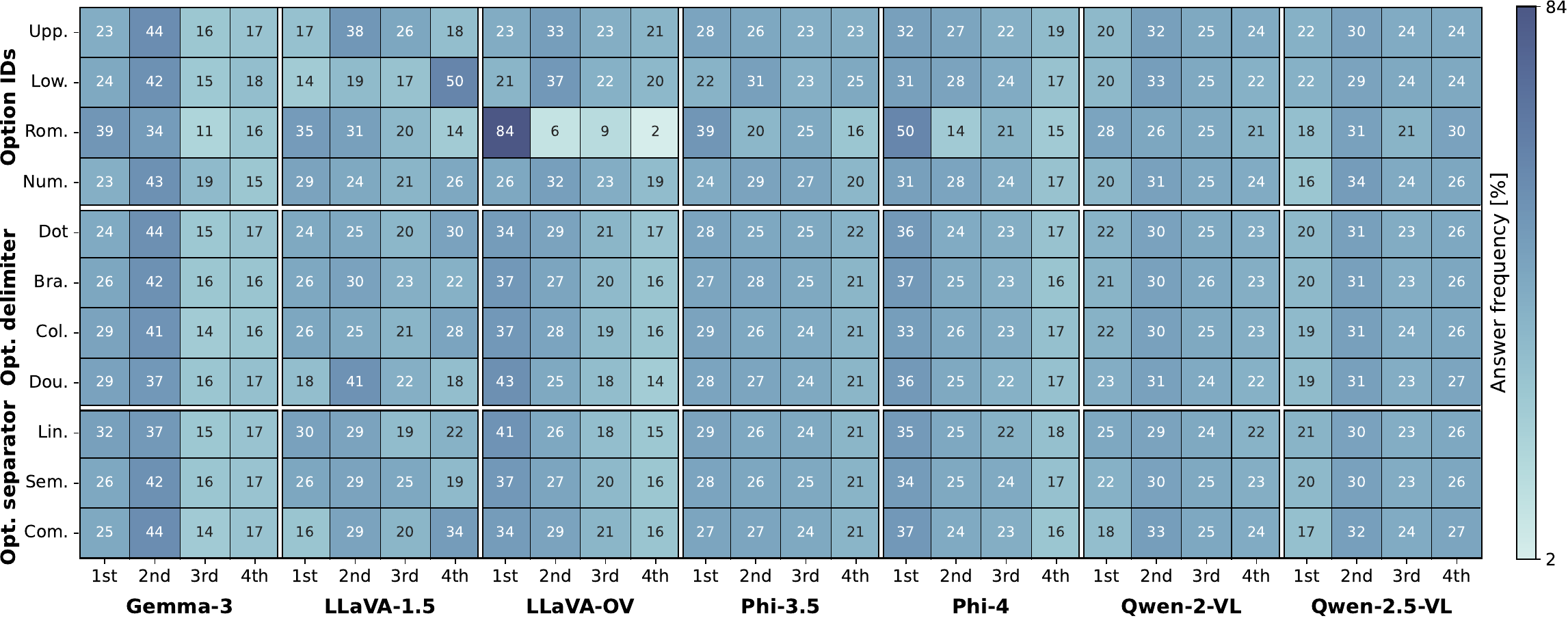}
    \vspace{-8pt}
    \caption{\textbf{Answer behavior of \acp{MLLM} on the V*Bench dataset.} We report the average answering frequency per option position for all seven \acp{MLLM} across the prompt format variations. The results are averaged over 12 evaluations for \textit{option IDs} and \textit{option delimiter}, and 16 for \textit{option separator}. Option position bias is mitigated via a circular evaluation scheme.}
    \label{fig:answer_freq_vstar}
\end{figure*}

We report the answering frequency for each option position across five \ac{MC-VQA} datasets, using a circular evaluation scheme to mitigate option-order bias, as previously noted.
We present the answer frequency results for A-OKVQA in \cref{fig:answer_freq_aokvqa}, HRBench-4K in \cref{fig:answer_freq_hrbench}, MM-Bench in \cref{fig:answer_freq_mmbench}, MME-RealWorld-Lite in \cref{fig:answer_freq_mmerwlite} and V*Bench in \cref{fig:answer_freq_vstar}.
For our analysis, we first count the number of times each option position is selected for each evaluation.
To obtain a meaningful measure of answer frequency, we normalize these counts.
This is necessary because some datasets contain a varying number of options per question.
Therefore, for each option position $i$, we compute its normalized answering frequency by dividing the number of times it was selected ($s_i$) by the number of times it appeared in questions ($n_i$):
\begin{equation}
    f_i = \frac{s_i}{n_i}.
\end{equation}
    
\noindent Next, we normalize these answering frequencies across all option positions so that they sum to $1$:

\begin{equation}
    \hat{f}_i = \frac{f_i}{\sum_{j=1}^{o} f_j},
\end{equation}

\noindent where $o$ is the maximal number of options in the datasets.
This ensures comparability of the answering frequency across datasets and positions.
In general, we observe a uniform answering behavior across all \acp{MLLM} for the datasets A-OKVQA, HRBench-4K, MM-Bench, and V*Bench.
However, under certain evaluation configurations—such as LLaVA-1.5 with lowercase option IDs or LLaVA-OV with Roman numeral option IDs—we detect significant preferences for specific option IDs.
Notably, on MME-RealWorld-Lite, most models display less uniform option-selection behavior compared to other benchmarks.

\subsection{Bias Mitigation Results}
\label{app:bias_mitigation_results}

\begin{figure}
    \centering
    \includegraphics[width=\linewidth]{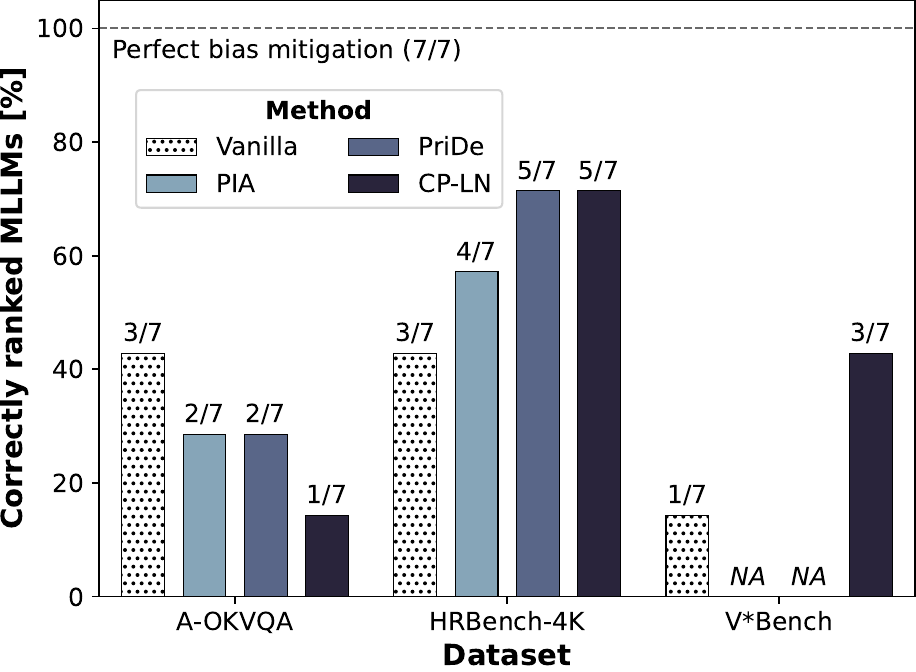}
    \vspace{-18pt}
    \caption{\textbf{Existing bias mitigation methods provide only limited mitigation of the prompt format-induced biases analyzed in this work.} We compare the effectiveness of bias mitigation methods (see \cref{app:bias_mitigation_methods}) in reducing the unexplored biases on three \ac{MC-VQA} datasets. While some methods improve rankings on individual datasets, no method consistently recovers the rankings obtained by averaging performance across all $48$ prompt formats (see \cref{tab:prompt_format_variations}).}
    \label{fig:bias_mitigation_results}
\end{figure}

Next, we examine the extent to which existing bias mitigation methods \ac{PIA}, \ac{PriDe}, and \ac{CP-LN} (see \cref{app:bias_mitigation_methods}) can mitigate the influence of the prompt format-induced biases across the datasets A-OKVQA, HRBench-4K, and V*Bench.
We assess their effectiveness by comparing method rankings to those obtained by averaging accuracy across all 48 prompt formats.
For reference, we also compare rankings based on each dataset's standard prompt format with those from the evaluation results across all prompt formats.
Our findings indicate that existing bias mitigation methods provide only limited robustness against prompt format-induced biases (see \cref{fig:bias_mitigation_results}).
Interestingly, on A-OKVQA, the vanilla prompt format correctly ranks $3$ models, while \ac{PIA} ranks $2$.
\ac{PriDe} performs best on this dataset, correctly ranking $4$ of the $7$ models, whereas \ac{CP-LN} ranks only $1$.
On HRBench-4K, both \ac{PriDe} and \ac{CP-LN} correctly rank $5$ models, outperforming \ac{PIA} ($4$) and the vanilla format ($3$).
This suggests that bias mitigation can improve ranking quality on some datasets, although the gains are not consistent across benchmarks.
\ac{PIA} and \ac{PriDe} are not compatible with evaluations on V*Bench due to their requirement for a fixed number of options (\textit{NA} in \cref{fig:bias_mitigation_results}), which varies between $2$ and $4$ in this dataset.
\ac{CP-LN} correctly ranks $3$, while the vanilla format ranks $1$ of the $7$ \acp{MLLM}.
Overall, none of the evaluated bias mitigation methods consistently reflects benchmarking outcomes under prompt format variations across all datasets. Although \ac{PriDe} and \ac{CP-LN} improve ranking accuracy on individual benchmarks, their effectiveness remains dataset-dependent and they do not reliably recover the reference rankings obtained by averaging over all $48$ prompt formats.

\noindent We present the full bias mitigation results on A-OKVQA in \cref{tab:bias_mitigation_aokvqa}, HRBench-4K in \cref{tab:bias_mitigation_hrbench_4k}, and V*Bench in \cref{tab:bias_mitigation_vstar}.
We assess the effectiveness of three bias mitigation methods by comparing method rankings to those obtained by averaging accuracy across all $48$ prompt format permutations.

\begin{table*}[hbtp]
    \centering
    \caption{\textbf{Full bias mitigation results.}
    This table compares the bias-mitigation effectiveness of the methods \ac{CP-LN}, \ac{PIA}, and \ac{PriDe} on the datasets A-OKVQA, HRBench-4K, and V*Bench.
    V*Bench contains questions with $2$ and $4$ options.
    The methods \ac{PIA} and \ac{PriDe} do not support mixed-number of options datasets.
    We assess effectiveness by counting the number of correctly ranked models relative to the pseudo GT.
    The pseudo GT is obtained by averaging accuracy across all $48$ prompt formats for each dataset (see \cref{tab:prompt_format_variations}).
    For correctly ranked models, the rank is highlighted in \textbf{bold}.}
    \label{tab:bias_mitigation_full_results}

    \newcommand{\subtablewidth}{0.94\linewidth}

    \begin{subtable}[b]{\subtablewidth}
        \centering
        \caption{Bias mitigation results on A-OKVQA.}
        \label{tab:bias_mitigation_aokvqa}
        \resizebox{\subtablewidth}{!}{
            \input{tables/bias_mitigation_results_aokvqa}%
        }
    \end{subtable}

    \vspace{.95em} 

    \begin{subtable}[b]{\subtablewidth}
        \centering
        \caption{Bias mitigation results on HRBench-4K.}
        \label{tab:bias_mitigation_hrbench_4k}
        \resizebox{\subtablewidth}{!}{
            \input{tables/bias_mitigation_results_hrbench_4k}%
        }
    \end{subtable}

        \vspace{.95em} 

    \begin{subtable}[b]{\subtablewidth}
        \centering
        \caption{Bias mitigation results on V*Bench.}
        \label{tab:bias_mitigation_vstar}
        \resizebox{\subtablewidth}{!}{
            \input{tables/bias_mitigation_results_vstar_bench}%
        }
    \end{subtable}

\end{table*}

\subsection{Full Benchmark Results}
\label{app:full_benchmark_results}

Further, we report the full benchmark results of all dataset-model evaluations across \textit{A-OKVQA} in \cref{tab:full_results_aokvqa}, \textit{HRBench-4K} in \cref{tab:full_results_hrbench}, \textit{MM-Bench} in \cref{tab:full_results_mmbench}, \textit{MME-RealWorld-Lite} in \cref{tab:full_results_mme}, and \textit{V*Bench} in \cref{tab:full_results_vstar}.
Each table shows the \ac{MLLM} performance under prompt format variations.
The reported metrics include accuracy (\%) and coverage (\%), where accuracy reflects the correctness of predictions and coverage indicates the instruction-following capability for each model.
These tables provide a comprehensive view of how prompt format variations influence model performance across all used \ac{MC-VQA} datasets.

\input{tables/full_results_aokvqa}

\input{tables/full_results_hr_bench_4k}

\input{tables/full_results_mmbench}

\input{tables/full_results_mme_rw_lite}

\input{tables/full_results_vstar}

\clearpage
\section{Mechanistic Analysis of Tokenization}
\label{app:mechanistic_tokenization}

This section provides detailed evidence for the tokenization and decoding mechanisms driving prompt format sensitivity (\cref{sect:mechanistic_sources}).
We analyze how model-specific tokenizers encode option IDs and how these representations translate into the tokens generated during decoding.
In particular, we examine differences in tokenization length and atomicity at the input level, as well as the number of tokens required to generate option IDs during output.
These analyses reveal how seemingly minor formatting choices alter the token sequences processed by the model and thereby influence multiple-choice prediction behavior.

\subsection{Tokenizer overview}
\label{app:tokenizer_overview}

\cref{tab:overview_tokenizers} provides an overview of the HuggingFace tokenizers used by the evaluated \acp{MLLM}.
These models rely on different tokenizer families that reflect their pretraining corpora and design choices.
In particular, several models use LLaMA- or Qwen-style tokenizers, while Gemma-3 and Phi-4 employ tokenizers solely used by them.
These tokenizers differ in vocabulary construction, subword segmentation strategy, and the handling of special tokens, including image placeholders and control symbols.
We explicitly adopt the original tokenizer of each model to preserve compatibility with pretrained checkpoints and to avoid introducing confounding effects in downstream evaluation.

\input{tables/tokenizer_overview}

\subsection{Tokenization of Option ID-Delimiter Pairs}
\label{app:tokenization_input}

\begin{figure}[t]
    \centering
    \includegraphics[width=0.95\linewidth]{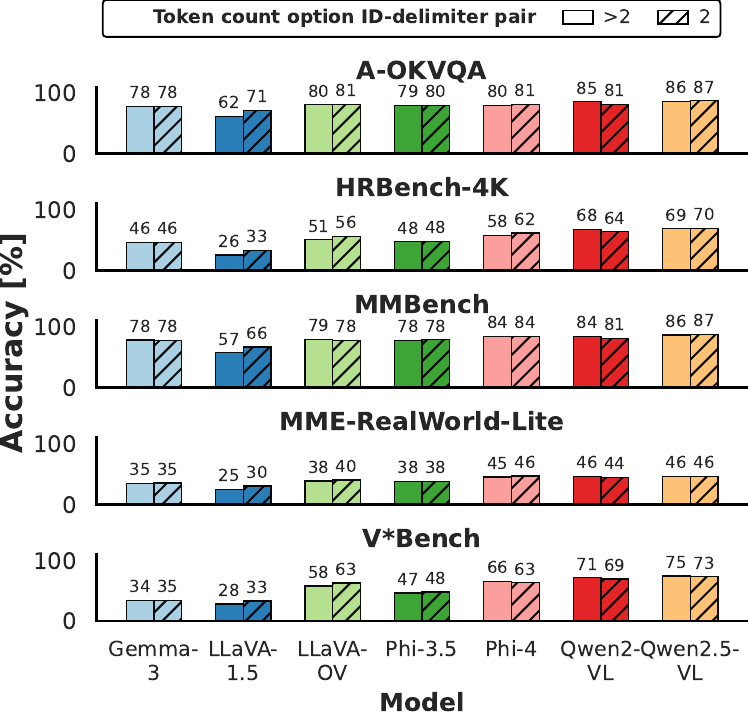}
    \vspace{-8pt}
    \caption{\textbf{Effect of option ID-delimiter tokenization length on accuracy.}
    Each pair of bars compares prompt formats where the option ID-delimiter pair tokenizes into exactly two tokens versus more than two tokens.
    Accuracy differences across these groups highlight how increases in tokenization length alter the input representation.
    Across several models and datasets, we observe only slight deviations across the two groups.}
    \label{fig:tokenization_length}
    \vspace{-15pt}
\end{figure}

In this section, we provide in-depth information about the input tokenization behavior of the four different tokenizers used by the \acp{MLLM} covered in this paper (see \cref{tab:overview_tokenizers}).
This analysis consists of two parts: (1) the number of tokens of option ID-delimiter pairs and (2) whether the option ID token(s) are atomically tokenized, i.e., they are contained as the bare and not as corrupted tokens.
\cref{tab:tokenization_encoding} visualizes the tokenization results of both analyses and we highlight token counts of option ID-delimiter pairs higher than 2 in bold and cases where for at least one option ID is not atomically tokenized as grey cells.

\noindent Across tokenizers, simple delimiters such as dots, colons, and single brackets generally lead to stable encodings with an average of two tokens, corresponding to a single token for the option ID and a single token for the delimiter.
In contrast, double-bracket delimiters or specific option ID sets frequently increase token counts beyond two and more often break atomic encoding of option IDs, particularly for numeric and Roman numeral formats.
These effects are tokenizer-specific and reflect differences in subword vocabularies and special character handling.

\input{tables/tokenization_prompt_formats}

\noindent To quantify how these encoding differences affect model performance, \cref{fig:tokenization_length} compares accuracy across prompt formats grouped by tokenization length.
Each pair of bars contrasts formats where the option ID-delimiter pair tokenizes into exactly two tokens versus more than two tokens.

\noindent We find that longer tokenizations often correlate with reduced accuracy, particularly for models such as LLaVA-1.5 and Phi-3.5.
This result shows that increasing tokenization length alters the input representation in a way that negatively affects downstream answer selection, even though the task semantics remain unchanged.
However, this effect is not uniform across all models, indicating that some architectures are more robust to non-minimal token encodings.
Also, this effect has less pronounced influences than the atomic option ID tokenization (see \cref{subsect:tokenization_effects}).

\subsection{Decoding of Option IDs}
\label{app:decoding_tokenization}

\input{tables/decoding_option_IDs}

Next, we analyze how option IDs are tokenized during output generation, focusing on the number of tokens required to generate each ID.
This factor directly affects decoding behavior, as multiple-choice answering often relies on reproducing option IDs that appear in the prompt.

\noindent \cref{tab:decoding_token_length_option_ids} reports the tokenization length for different option ID sets across all evaluated models.
Most models generate letter-based and Roman numeral IDs using a single token.
However, some models, such as LLaVA-1.5 and Phi-3.5, require multiple tokens for numeric option IDs.

\noindent This increase in tokenization length changes the generation process: instead of copying a single token, the model must generate a short sequence of tokens.
This can reduce the reliability of copy-style decoding and increase the likelihood of errors.
These effects are consistent with the instability observed in \cref{sect:benchmark_instability} and with the copy-attention behavior analyzed in \cref{subsect:copy_attention_heads}.

\noindent Overall, decoding tokenization can introduce an additional source of variability, as prompt format choices that affect output token length can influence the ease of option selection even when input representations remain unchanged.

\clearpage
\section{Open-Ended Evaluation}
\label{app:open_ended_eval}

\begin{figure}[b]
    \centering
    \begin{VQABox}{Prompt Layout: open-ended evaluation}
    \texttt{You are given an image and a question about that image.\\
    \\
    \textbf{Question:}\\
    What is the color of the "\texttt{<object>}"?\\
    \\
    Provide an answer based solely on the image.}
    \end{VQABox}
    \caption{\textbf{Prompt template for open-ended VQA evaluation.} The model answers freely without being provided with answer options or additional contextual information, relying only on the image and the question.}
    \label{fig:oe_evaluation}
\end{figure}

\noindent To evaluate the alignment between multiple-choice and open-ended (i.e., free-form) evaluation schemes, we assess all \acp{MLLM} on A-OKVQA, HRBench-4K, and V*Bench using an open-ended VQA protocol. This analysis is motivated by the fact that many benchmarks are reported in multiple-choice form, while real-world deployment typically requires free-form generation.
Therefore, it is essential for interpreting benchmark results to understand whether model behavior remains consistent across evaluation settings.
In the open-ended setting, models are provided only with the image and the question.
Thus, they are required to generate an answer without being exposed to candidate options.
This removes inductive biases introduced by option ordering, tokenization effects, or information leakage from the answer options themselves that can arise in multiple-choice evaluation.
We use a fixed prompt template (see \cref{fig:oe_evaluation}) across all models to ensure comparability.
For all experiments, we set the maximum number of generated tokens to $256$ and disable sampling to obtain deterministic outputs.
We perform evaluation by comparing generated answers against open-ended ground-truth annotations as described below.

\subsection{Conversion to open-ended answers}
\label{app:conversion_open_ended}

\ac{MC-VQA} datasets do not directly provide free-form ground-truth answers suitable for open-ended evaluation.
To enable a consistent comparison across datasets and evaluation schemes, we construct open-ended ground-truth answers for A-OKVQA, HRBench-4K, and V*Bench.

\noindent For V*Bench, we use the original open-ended annotations provided by the dataset.
For A-OKVQA and HRBench-4K, we convert the correct multiple-choice option into a natural-language answer.
This conversion is performed using \textit{Qwen2.5‑32B‑Instruct}, conditioned on the question and the correct answer option.
The conversion prompt enforces strict constraints to prevent the introduction of new information and to remove any reference to the original multiple-choice structure (see \cref{fig:conversion_mcq_oe}).
All generated answers are manually verified to ensure semantic equivalence with the original correct option and to correct occasional formatting or phrasing artifacts.
This procedure yields a set of open-ended ground-truth answers that are faithful to the original annotations while being suitable for free-form evaluation.

\begin{figure*}[b]
    \centering
    \begin{VQABox}{Prompt Layout: Conversion of multiple-choice VQA to open-ended evaluation setting}
    \texttt{\textbf{System Prompt (Answer Generator):}\\
    You are a ground-truth answer generator.
    Your task is to convert a correct multiple-choice answer option into
    a clear, natural, open-ended answer.
    Always write a full, self-contained sentence.
    Never refer to option IDs or multiple-choice structure.\\
    \\
    \textbf{Rules:}\\
    Use only the information contained in the provided correct option.
    Do not introduce additional explanations or external knowledge.
    Write the answer as an authoritative reference suitable as ground truth.\\
    \\
    \textbf{User Prompt:}\\
    \\
    \textbf{[QUESTION]}\\
    What is the color of the object?\\
    \\
    \textbf{[CORRECT ANSWER OPTION]}\\
    "\textit{<correct option text>}"\\
    \\
    \textbf{[INSTRUCTION]}\\
    Write the ground-truth open-ended answer.
    Return only the answer text.}
    \end{VQABox}
    \caption{\textbf{Prompt template for generating open-ended ground-truth answers from multiple-choice VQA annotations.} A language model converts the correct answer option into a natural, self-contained sentence without referencing the original multiple-choice format.}
    \label{fig:conversion_mcq_oe}
\end{figure*}

\subsection{Correctness assessment}
\label{app:correctness_assessment}

In the next step, we must assess the correctness of the free-form answer given by each \ac{MLLM} in the open-ended evaluation scheme.
Therefore, we use an LLM-as-a-judge approach with Qwen2.5-32B-Instruct as the LLM-judge.
We define a prompt template (see \cref{fig:llm-judge-prompt}), in which we provide the \ac{LLM} detailed instructions, in-context examples, as well as the given free-form answer and the ground-truth open-ended answer.
This setup constrains the evaluation model to assess only the semantic correctness of the generated answer relative to the ground-truth reference, ensuring consistent and objective judgments without introducing additional interpretation bias.
We provide the open-ended results in \cref{tab:open_ended_evaluation}.

\begin{table}[hbtp]
    \centering
    \caption{\textbf{Open-ended evaluation.}
    We compare standard \ac{MC-VQA} with open-ended generation (difference reported in the brackets).
    Open-ended accuracy is consistently lower, and \ac{MC-VQA} does not reliably correlate with open-ended rankings.}
    \label{tab:open_ended_evaluation}
    \resizebox{\linewidth}{!}{%
    \begin{tabular}{l c c c}
        \toprule
        & \multicolumn{3}{c}{\textbf{Open-ended accuracy $[\%]$}}\\
        \textbf{Model} & A-OKVQA & HRBench-4K & V*Bench\\
        \midrule
        Gemma-3 & $32.01$ $(\signed{-45.98})$ & $30.75$ $(\signed{-15.16})$ & $31.15$ $(\signed{-3.26})$\\
        LLaVA-1.5 & $33.23$ $(\signed{-33.85})$ & $20.00$ $(\signed{-9.90})$ & $39.00$ $(\signed{+7.85})$\\
        LLaVA-OV & $33.49$ $(\signed{-47.79})$ & $13.25$ $(\signed{-40.61})$ & $61.52$ $(\signed{-0.84})$\\
        Phi-3.5 & $34.45$ $(\signed{-44.93})$ & $30.25$ $(\signed{-17.65})$ & $46.86$ $(\signed{-0.60})$ \\
        Phi-4 & $51.15$ $(\signed{-29.36})$ & $42.00$ $(\signed{-19.13})$ & $63.09$ $(\signed{-6.74})$ \\
        Qwen2-VL & $41.97$ $(\signed{-39.99})$ & $50.75$ $(\signed{-13.61})$ & $65.97$ $(\signed{-3.61})$ \\
        Qwen2.5-VL & $30.26$ $(\signed{-56.51})$ & $46.00$ $(\signed{-23.61})$ & $65.45$ $(\signed{-11.88})$ \\
        \bottomrule
    \end{tabular}%
    }
\end{table}

\begin{figure*}[hbtp]
    \centering
    \begin{VQABox}{Prompt Layout: Correctness assessment of open-ended answer via LLM-as-a-judge}
    \texttt{\textbf{System Prompt (Judge Model):}\\
    You are a strict and consistent evaluator for open-ended Visual Question Answering (VQA).
    Your task is to assess the correctness of a model-generated answer by comparing it to
    a ground-truth reference for the same question.
    Judge correctness, not style.
    Accept semantically equivalent answers.
    Never reveal chain-of-thought; provide only a brief rationale.\\
    \\
    \textbf{User Prompt:}\\
    Evaluate the VQA response.\\
    \\
    \textbf{[Abstract Few-Shot Examples]}\\
    Several example evaluations are provided to illustrate valid judgments,
    including cases of correct, partially correct, and incorrect answers.
    Each example specifies a question, a ground-truth answer, a model-generated answer,
    and the expected JSON verdict.\\
    \\
    \textbf{[Evaluation Instance]}\\
    \textbf{QUESTION:}\\
    What is the color of the object?\\
    \\
    \textbf{GROUND\_TRUTH\_ANSWER:}\\
    "\textit{<reference answer>}"\\
    \\
    \textbf{GIVEN\_ANSWER (model output):}\\
    "\textit{<open-ended prediction>}"\\
    \\
    Return a valid JSON object containing:
    a correctness verdict, a numeric score, error categories,
    and a brief explanation. Output JSON only.}
    \end{VQABox}
    \caption{\textbf{Prompt template for LLM-as-a-judge evaluation of open-ended VQA.} A separate language model evaluates the correctness of a model-generated answer relative to a ground-truth reference, guided by abstract few-shot examples that define the evaluation criteria.}
    \label{fig:llm-judge-prompt}
\end{figure*}

\subsection{Grounded MC-VQA evaluation}
\label{app:grounded_mc_vqa}

We next examine whether incorporating reasoning signals can stabilize model behavior under prompt format variations.
Specifically, we test whether conditioning multiple-choice predictions on a generated open-ended answer improves robustness.

\noindent To this end, we construct an extended multiple-choice prompt that includes the open-ended response of the model as additional context (see \cref{fig:grounded-prompt}).
This grounded setting separates two components of the task: (i) generating a candidate answer in open-ended form, and (ii) selecting the corresponding option ID in the multiple-choice format.
By providing the open-ended answer explicitly, the model no longer needs to rely solely on the option representation to recover its reasoning, which allows us to isolate the effect of option selection.

\noindent This setup constrains the model to verify or correct its prior answer based on the image and the given options, rather than generating a decision solely from the prompt structure.
As a result, any remaining performance differences across prompt formats reflect variability in the option-selection process rather than differences in underlying reasoning.

\begin{figure*}[b]
    \centering
    \begin{VQABox}{Prompt Layout: Grounded multiple-choice VQA evaluation}
    \texttt{You are given an image and a question about that image. {\color{blue}A previously generated open-ended answer is provided as additional context.}\\
    \\
    \textbf{Question:}\\
    What is the color of the "<object>"?\\
    \\
    {\color{blue}
    \textbf{Contextual Description (previous open-ended answer):}\\
    "\textit{<model-generated answer>}"\\
    }
    \\
    \textbf{Options:}\\
    a.\ "<option\_1>"\\
    b.\ "<option\_2>"\\
    c.\ "<option\_3>"\\
    d.\ "<option\_4>"\\
    \\
    Select the best answer to the above multiple-choice question based on the image. Respond with only the letter (e.g., a, b, c or d) of the correct option and no bracket, colon, or dot.}
    \end{VQABox}
    \caption{\textbf{Example prompt for grounded evaluation}. The model is conditioned on its own prior open-ended answer (context), which must be verified or corrected using visual evidence from the image.}
    \label{fig:grounded-prompt}
\end{figure*}

\noindent The results of the grounded \ac{MC-VQA} evaluation indicate that the context does not stabilize \ac{MC-VQA} across prompt formats (see \cref{tab:full_results_grounded_mc_vqa}).

\input{tables/results_grounded_mc_vqa}

\clearpage
\section{Additional Analyses}
\label{app:additional_analyses}

In this section, we provide detailed specifications and results for three analyses. Specifically, we report results for training exposure (\cref{subsect:training_exposure_option_IDs}, \cref{app:training_exposure}), copy attention heads (\cref{subsect:copy_attention_heads}, \cref{app:copy_attention_heads}), and models’ ability to infer effective prompt factors (\cref{subsec:models_infer_good_prompt_factors}, \cref{app:models_infer_good_prompt_factors}).

\subsection{Effects Go Beyond Training Exposure}
\label{app:training_exposure}

In \cref{subsect:training_exposure_option_IDs}, we examine whether the performance of option ID sets in \ac{MC-VQA} correlates with their exposure during model training.
To this end, we introduce additional option ID sets that are unlikely to have appeared in multiple-choice contexts (text-only or vision-language) during pretraining or instruction tuning of the evaluated \acp{MLLM}.

\noindent These option ID sets differ in their degree of semantic leakage and structural properties (see \cref{tab:additional_option_ids}).
Specifically, Greek-letter IDs exhibit minimal semantic leakage and a clear ordinal structure.
Directional words introduce moderate semantic leakage and relational structure but lack a strict ordering.
Finally, word-based IDs have high semantic leakage and no inherent ordinal structure.

\noindent All IDs tokenize into a single token for Phi‑4 and Qwen2.5‑VL, which controls for decoding length effects and isolates the influence of ID semantics and structure on multiple-choice prediction.

\begin{table}[hbtp]
    \centering
    \caption{\textbf{Additional option ID sets grouped by semantic leakage and structural properties.}
    We categorize option ID sets by their degree of semantic leakage and inherent structure to disentangle training exposure effects from tokenization and decoding artifacts.}
    \label{tab:additional_option_ids}
    \resizebox{\linewidth}{!}{%
    \begin{tabular}{p{0.3\linewidth} p{0.55\linewidth} p{0.30\linewidth}}
        \toprule   
        \textbf{Option ID set} & \textbf{Semantic and structural properties} & \textbf{Example IDs} \\
        \midrule
        Greek letters 
        & Minimal semantic leakage; strong ordinal structure. 
        & {\ttfamily alpha/\allowbreak beta/\allowbreak gamma/\allowbreak delta} \\
        Directional words 
        & Moderate semantic leakage; relational semantics without strict ordinal structure. 
        & {\ttfamily north/\allowbreak east/\allowbreak south/\allowbreak west} \\
        Word-based IDs 
        & High semantic leakage; no inherent ordinal or relational structure. 
        & {\ttfamily key/\allowbreak desk/\allowbreak tree/\allowbreak sea} \\
        \bottomrule
    \end{tabular}%
    }
\end{table}

\begin{table}[t]
    \centering
    \caption{\textbf{Performance of additional option ID sets on V*Bench.}
    We report mean, standard deviation, minimum, and maximum accuracy across $12$ prompt format variants for alternative option ID sets with varying semantic leakage.
    Greek-letter IDs achieve accuracy comparable to standard letter- and number-based IDs, while direction- and word-based IDs show substantially lower performance and higher variance.}
    \label{tab:results_additional_option_ids}
    \resizebox{\linewidth}{!}{
    \begin{tabular}{l l | c c c c}
        \toprule
        & & \multicolumn{4}{c}{\textbf{Accuracy} $[\%]\uparrow$} \\
        \textbf{Model} & \textbf{Option IDs} & Mean & Std. & Min. & Max. \\
        \midrule

        \multirow{7}{*}{\textbf{Phi-4}} & Greek & $70.90$ & $1.72$ & $68.13$ & $73.15$\\
        & Direct. & $49.18$ & $11.67$ & $12.42$ & $54.70$ \\
        & Words & $44.97$ & $12.88$ & $29.20$ & $68.96$ \\
        & \textcolor{darkgray}{Uppercase} & \basevqacell{73.21} & \basevqacell{0.63} & \basevqacell{72.32} & \basevqacell{74.33} \\
        & \textcolor{darkgray}{Lowercase} & \basevqacell{72.58} & \basevqacell{0.88} & \basevqacell{71.48} & \basevqacell{74.16} \\
        & \textcolor{darkgray}{Numbers} & \basevqacell{71.55} & \basevqacell{1.01} & \basevqacell{69.97} & \basevqacell{73.15} \\
        & \textcolor{darkgray}{Roman} & \basevqacell{62.00} & \basevqacell{4.36} & \basevqacell{57.05} & \basevqacell{70.97} \\
        \midrule

        \multirow{7}{*}{\textbf{Qwen2.5-VL}} & Greek & $76.69$ & $1.33$ & $74.83$ & $78.69$ \\
        & Direct. & $65.27$ & $4.14$ & $57.21$ & $72.65$ \\
        & Words & $60.75$ & $9.44$ & $44.13$ & $71.81$ \\
        & \textcolor{darkgray}{Uppercase} & \basevqacell{79.85} & \basevqacell{1.48} & \basevqacell{77.18} & \basevqacell{81.71} \\
        & \textcolor{darkgray}{Lowercase} & \basevqacell{80.10} & \basevqacell{1.24} & \basevqacell{78.02} & \basevqacell{81.38} \\
        & \textcolor{darkgray}{Numbers} & \basevqacell{75.45} & \basevqacell{0.79} & \basevqacell{73.99} & \basevqacell{77.18} \\
        & \textcolor{darkgray}{Roman} & \basevqacell{74.34} & \basevqacell{0.85} & \basevqacell{72.65} & \basevqacell{75.67} \\
        \bottomrule
    \end{tabular}%
    }
\end{table}

\noindent In \cref{fig:option_id_sets_accuracy}, we visualize the accuracy distribution of the additional option ID sets while varying the option delimiter and separator, resulting in $12$ prompt formats per ID set.
We evaluate this setting on V*Bench using Phi-4 and Qwen2.5-VL, as these are the only models for which all evaluated option IDs tokenize into a single token.

\noindent \cref{tab:results_additional_option_ids} summarizes the mean, standard deviation, and range of accuracy across the $12$ prompt formats for each ID set.
Greek-letter IDs achieve performance that is close to standard uppercase, lowercase, and numeric IDs for both models, with relatively low variance across prompt formats.
This result indicates that effective multiple-choice selection does not depend on exposure to conventional letter-based IDs during training.

\noindent In contrast, directional words and word-based IDs lead to substantially lower average accuracy and markedly higher variance.
These IDs introduce semantic content that is unrelated to the multiple-choice structure, which appears to interfere with reliable option selection.
Overall, these results show that option ID performance cannot be explained by training exposure alone: IDs with minimal semantic leakage and clear structure (such as Greek letters) generalize well, while semantically loaded IDs degrade performance even when tokenization effects are controlled.

\subsection{Copy Attention Heads}
\label{app:copy_attention_heads}

In \cref{subsect:copy_attention_heads}, we show that the attention in specific attention heads, which we call "copy" attention heads, varies across different option ID sets in Qwen2.5-VL.
In this subsection, we will give more detailed results and more in-depth information about the procedure of how we obtained these values for our analysis.
Our analysis scope covers the four option ID sets "Uppercase", "Lowercase", "Numbers", and "Roman", while we have option delimiter fixed as "colon" and option separator as "line break".

\begin{algorithm}[hbtp]
    \caption{\textbf{Extraction of option ID attention margin}}
    \label{alg:attention_extraction}
    \begin{algorithmic}[1]
        \Require Prompt $x$, image $I$, options $\mathcal{O}$, model $f_\theta$
        
        \State Construct conversation input $(x, I)$
        \State Run forward pass $\to$ obtain attention $\{A_{l,h}\}$
        \State Identify query position $q$
        
        \Statex
        \State \textbf{Step 1: Extract option ID token positions}
        \ForAll{$o \in \mathcal{O}$}
            \State $\mathcal{T}_o \gets$ token positions corresponding to $o$
        \EndFor
        
        \Statex
        \State \textbf{Step 2: Compute attention scores}
        
        \For{$l = 1, \dots, L$}
            \For{$h = 1, \dots, H$}
        
                \State Compute option scores:
                \ForAll{$o \in \mathcal{O}$}
                    \State $a_{l,h}(o) \gets \textsc{ExtractAttn}(q, \mathcal{T}_o)$
                \EndFor
        
                \State $s_{l,h} \gets \textsc{Margin}\big(a_{l,h}(o)\big)$
        
            \EndFor
        \EndFor
        
        \State \Return $S \in \mathbb{R}^{L \times H}$
    
    \end{algorithmic}
\end{algorithm}

\noindent \cref{alg:attention_extraction} summarizes the procedure used to extract option-identifier attention signals for a single V*Bench sample.
For each sample, we construct the model-specific multimodal prompt, including the image and multiple-choice options, and run a forward pass through the model to obtain attention maps for all layers and heads.

\noindent We identify the query position $q$ corresponding to the generated option identifier token and determine the token positions $\mathcal{T}_o$ associated with each option identifier $o \in \mathcal{O}$ in the input sequence.
For each attention layer $l$ and head $h$, we compute the average attention mass assigned to the tokens of each option identifier:
\[
\textsc{ExtractAttn}(q, \mathcal{T}_o)
= \frac{1}{|\mathcal{T}_o|} \sum_{t \in \mathcal{T}_o} A_{l,h}(q,t),
\]
where $A_{l,h}(q,t)$ denotes the attention weight from the query position $q$ to token position $t$.

\noindent To quantify how decisively the model attends to a single option identifier, we compute an attention margin for each layer-head pair as the difference between the highest and second-highest option attention scores:
\[
\textsc{Margin}\big(a_{l,h}(o)\big)
=
\max_{o \in \mathcal{O}} a_{l,h}(o)
-
\max_{o \in \mathcal{O'}} a_{l,h}(o)
\]
where $o^*$ denotes the option with the highest attention score and $\mathcal{O'} = \mathcal{O} \setminus \{o^*\}$ represents the option set without the option with the highest attention score.
The resulting margin tensor $S \in \mathbb{R}^{L \times H}$ provides a layer- and head-wise measure of how sharply attention is concentrated on a single option identifier during decoding.

\begin{figure*}[t]
    \centering
    \begin{subfigure}[t]{0.48\textwidth}
        \centering
        \includegraphics[width=\linewidth]{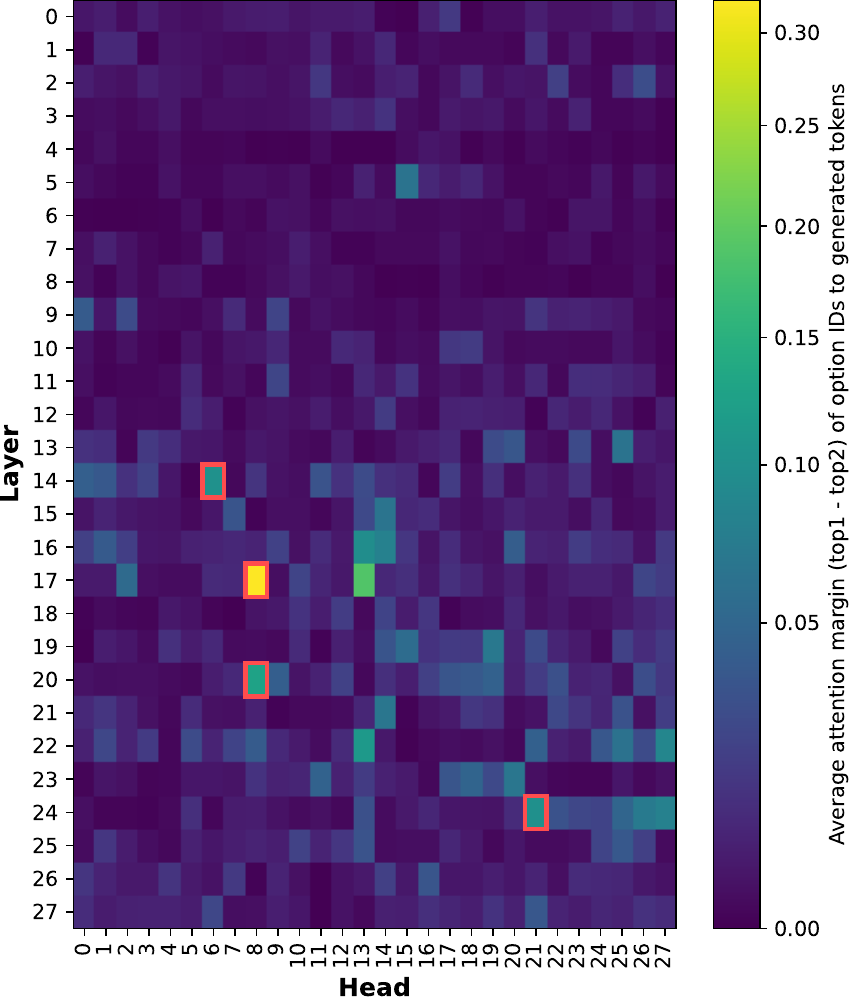}
        \caption{Uppercase letter option IDs}
        \label{fig:sub1}
    \end{subfigure}
    \hfill
    \begin{subfigure}[t]{0.48\textwidth}
        \centering
        \includegraphics[width=\linewidth]{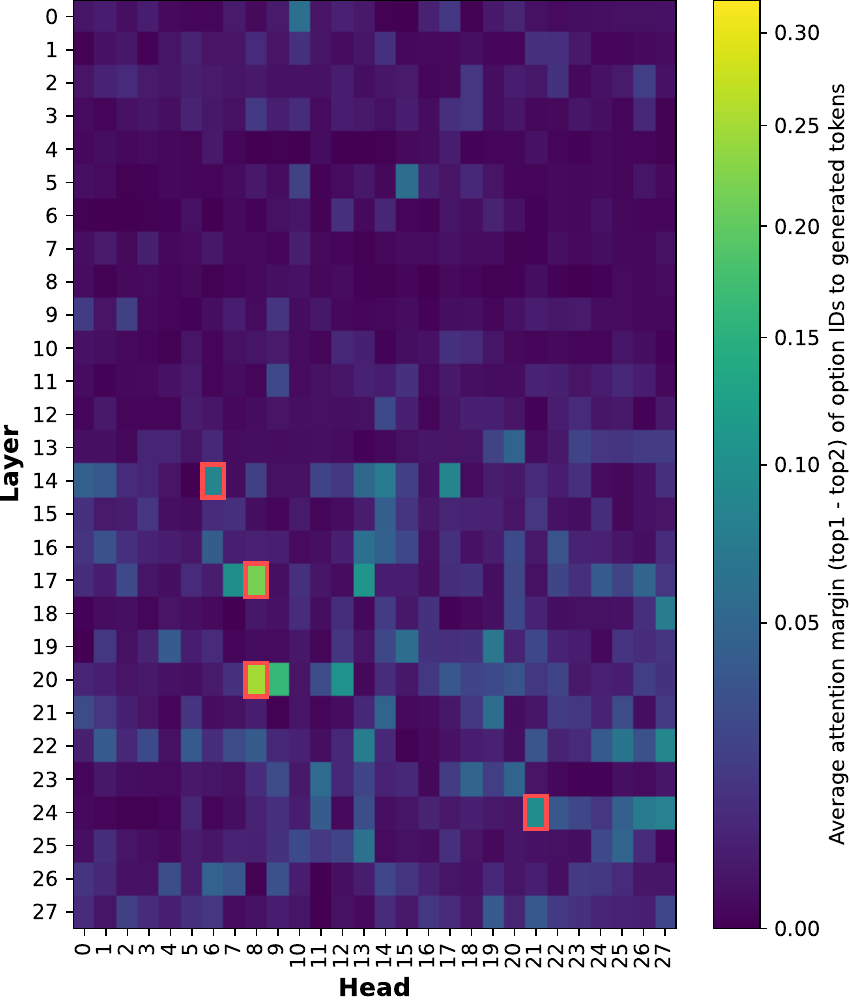}
        \caption{Lowercase letter option IDs}
        \label{fig:sub2}
    \end{subfigure}

    \vspace{0.5em}

    \begin{subfigure}[t]{0.48\textwidth}
        \centering
        \includegraphics[width=\linewidth]{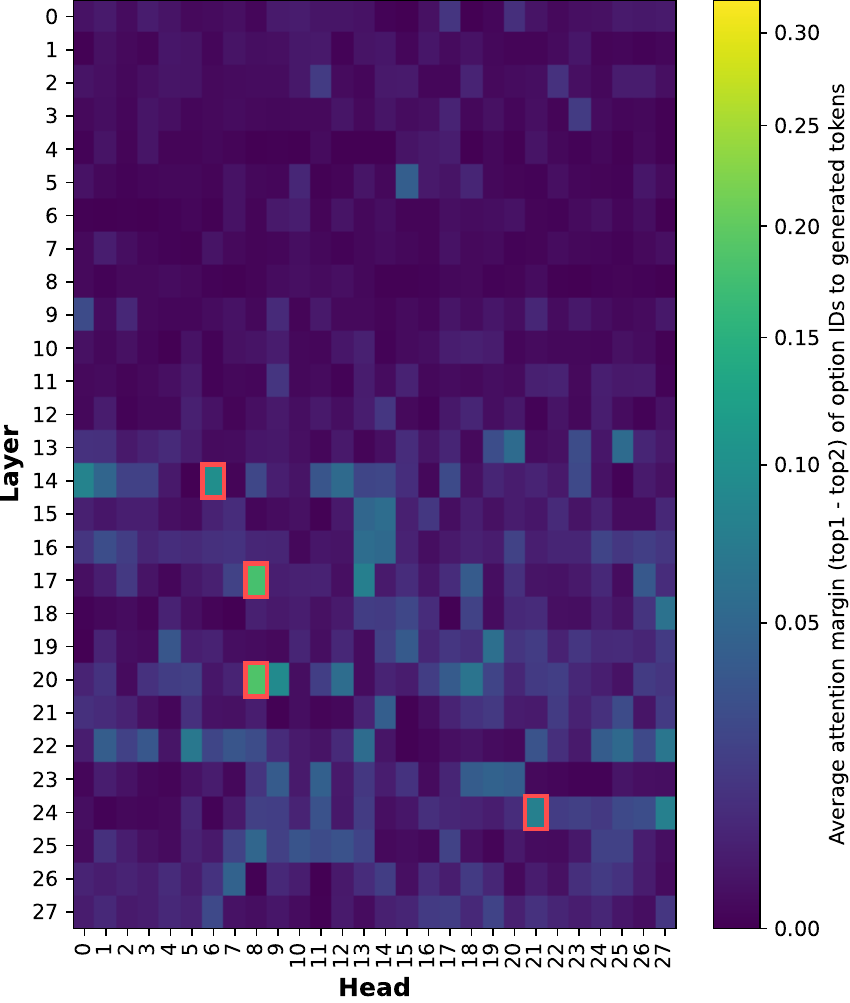}
        \caption{Number option IDs}
        \label{fig:sub3}
    \end{subfigure}
    \hfill
    \begin{subfigure}[t]{0.48\textwidth}
        \centering
        \includegraphics[width=\linewidth]{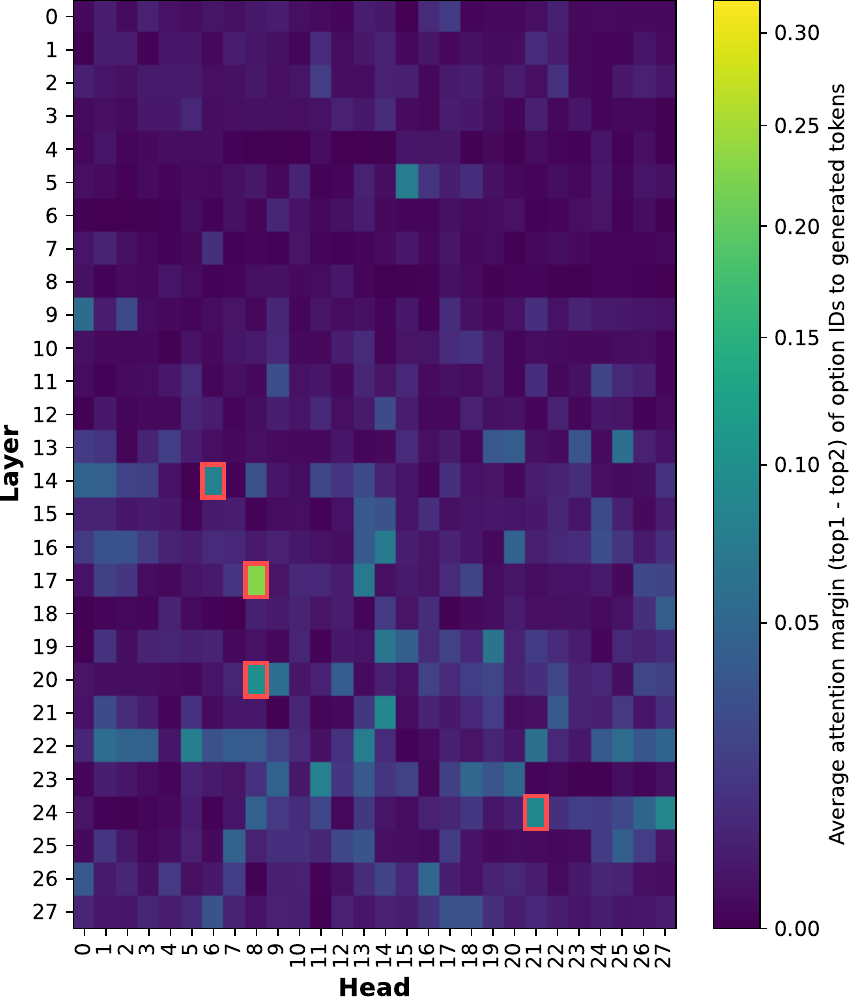}
        \caption{Roman numeral option IDs}
        \label{fig:sub4}
    \end{subfigure}

    \caption{\textbf{Average option ID attention per attention head in Qwen2.5-VL.} We show the average attention margin between top-1 and top-2 attention values across all option ID tokens. We vary the option ID set to uppercase letters \textbf{(a)}, lowercase letters \textbf{(b)}, numbers \textbf{(c)}, and Roman numerals \textbf{(d)}, while keeping the option delimiter (colon) and option separator (line separator) fixed. We highlight the selected "copy" attention heads in \textcolor{highlightcolor}{\textbf{red}}.}
    \label{fig:attention_heads_overview}
\end{figure*}

\noindent We next identify attention heads that exhibit copy-style behavior in \ac{MC-VQA}.
For each evaluation with a fixed option ID set, we average the margin tensor $S$ across all dataset samples.
We then select the top-10 attention heads with the largest average margins for each option ID set and analyze the overlap of these sets across evaluations.
Attention heads that consistently appear among the top-10 across all option ID sets are identified as copy attention heads.
Using this criterion, we identify four such heads at layer-head indices $(14, 6)$, $(17, 8)$, $(20, 8)$, and $(24, 21)$.
\cref{fig:attention_heads_overview} reports the average attention margin across all attention heads for each option ID set, with the identified copy attention heads highlighted.
In \cref{tab:copy_attention_margin}, we report the cumulative attention margin across all four copy attention heads.

\subsection{MLLMs Can Reason About Favorable Prompt Formats For Multiple-Choice}
\label{app:models_infer_good_prompt_factors}

\begin{table}[b]
    \centering
    \caption{\textbf{Model-selected prompt formats for additional ID sets.}
    Each model selects a preferred option ID set, delimiter, and separator from the design space in \cref{tab:prompt_format_variations}.
    Most models choose configurations with strong ranks on V*Bench, which suggests that prompt format preferences are systematic and partly recoverable from the models themselves.
    Lower average rank indicates better performance.}
    \vspace{-7pt}
    \label{tab:model_selected_prompt_formats_2}
    \resizebox{\linewidth}{!}{
    \begin{tabular}{l c c c c c}
        \toprule
        \textbf{Model} & \textbf{Option IDs} & \textbf{Delimiter} & \textbf{Separator} & \textbf{\makecell{Avg.\\ Rank}} & \textbf{\makecell{Rel. gap\\to optimal}}\\
        \midrule
        Phi-4      & Greek   & Colon     & Comma     & \colornum{13.0} & $-6.88$ \\
        Qwen2.5-VL & Greek   & Colon     & Linebreak & \colornum{4.0} & $-1.28$  \\
        \bottomrule
    \end{tabular}%
    }
\end{table}

In \cref{subsec:models_infer_good_prompt_factors}, we analyze whether \acp{MLLM} can anticipate which prompt formats are more favorable for their own multiple-choice performance.
To this end, we present each model with a prompt that contains the explicit design space of semantically identical \ac{MC-VQA} prompt formats options (see \cref{fig:model_prompt_self_selection}).
Without access to any experimental results, the model is asked to predict which combination of formatting choices is most likely to yield the highest accuracy based solely on its internal processing characteristics.
As shown in \cref{tab:model_selected_prompt_formats}, all \acp{MLLM} except Phi‑4 select similar or identical prompt formats, and for each of these models the selected format performs significantly better than random guessing.
We observe similar results for the additional option ID sets (see \cref{tab:additional_option_ids}), where Phi-4 and Qwen2.5-VL both select best-performing Greek option IDs (see \cref{tab:model_selected_prompt_formats_2}).

\begin{figure*}[hbtp]
    \centering
    \begin{VQABox}{Prompt Layout: Model self-selection of optimal MC-VQA prompt format}
    \texttt{\textbf{System Prompt:}\\
    You are a multimodal large language model evaluated on multiple-choice Visual Question Answering (VQA).
    Your task is to reason about how different prompt formatting choices affect your own performance.\\
    \\
    All prompt formats are semantically identical and differ only in superficial formatting choices.
    These differences do not change the meaning of the question or the answer options, but they may influence your accuracy.\\
    \\
    \textbf{Design Space:}\\
    The prompt formats vary along three independent factors.\\
    \\
    \textbf{Option ID set (answer labels):}\\
    - Uppercase letters: A / B / C / D\\
    - Lowercase letters: a / b / c / d\\
    - Numbers: 1 / 2 / 3 / 4\\
    - Roman numerals: I / II / III / IV\\
    \\
    \textbf{Option delimiter (ID--option separator):}\\
    - Colon: \texttt{"A: Paris"}\\
    - Dot: \texttt{"A. Paris"}\\
    - Bracket: \texttt{"A) Paris"}\\
    - Double brackets: \texttt{"(A) Paris"}\\
    \\
    \textbf{Option separator (between options):}\\
    - Line break: each option on a new line\\
    - Comma: \texttt{"A) Paris, B) Berlin"}\\
    - Semicolon: \texttt{"A) Paris; B) Berlin"}\\
    \\
    \textbf{Task:}\\
    Based on your internal processing characteristics, predict which combination of formatting choices is most likely to yield the highest multiple-choice accuracy for you.
    You are not given any experimental results.
    Base your decision solely on your understanding of how you process prompts.\\
    \\
    \textbf{Output format:}\\
    Return \emph{only} a valid JSON object with the following fields:\\
    \\
    \{"best\_option\_id\_set": one of ["Uppercase", "Lowercase", "Numbers", "Roman numerals"],\\
    "best\_option\_delimiter": one of ["Colon", "Dot", "Bracket", "Double brackets"],\\
    "best\_option\_separator": one of ["Line break", "Comma", "Semicolon"],\\
    "confidence": a number between 0 and 1,\\
    "rationale": "Brief explanation (1--3 sentences)"\}\\
    \\
    Do not output anything else besides valid JSON.}
    \end{VQABox}
    \caption{\textbf{Prompt template for model self-selection of optimal multiple-choice formatting.}
    The model is asked to predict which semantically neutral prompt format configuration best matches its own internal processing characteristics, without access to experimental results.
    This setup tests whether prompt format sensitivity is structured and partly accessible to the model itself.}
    \label{fig:model_prompt_self_selection}
\end{figure*}

%% file: tables/instruction_prompts.tex
\begin{table*}[!b]
    \centering
    \caption{\textbf{Instruction prompt used across all \ac{MC-VQA} datasets.}
    We employ this optimized instruction prompt to mitigate poor instruction-following behavior.
    Additionally, we include the reference prompt from V*Bench for comparison.
    Finally, we demonstrate the adaptability of our prompt to different question formats, accommodating variations in the number and sets of option IDs.}
    \label{tab:instruction_prompts}
    \begin{tabularx}{\linewidth}{l X}
         \toprule
         \textbf{Option IDs} & \textbf{Instruction prompt} \\
         \midrule
         \textit{\makecell[l]{Reference \\ (V*Bench \cite{duan_vlmevalkit_2024})}} & \textit{Please select the correct answer from the options above.} \\
         \midrule
         Template & Select the best answer to the above multiple-choice question based on the image. Respond with only the letter (e.g., \texttt{\{opt\}}, \texttt{\{opt\}}, \texttt{\{opt\}} or \texttt{\{opt\}}) of the correct option and no bracket, colon, or dot. \\
         \midrule
         \midrule
         \texttt{A/B/C/D} & Select the best answer to the above multiple-choice question based on the image. Respond with only the letter (e.g., \texttt{A}, \texttt{B}, \texttt{C} or \texttt{D}) of the correct option and no bracket, colon, or dot. \\
         \midrule
         \texttt{A/B} & Select the best answer to the above multiple-choice question based on the image. Respond with only the letter (e.g., \texttt{A}, or \texttt{B}) of the correct option and no bracket, colon, or dot. \\
         \bottomrule
    \end{tabularx}
\end{table*}

%% file: tables/model_overview.tex
\begin{table}[t]
    \centering
    \renewcommand{\arraystretch}{1.2}
    \caption{\textbf{Model overview.} We report the specific versions used of each \ac{MLLM} for reproducibility reasons of this work.}
    \label{tab:model_overview}
    \vspace{-5pt}
    \resizebox{\linewidth}{!}{
    \begin{tabular}{l c c c}
        \toprule
        \textbf{Model} & \textbf{Year} & \textbf{HuggingFace model name} & \textbf{\makecell{LLM\\ size [B]}} \\
        \midrule
         Gemma-3 & 2025 & \textit{google/gemma-3-4b-it} & $4$\\
         LLaVA-1.5 & 2023 & \textit{llava-hf/llava-1.5-7b-hf} & $7$\\
         LLaVA-OV & 2024 & \textit{lmms-lab/llava-onevision-qwen2-7b-ov} & $7$\\
         Phi-3.5 & 2024 & \textit{microsoft/Phi-3.5-vision-instruct} & $3.8$\\
         Phi-4 & 2025 & \textit{microsoft/Phi-4-multimodal-instruct} & $3.8$\\
         Qwen2-VL & 2024 & \textit{Qwen/Qwen2-VL-7B-Instruct} & $7$\\
         Qwen2.5-VL & 2025 & \textit{Qwen/Qwen2.5-VL-7B-Instruct} & $7$\\
        \bottomrule
    \end{tabular}%
    }
\end{table}

%% file: tables/datasets.tex
\begin{table*}[!t]
    \centering
    \renewcommand{\arraystretch}{1.2}
    \caption{\textbf{Statistics of the \ac{MC-VQA} datasets.} The datasets vary significantly in the number of samples and image resolution. We report the number of question-answer (QA) pairs for both the single-prompt and circular evaluation schemes. Additionally, we indicate the prompt format variations that are standard for each dataset.}
    \vspace{-5pt}
    \label{tab:dataset_overview}
    \resizebox{\linewidth}{!}{
    \begin{tabular}{l c c c @{}c c c @{}c c c c}
        \toprule
        & & \multicolumn{2}{c}{\textbf{QA pairs $[\#]$}} &
        & \multicolumn{2}{c}{\textbf{Image size} $[\text{px}]$} &
        & \multicolumn{3}{c}{\textbf{Prompt format}} \\
        \cline{3-4} \cline{6-7} \cline{9-11} \vspace{-2.5ex} \\
        \textbf{Dataset} & \textbf{No. options} & Single & Circ. eval. & & \textbf{$\overline{H}$} & \textbf{$\overline{W}$} & & Option IDs & Option del. & Option sep.\\
        \midrule
        A-OKVQA \cite{schwenk_aokvqa_2022} & $4$ & \num[round-precision=0]{1145} & \num[round-precision=0]{4580} & & \num[round-precision=0]{582} & \num[round-precision=0]{480} & & Low. & Double b. & Line break \\ 
        HR-Bench-4k \cite{wang_hrbench_2025} & $4$ & \num[round-precision=0]{200} & \num[round-precision=0]{800} & & \num[round-precision=0]{4024} & \num[round-precision=0]{3503} & & Upp. & Dot & Line break \\
        MMBench-DEV \cite{liu_mmbench_2024} & $2-4$ &\num[round-precision=0]{1292} & \num[round-precision=0]{4876} & & \num[round-precision=0]{464} & \num[round-precision=0]{349} & & Upp. & Dot & Semicolon \\ 
        MME-RW-Lite \cite{zhang_mmerealworld_2025} & $5$ & \num[round-precision=0]{1919} & \num[round-precision=0]{9595} & & \num[round-precision=0]{2836} & \num[round-precision=0]{1566} & & Upp. & Double b. & Line break \\
        V*Bench \cite{wu_vstar_2024} & $2-4$ & \num[round-precision=0]{192} & \num[round-precision=0]{596} & & \num[round-precision=0]{2246} & \num[round-precision=0]{1582} & & Upp. & Dot & Line break \\
        \bottomrule
    \end{tabular}%
    }
\end{table*}

%% file: tables/computation_report.tex
\begin{table*}[!b]
    \centering
    \caption{\textbf{Computation Report.} Computational complexity for the comprehensive analysis presented in \cref{sect:benchmark_instability}.}
    \label{tab:computation_report}
    \resizebox{0.9\textwidth}{!}{
    \begin{tabular}{l c c c c}
        \toprule
        & \multicolumn{2}{c}{\textbf{Question-level evaluations}} & \multicolumn{2}{c}{\textbf{Inference time estimates}} \\
        \textbf{Dataset} & Per prompt format variant & Total evaluations & Per question [s] & Total time [h] \\
        \midrule
         A-OKVQA & \num[round-precision=0]{4580} & \num[round-precision=0]{1538880} & \num{0.5} & \num[round-precision=0]{204.983}\\
         HRBench-4K & \num[round-precision=0]{800} & \num[round-precision=0]{268800} & \num{2} & \num[round-precision=0]{149.333}\\
         MM-Bench & \num[round-precision=0]{4876} & \num[round-precision=0]{1638336} & \num{1} & \num[round-precision=0]{455.093}\\
         MME-RW-Lite & \num[round-precision=0]{9595} & \num[round-precision=0]{3223920} & \num{1.5} & \num[round-precision=0]{1343.3}\\
         V*Bench & \num[round-precision=0]{596} & \num[round-precision=0]{200256} & \num{2} & \num[round-precision=0]{111.253}\\
        \midrule
        \textbf{Sum} & $-$ & \num[round-precision=0]{6807192} & $-$ & \num[round-precision=0]{2263.96333}\\
        \bottomrule
    \end{tabular}%
    }
\end{table*}

%% file: tables/bias_mitigation_results_aokvqa.tex
\begin{tabular}{l c c c c c c c c c c}
    \toprule
     & \multicolumn{2}{c}{\textbf{Pseudo GT}} & \multicolumn{2}{c}{\textbf{Vanilla}} & \multicolumn{2}{c}{\textbf{\ac{PIA}}} & \multicolumn{2}{c}{\textbf{\ac{PriDe}}} & \multicolumn{2}{c}{\textbf{\ac{CP-LN}}}\\
    \textbf{Model} & Acc. $[\%]$ & Rank & Acc. $[\%]$ & Rank & Acc. $[\%]$ & Rank & Acc. $[\%]$ & Rank & Acc. $[\%]$ & Rank \\
    \midrule
    Gemma-3 & \num{77.9904} & 6 & \num{78.08} & 3 & \num{61.00} & 4 & \num{78.8646} & \textbf{6} & \num{67.78} & \textbf{6}\\
    LLaVA-1.5 & \num{48.47} & 7 & \num{12.29} & \textbf{7} & \num{10.93} & \textbf{7} & \num{77.1179} & \textbf{7} & \num{68.21} & 5\\
    LLaVA-OV & \num{81.28} & 3 & \num{42.40} & 6 & \num{37.91} & 5 & \num{91.35} & 1 & \num{96.59} & 1 \\
    Phi-3.5 & \num{79.38} & 5 & \num{78.67} & 2 & \num{64.37} & 2 & \num{80.698} & \textbf{5} & \num{67.69} & 7 \\
    Phi-4 & \num{80.5} & 4 & \num{75.04} & \textbf{4} & \num{61.14} & 3 & \num{83.40} & \textbf{4} & \num{85.94} & 2\\
    Qwen-2-VL & \num{81.96} & 2 & \num{44.26} & 5 & \num{36.31} & 6 & \num{87.16} & 3 & \num{78.52} & 3\\
    Qwen-2.5-VL & \num{86.77} & 1 & \num{87.93} & \textbf{1} & \num{77.31} & \textbf{1} & \num{87.685} & 2 & \num{74.93} & 4\\
    \midrule
    \textit{Complexity $[\#] \downarrow$} & \multicolumn{2}{c}{\num[round-precision=0]{219840}} & \multicolumn{2}{c}{\num[round-precision=0]{1145}} & \multicolumn{2}{c}{\num[round-precision=0]{4580}} & \multicolumn{2}{c}{\num[round-precision=0]{1145}} & \multicolumn{2}{c}{\num[round-precision=0]{4580}} \\
    \midrule
    \midrule
    \multicolumn{11}{l}{\textbf{Comparison to Pseudo GT}} \\
    \multicolumn{3}{l}{Ranking correlation $\uparrow$} & \multicolumn{2}{c}{\num{0.36}} & \multicolumn{2}{c}{\num{0.39}} & \multicolumn{2}{c}{\num{0.89}} & \multicolumn{2}{c}{\num{0.54}} \\
    \multicolumn{3}{l}{Correctly ranked $\uparrow$} & \multicolumn{2}{c}{$(3/7)$} & \multicolumn{2}{c}{$(2/7)$} & \multicolumn{2}{c}{$(4/7)$} & \multicolumn{2}{c}{$(1/7)$} \\
    \bottomrule
\end{tabular}

%% file: tables/bias_mitigation_results_hrbench_4k.tex
\begin{tabular}{l c c c c c c c c c c}
    \toprule
     & \multicolumn{2}{c}{\textbf{Pseudo GT}} & \multicolumn{2}{c}{\textbf{Vanilla}} & \multicolumn{2}{c}{\textbf{\ac{PIA}} } & \multicolumn{2}{c}{\textbf{\ac{PriDe}}} & \multicolumn{2}{c}{\textbf{\ac{CP-LN}}}\\
    \textbf{Model} & Acc. $[\%]$ & Rank & Acc. $[\%]$ & Rank & Acc. $[\%]$ & Rank & Acc. $[\%]$ & Rank & Acc. $[\%]$ & Rank \\
    \midrule
    Gemma-3 & \num{45.91} & 6 & \num{44.25} & \textbf{6} & \num{4.99} & \textbf{6} & \num{44.25} & \textbf{6} & \num{52.00} & \textbf{6} \\
    LLaVA-1.5 & \num{29.90} & 7 & 36.75 & \textbf{7} & \num{3.41} & \textbf{7} & \num{37.125} & \textbf{7} & \num{34.50} & \textbf{7} \\
    LLaVA-OV & \num{55.21} & 4 & 65.00 & 1 & \num{10.61} & 3 & \num{65.5} & 2 & \num{64.00} & 3 \\
    Phi-3.5 & \num{47.90} & 5 & 50.50 & \textbf{5} & \num{6.4} & \textbf{5} & \num{50.125} & \textbf{5} & \num{53.00} & \textbf{5} \\
    Phi-4 & \num{61.13} & 3 & 65.00 & 1 & \num{10.66} & 2 & \num{65.25} & 4 & \num{58.00} & 4 \\
    Qwen-2-VL & \num{64.36} & 2 & 62.00 & 4 & \num{9.63} & 4 & \num{65.5} & \textbf{2} & \num{68.06} & \textbf{2} \\
    Qwen-2.5-VL & \num{69.61} & 1 & 62.75 & 3 & \num{39.47} & \textbf{1} & \num{72.5} & \textbf{1} & \num{68.5} & \textbf{1} \\
    \midrule
    \textit{Complexity $[\#] \downarrow$} & \multicolumn{2}{c}{\num[round-precision=0]{38400}} & \multicolumn{2}{c}{\num[round-precision=0]{200}} & \multicolumn{2}{c}{\num[round-precision=0]{800}} & \multicolumn{2}{c}{\num[round-precision=0]{200}} & \multicolumn{2}{c}{\num[round-precision=0]{800}} \\
    \midrule
    \midrule
    \multicolumn{11}{l}{\textbf{Comparison to Pseudo GT}} \\
    \multicolumn{3}{l}{Ranking correlation $\uparrow$} & \multicolumn{2}{c}{\num{0.66}} & \multicolumn{2}{c}{\num{0.86}} & \multicolumn{2}{c}{\num{0.98}} & \multicolumn{2}{c}{\num{1.00}} \\
    \multicolumn{3}{l}{Correctly ranked $\uparrow$} & \multicolumn{2}{c}{$(3/7)$} & \multicolumn{2}{c}{$(4/7)$} & \multicolumn{2}{c}{$(5/7)$} & \multicolumn{2}{c}{$(5/7)$} \\
    \bottomrule
\end{tabular}

%% file: tables/bias_mitigation_results_vstar_bench.tex
\begin{tabular}{l c c c c c c c c c c}
    \toprule
     & \multicolumn{2}{c}{\textbf{Pseudo GT}} 
     & \multicolumn{2}{c}{\textbf{Vanilla}} 
     & \multicolumn{2}{c}{\textbf{\ac{PIA}} } 
     & \multicolumn{2}{c}{\textbf{\ac{PriDe}}} 
     & \multicolumn{2}{c}{\textbf{\ac{CP-LN}}}\\
    \textbf{Model} & Acc. $[\%]$ & Rank & Acc. $[\%]$ & Rank & Acc. $[\%]$ & Rank & Acc. $[\%]$ & Rank & Acc. $[\%]$ & Rank \\
    \midrule
    Gemma-3 & \num{34.4135} & 6 & \num{36.13} & 7 & $-$ & $-$ & $-$ & $-$ & \num{42.93} & 7 \\
    LLaVA-1.5 & \num{31.15} & 7 & \num{40.84} & 6 & $-$ & $-$ & $-$ & $-$ & \num{44.50} & 6 \\
    LLaVA-OV & \num{62.36} & 4 & \num{72.73} & 3 & $-$ & $-$ & $-$ & $-$ & \num{68.06} & 2 \\
    Phi-3.5 & \num{47.46} & 5 & \num{52.88} & \textbf{5} & $-$ & $-$ & $-$ & $-$ & \num{53.40} & \textbf{5} \\
    Phi-4 & \num{69.8336} & 2 & \num{73.30} & 1 & $-$ & $-$ & $-$ & $-$ & \num{62.83} & 4 \\
    Qwen-2-VL & \num{69.58} & 3 & \num{54.45} & 4 & $-$ & $-$ & $-$ & $-$ & \num{65.97} & \textbf{3} \\
    Qwen-2.5-VL & \num{77.4364} & 1 & \num{72.77} & 2 & $-$ & $-$ & $-$ & $-$ & \num{69.11} & \textbf{1} \\
    \midrule
    \textit{Complexity $[\#] \downarrow$} & \multicolumn{2}{c}{\num[round-precision=0]{28608}} & \multicolumn{2}{c}{\num[round-precision=0]{191}} & \multicolumn{2}{c}{$-$} & \multicolumn{2}{c}{$-$} & \multicolumn{2}{c}{\num[round-precision=0]{596}} \\
    \midrule
    \midrule
    \multicolumn{11}{l}{\textbf{Comparison to Pseudo GT}}\\
    \multicolumn{3}{l}{Ranking correlation $\uparrow$} & \multicolumn{2}{c}{\num{0.71}} & \multicolumn{2}{c}{$-$} & \multicolumn{2}{c}{$-$} & \multicolumn{2}{c}{\num{0.93}}\\
    \multicolumn{3}{l}{Correctly ranked $\uparrow$} & \multicolumn{2}{c}{$(1/7)$} & \multicolumn{2}{c}{$-$} & \multicolumn{2}{c}{$-$} & \multicolumn{2}{c}{$(3/7)$}\\
    \bottomrule
\end{tabular}

%% file: tables/full_results_aokvqa.tex
\begin{table*}[hbtp]
    \centering
    \caption{\textbf{Full results of prompt format variations on A-OKVQA dataset.} 
    This table presents the performance of seven \acp{MLLM} under different option separators (\textit{Comma}, \textit{Line break}, \textit{Semicolon}) and delimiters (\textit{Dot}, \textit{Colon}, \textit{Bracket}, \textit{Double brackets}).
    Option ID sets are varied across four formats: \textit{Uppercase}, \textit{Lowercase}, \textit{Numbers}, and \textit{Roman numbers}.
    Metrics include accuracy (\%) and coverage (\%), with coverage consistently near 100\% and accuracy varying by format.
    The top-performing model changes across prompt format variations,alternating between Qwen2.5-VL, Qwen2-VL,and LLaVA-OV.The accuracy gap to the second-best \ac{MLLM} ranges from $0.00$ \ac{pp} (tie between two models) to $5.39$ \ac{pp}. The highest accuracy per prompt format variation is highlighted in bold.}
    \label{tab:full_results_aokvqa}
    \resizebox{\textwidth}{!}{%
    \begin{tabular}{l l l S S S S @{} c S S S S @{} c S S S S @{} c S S S S}
    \toprule
    & & & \multicolumn{19}{c}{Option delimiter} \\
    \cline{4-22} \vspace{-2.5ex}\\
    & & & \multicolumn{4}{c}{\textbf{Dot}} & & \multicolumn{4}{c}{\textbf{Colon}} & & \multicolumn{4}{c}{\textbf{Bracket}} & & \multicolumn{4}{c}{\textbf{Double brackets}} \\
    & & & \multicolumn{4}{c}{Option IDs} & & \multicolumn{4}{c}{Option IDs} & & \multicolumn{4}{c}{Option IDs} & & \multicolumn{4}{c}{Option IDs}\\
    \cline{4-7} \cline{9-12} \cline{14-17} \cline{19-22} \vspace{-2.5ex} \\
    \textbf{Option sep.} & \textbf{Model} & & \textbf{Upp.} & \textbf{Low.} & \textbf{Num.} & \textbf{Rom.}
     & & \textbf{Upp.} & \textbf{Low.} & \textbf{Num.} & \textbf{Rom.}
     & & \textbf{Upp.} & \textbf{Low.} & \textbf{Num.} & \textbf{Rom.}
     & & \textbf{Upp.} & \textbf{Low.} & \textbf{Num.} & \textbf{Rom.} \\
    \midrule
    \multirow{14}{*}{Comma}
     & \multirow{2}{*}{Gemma-3} & Acc. [\%] & 77.969 & 78.362 & 77.838 & 77.336 &  & 77.511 & 77.838 & 77.664 & 77.14 &  & 78.079 & 78.559 & 77.882 & 77.445 &  & 78.559 & 78.559 & 77.969 & 77.773 \\
    & & Cov. [\%] & 100.0 & 100.0 & 100.0 & 100.0 & & 100.0 & 100.0 & 100.0 & 100.0 & & 100.0 & 100.0 & 100.0 & 100.0 & & 100.0 & 99.978 & 100.0 & 99.956 \\
    \cline{2-22} \vspace{-2.5ex}\\
     & \multirow{2}{*}{LLaVA-1.5} & Acc. [\%] & 73.821 & 45.175 & 71.201 & 71.812 &  & 72.183 & 52.489 & 70.349 & 71.048 &  & 73.799 & 55.568 & 73.886 & 72.009 &  & 75.415 & 8.275 & 73.472 & 46.026 \\
    & & Cov. [\%] & 100.0 & 67.096 & 99.76 & 99.869 & & 100.0 & 80.087 & 99.913 & 99.978 & & 100.0 & 80.546 & 99.869 & 99.934 & & 99.956 & 10.262 & 99.585 & 55.437 \\
    \cline{2-22} \vspace{-2.5ex}\\
     & \multirow{2}{*}{LLaVA-OV} & Acc. [\%] & \bfseries 89.847 & \bfseries 89.541 & \bfseries 89.76 & 80.721 &  & \bfseries 90.415 & \bfseries 90.066 & \bfseries 89.803 & 74.279 &  & \bfseries 90.611 & \bfseries 90.328 & \bfseries 90.153 & 77.576 &  & \bfseries 90.59 & 3.821 & \bfseries 89.716 & 73.144 \\
    & & Cov. [\%] & 100.0 & 99.978 & 99.782 & 99.934 & & 100.0 & 100.0 & 99.891 & 99.782 & & 100.0 & 100.0 & 99.934 & 99.956 & & 100.0 & 4.389 & 99.782 & 99.454 \\
    \cline{2-22} \vspace{-2.5ex}\\
     & \multirow{2}{*}{Phi-3.5} & Acc. [\%] & 76.921 & 78.253 & 79.214 & 78.996 &  & 79.541 & 79.694 & 79.498 & 79.127 &  & 78.297 & 79.192 & 79.52 & 79.323 &  & 79.651 & 69.847 & 79.214 & 78.559 \\
    & & Cov. [\%] & 100.0 & 100.0 & 100.0 & 100.0 & & 100.0 & 100.0 & 100.0 & 100.0 & & 100.0 & 100.0 & 100.0 & 100.0 & & 100.0 & 84.52 & 99.934 & 97.991 \\
    \cline{2-22} \vspace{-2.5ex}\\
     & \multirow{2}{*}{Phi-4} & Acc. [\%] & 82.773 & 82.489 & 82.358 & 72.118 &  & 82.773 & 82.86 & 82.293 & 73.035 &  & 82.817 & 82.729 & 82.62 & 69.541 &  & 83.035 & 82.838 & 82.576 & 76.572 \\
    & & Cov. [\%] & 100.0 & 100.0 & 100.0 & 89.847 & & 100.0 & 100.0 & 100.0 & 90.349 & & 100.0 & 100.0 & 100.0 & 87.162 & & 100.0 & 99.934 & 100.0 & 94.76 \\
    \cline{2-22} \vspace{-2.5ex}\\
     & \multirow{2}{*}{Qwen-2-VL} & Acc. [\%] & 83.581 & 84.258 & 85.59 & 85.328 &  & 84.148 & 85.349 & 85.197 & \bfseries 85.699 &  & 84.258 & 85.087 & 85.59 & \bfseries 85.59 &  & 65.939 & 35.764 & 84.214 & \bfseries 85.939 \\
    & & Cov. [\%] & 100.0 & 100.0 & 99.825 & 100.0 & & 100.0 & 100.0 & 99.825 & 100.0 & & 100.0 & 100.0 & 99.825 & 100.0 & & 78.079 & 44.061 & 98.428 & 99.913 \\
    \cline{2-22} \vspace{-2.5ex}\\
     & \multirow{2}{*}{Qwen-2.5-VL} & Acc. [\%] & 87.314 & 87.205 & 85.699 & \bfseries 85.939 &  & 86.965 & 87.031 & 85.677 & 85.131 &  & 87.227 & 86.987 & 86.048 & \bfseries 85.59 &  & 87.555 & \bfseries 87.358 & 85.786 & 85.131 \\
    & & Cov. [\%] & 100.0 & 100.0 & 99.869 & 100.0 & & 100.0 & 99.978 & 99.869 & 100.0 & & 100.0 & 100.0 & 99.869 & 100.0 & & 100.0 & 100.0 & 99.847 & 100.0 \\
    \midrule
    \multirow{14}{*}{Line break}
     & \multirow{2}{*}{Gemma-3} & Acc. [\%] & 78.559 & 78.188 & 78.166 & 77.751 &  & 78.472 & 78.406 & 78.297 & 77.817 &  & 78.624 & 78.646 & 78.297 & 77.817 &  & 78.144 & 78.079 & 78.122 & 78.057 \\
    & & Cov. [\%] & 100.0 & 100.0 & 100.0 & 100.0 & & 100.0 & 100.0 & 100.0 & 100.0 & & 100.0 & 100.0 & 100.0 & 100.0 & & 99.978 & 100.0 & 100.0 & 100.0 \\
    \cline{2-22} \vspace{-2.5ex}\\
     & \multirow{2}{*}{LLaVA-1.5} & Acc. [\%] & 79.105 & 69.978 & 76.9 & 77.576 &  & 78.559 & 76.048 & 77.227 & 77.031 &  & 78.886 & 74.258 & 77.271 & 77.358 &  & 78.886 & 12.293 & 76.703 & 38.603 \\
    & & Cov. [\%] & 100.0 & 89.498 & 99.738 & 99.934 & & 100.0 & 97.467 & 99.934 & 100.0 & & 100.0 & 94.476 & 99.891 & 99.978 & & 100.0 & 13.865 & 99.301 & 42.031 \\
    \cline{2-22} \vspace{-2.5ex}\\
     & \multirow{2}{*}{LLaVA-OV} & Acc. [\%] & \bfseries 91.07 & \bfseries 90.764 & \bfseries 90.022 & 59.847 &  & \bfseries 90.961 & \bfseries 90.611 & \bfseries 90.502 & 62.795 &  & \bfseries 90.873 & \bfseries 90.895 & \bfseries 90.371 & 63.581 &  & \bfseries 90.742 & 42.402 & \bfseries 90.262 & 67.271 \\
    & & Cov. [\%] & 100.0 & 100.0 & 99.716 & 99.41 & & 100.0 & 100.0 & 99.913 & 99.607 & & 100.0 & 100.0 & 99.847 & 99.694 & & 100.0 & 47.467 & 99.738 & 99.585 \\
    \cline{2-22} \vspace{-2.5ex}\\
     & \multirow{2}{*}{Phi-3.5} & Acc. [\%] & 81.004 & 81.048 & 81.026 & 80.087 &  & 80.873 & 80.983 & 80.437 & 79.847 &  & 81.07 & 80.873 & 80.808 & 80.131 &  & 81.048 & 78.668 & 80.983 & 79.847 \\
    & & Cov. [\%] & 100.0 & 100.0 & 100.0 & 100.0 & & 100.0 & 100.0 & 100.0 & 100.0 & & 100.0 & 100.0 & 100.0 & 100.0 & & 100.0 & 96.179 & 100.0 & 99.716 \\
    \cline{2-22} \vspace{-2.5ex}\\
     & \multirow{2}{*}{Phi-4} & Acc. [\%] & 82.62 & 82.336 & 82.183 & 78.166 &  & 82.882 & 82.707 & 82.533 & 79.105 &  & 82.664 & 82.533 & 81.965 & 76.201 &  & 82.904 & 82.598 & 82.598 & 75.655 \\
    & & Cov. [\%] & 100.0 & 100.0 & 100.0 & 96.157 & & 100.0 & 100.0 & 100.0 & 96.79 & & 100.0 & 100.0 & 100.0 & 93.777 & & 100.0 & 100.0 & 100.0 & 93.712 \\
    \cline{2-22} \vspace{-2.5ex}\\
     & \multirow{2}{*}{Qwen-2-VL} & Acc. [\%] & 86.572 & 86.834 & 85.633 & 84.978 &  & 86.616 & 86.616 & 85.59 & 85.349 &  & 86.659 & 86.419 & 85.808 & 85.131 &  & 78.974 & 44.258 & 85.218 & 85.611 \\
    & & Cov. [\%] & 100.0 & 100.0 & 99.913 & 100.0 & & 100.0 & 100.0 & 99.891 & 100.0 & & 100.0 & 100.0 & 99.913 & 100.0 & & 91.769 & 53.952 & 99.258 & 99.847 \\
    \cline{2-22} \vspace{-2.5ex}\\
     & \multirow{2}{*}{Qwen-2.5-VL} & Acc. [\%] & 87.991 & 87.773 & 86.485 & \bfseries 86.157 &  & 88.122 & 87.795 & 86.528 & \bfseries 86.092 &  & 87.904 & 87.86 & 86.528 & \bfseries 86.026 &  & 88.166 & \bfseries 87.991 & 86.834 & \bfseries 85.677 \\
    & & Cov. [\%] & 100.0 & 100.0 & 99.869 & 99.978 & & 100.0 & 100.0 & 99.847 & 100.0 & & 100.0 & 100.0 & 99.869 & 99.978 & & 100.0 & 100.0 & 99.847 & 99.956 \\
    \midrule
    \multirow{14}{*}{Semicolon}
     & \multirow{2}{*}{Gemma-3} & Acc. [\%] & 77.969 & 78.297 & 77.402 & 77.555 &  & 77.489 & 77.86 & 77.555 & 77.511 &  & 78.1 & 78.428 & 77.533 & 77.511 &  & 78.297 & 78.341 & 77.904 & 77.86 \\
    & & Cov. [\%] & 100.0 & 100.0 & 100.0 & 100.0 & & 100.0 & 100.0 & 100.0 & 100.0 & & 100.0 & 100.0 & 100.0 & 100.0 & & 100.0 & 100.0 & 100.0 & 99.978 \\
    \cline{2-22} \vspace{-2.5ex}\\
     & \multirow{2}{*}{LLaVA-1.5} & Acc. [\%] & 77.467 & 60.742 & 73.93 & 73.908 &  & 75.633 & 68.144 & 73.734 & 73.013 &  & 76.114 & 68.712 & 74.41 & 73.712 &  & 76.397 & 12.162 & 74.367 & 54.323 \\
    & & Cov. [\%] & 100.0 & 80.786 & 99.847 & 99.934 & & 100.0 & 90.895 & 99.956 & 100.0 & & 100.0 & 90.917 & 99.869 & 99.956 & & 99.978 & 14.323 & 99.891 & 65.808 \\
    \cline{2-22} \vspace{-2.5ex}\\
     & \multirow{2}{*}{LLaVA-OV} & Acc. [\%] & \bfseries 90.0 & \bfseries 90.24 & \bfseries 89.76 & 79.389 &  & \bfseries 90.852 & \bfseries 90.328 & \bfseries 90.175 & 73.581 &  & \bfseries 90.939 & \bfseries 90.655 & \bfseries 90.022 & 77.424 &  & \bfseries 90.721 & 11.659 & \bfseries 89.825 & 72.664 \\
    & & Cov. [\%] & 100.0 & 100.0 & 99.891 & 99.825 & & 100.0 & 100.0 & 99.956 & 99.825 & & 100.0 & 100.0 & 99.913 & 99.956 & & 100.0 & 13.297 & 99.782 & 99.716 \\
    \cline{2-22} \vspace{-2.5ex}\\
     & \multirow{2}{*}{Phi-3.5} & Acc. [\%] & 78.886 & 79.389 & 79.782 & 79.323 &  & 79.694 & 79.913 & 79.76 & 79.301 &  & 79.563 & 79.76 & 80.175 & 79.782 &  & 79.76 & 72.533 & 80.044 & 79.148 \\
    & & Cov. [\%] & 100.0 & 100.0 & 100.0 & 100.0 & & 100.0 & 100.0 & 100.0 & 100.0 & & 100.0 & 100.0 & 100.0 & 100.0 & & 100.0 & 87.489 & 99.934 & 98.865 \\
    \cline{2-22} \vspace{-2.5ex}\\
     & \multirow{2}{*}{Phi-4} & Acc. [\%] & 82.598 & 82.358 & 82.205 & 71.834 &  & 82.838 & 82.62 & 82.424 & 72.751 &  & 82.729 & 82.511 & 82.62 & 68.297 &  & 82.729 & 82.62 & 82.576 & 77.598 \\
    & & Cov. [\%] & 100.0 & 100.0 & 100.0 & 89.389 & & 100.0 & 100.0 & 100.0 & 89.738 & & 100.0 & 100.0 & 100.0 & 84.978 & & 100.0 & 99.934 & 100.0 & 95.502 \\
    \cline{2-22} \vspace{-2.5ex}\\
     & \multirow{2}{*}{Qwen-2-VL} & Acc. [\%] & 84.869 & 86.266 & 85.895 & 85.437 &  & 86.114 & 86.55 & 85.59 & 85.655 &  & 86.288 & 86.463 & 85.983 & 85.721 &  & 73.646 & 41.092 & 84.956 & \bfseries 85.917 \\
    & & Cov. [\%] & 100.0 & 100.0 & 99.869 & 100.0 & & 100.0 & 100.0 & 99.847 & 100.0 & & 100.0 & 100.0 & 99.869 & 100.0 & & 86.004 & 50.022 & 98.908 & 99.956 \\
    \cline{2-22} \vspace{-2.5ex}\\
     & \multirow{2}{*}{Qwen-2.5-VL} & Acc. [\%] & 87.838 & 87.576 & 86.114 & \bfseries 85.873 &  & 87.598 & 87.424 & 85.895 & \bfseries 85.721 &  & 87.402 & 87.576 & 86.332 & \bfseries 85.917 &  & 87.62 & \bfseries 87.555 & 86.397 & 85.502 \\
    & & Cov. [\%] & 100.0 & 100.0 & 99.869 & 100.0 & & 100.0 & 100.0 & 99.847 & 100.0 & & 100.0 & 100.0 & 99.869 & 100.0 & & 100.0 & 100.0 & 99.847 & 99.978 \\
    \bottomrule
    \end{tabular}%
    }
\end{table*}

%% file: tables/full_results_hr_bench_4k.tex
\begin{table*}[hbtp]
    \centering
    \caption{\textbf{Full results of prompt format variations on HR-Bench-4K dataset.} 
    This table presents the performance of seven \acp{MLLM} under different option separators (\textit{Comma}, \textit{Line break}, \textit{Semicolon}) and delimiters (\textit{Dot}, \textit{Colon}, \textit{Bracket}, \textit{Double brackets}).
    Option ID sets are varied across four formats: \textit{Uppercase}, \textit{Lowercase}, \textit{Numbers}, and \textit{Roman numbers}.
    Metrics include accuracy (\%) and coverage (\%), with coverage consistently near 100\% and accuracy varying by format.
    Qwen-2.5-VL is consistently the top-performing \ac{MLLM}, demonstrating strong robustness.
    The accuracy gap to the second-best \ac{MLLM} ranges from $0.25$ \ac{pp} to $6.88$ \ac{pp}.
    The highest accuracy per prompt format variation is highlighted in bold.}
    \label{tab:full_results_hrbench}
    \resizebox{\textwidth}{!}{%
    \begin{tabular}{l l l S S S S @{} c S S S S @{} c S S S S @{} c S S S S}
    \toprule
    & & & \multicolumn{19}{c}{Option delimiter} \\
    \cline{4-22} \vspace{-2.5ex}\\
    & & & \multicolumn{4}{c}{\textbf{Dot}} & & \multicolumn{4}{c}{\textbf{Colon}} & & \multicolumn{4}{c}{\textbf{Bracket}} & & \multicolumn{4}{c}{\textbf{Double brackets}} \\
    & & & \multicolumn{4}{c}{Option IDs} & & \multicolumn{4}{c}{Option IDs} & & \multicolumn{4}{c}{Option IDs} & & \multicolumn{4}{c}{Option IDs}\\
    \cline{4-7} \cline{9-12} \cline{14-17} \cline{19-22} \vspace{-2.5ex} \\
    \textbf{Option sep.} & \textbf{Model} & & \textbf{Upp.} & \textbf{Low.} & \textbf{Num.} & \textbf{Rom.}
     & & \textbf{Upp.} & \textbf{Low.} & \textbf{Num.} & \textbf{Rom.}
     & & \textbf{Upp.} & \textbf{Low.} & \textbf{Num.} & \textbf{Rom.}
     & & \textbf{Upp.} & \textbf{Low.} & \textbf{Num.} & \textbf{Rom.} \\
    \midrule
    \multirow{14}{*}{Comma}
     & \multirow{2}{*}{Gemma-3} & Acc. [\%] & 45.75 & 46.0 & 43.5 & 46.375 &  & 46.25 & 46.375 & 44.25 & 46.25 &  & 46.375 & 46.25 & 46.375 & 46.875 &  & 45.625 & 46.375 & 45.0 & 46.875 \\
    & & Cov. [\%] & 100.0 & 100.0 & 100.0 & 100.0 & & 100.0 & 100.0 & 100.0 & 100.0 & & 100.0 & 100.0 & 100.0 & 100.0 & & 100.0 & 100.0 & 100.0 & 100.0 \\
    \cline{2-22} \vspace{-2.5ex}\\
     & \multirow{2}{*}{LLaVA-1.5} & Acc. [\%] & 33.0 & 24.375 & 30.5 & 33.375 &  & 33.125 & 25.375 & 30.875 & 31.75 &  & 35.125 & 27.0 & 32.25 & 33.375 &  & 35.875 & 1.875 & 32.875 & 11.875 \\
    & & Cov. [\%] & 100.0 & 84.875 & 97.25 & 99.875 & & 100.0 & 89.25 & 100.0 & 100.0 & & 100.0 & 87.25 & 99.75 & 100.0 & & 100.0 & 4.75 & 97.125 & 32.125 \\
    \cline{2-22} \vspace{-2.5ex}\\
     & \multirow{2}{*}{LLaVA-OV} & Acc. [\%] & 64.25 & 62.375 & 60.25 & 47.75 &  & 64.25 & 63.875 & 60.375 & 44.625 &  & 64.5 & 63.875 & 61.375 & 46.625 &  & 64.125 & 4.375 & 62.0 & 40.125 \\
    & & Cov. [\%] & 100.0 & 100.0 & 100.0 & 99.625 & & 100.0 & 100.0 & 100.0 & 99.625 & & 100.0 & 100.0 & 100.0 & 99.875 & & 100.0 & 6.625 & 99.625 & 99.625 \\
    \cline{2-22} \vspace{-2.5ex}\\
     & \multirow{2}{*}{Phi-3.5} & Acc. [\%] & 46.875 & 48.0 & 46.625 & 45.5 &  & 47.75 & 47.25 & 45.25 & 45.375 &  & 47.25 & 48.25 & 48.375 & 45.875 &  & 50.25 & 39.625 & 47.875 & 45.125 \\
    & & Cov. [\%] & 100.0 & 100.0 & 100.0 & 100.0 & & 100.0 & 100.0 & 99.5 & 100.0 & & 100.0 & 100.0 & 100.0 & 100.0 & & 100.0 & 78.125 & 97.375 & 99.75 \\
    \cline{2-22} \vspace{-2.5ex}\\
     & \multirow{2}{*}{Phi-4} & Acc. [\%] & 62.0 & 63.25 & 60.375 & 53.25 &  & 63.5 & 63.625 & 63.125 & 54.375 &  & 63.5 & 62.75 & 62.5 & 49.625 &  & 63.125 & 63.25 & 61.875 & 50.875 \\
    & & Cov. [\%] & 100.0 & 100.0 & 100.0 & 94.875 & & 100.0 & 100.0 & 100.0 & 94.0 & & 100.0 & 100.0 & 100.0 & 92.375 & & 100.0 & 100.0 & 100.0 & 93.5 \\
    \cline{2-22} \vspace{-2.5ex}\\
     & \multirow{2}{*}{Qwen-2-VL} & Acc. [\%] & 66.0 & 66.125 & 67.625 & 66.375 &  & 65.5 & 66.75 & 67.5 & 66.625 &  & 66.25 & 66.25 & 66.875 & 66.375 &  & 59.875 & 25.625 & 66.0 & \bfseries 67.625 \\
    & & Cov. [\%] & 100.0 & 100.0 & 100.0 & 100.0 & & 100.0 & 100.0 & 100.0 & 100.0 & & 100.0 & 100.0 & 99.875 & 100.0 & & 92.25 & 43.875 & 100.0 & 100.0 \\
    \cline{2-22} \vspace{-2.5ex}\\
     & \multirow{2}{*}{Qwen-2.5-VL} & Acc. [\%] & \bfseries 70.625 & \bfseries 69.875 & \bfseries 68.875 & \bfseries 67.875 &  & \bfseries 69.625 & \bfseries 69.625 & \bfseries 67.75 & \bfseries 67.375 &  & \bfseries 70.5 & \bfseries 70.375 & \bfseries 68.875 & \bfseries 67.875 &  & \bfseries 70.5 & \bfseries 70.0 & \bfseries 69.375 & 66.5 \\
    & & Cov. [\%] & 100.0 & 100.0 & 99.75 & 100.0 & & 100.0 & 99.875 & 99.875 & 100.0 & & 100.0 & 100.0 & 99.875 & 100.0 & & 100.0 & 99.875 & 100.0 & 100.0 \\
    \midrule
    \multirow{14}{*}{Line break}
     & \multirow{2}{*}{Gemma-3} & Acc. [\%] & 44.25 & 44.875 & 45.125 & 46.875 &  & 46.0 & 45.875 & 46.0 & 46.125 &  & 44.875 & 44.625 & 45.625 & 46.625 &  & 45.5 & 44.75 & 45.125 & 46.75 \\
    & & Cov. [\%] & 100.0 & 100.0 & 100.0 & 100.0 & & 100.0 & 100.0 & 100.0 & 100.0 & & 100.0 & 100.0 & 100.0 & 100.0 & & 100.0 & 100.0 & 100.0 & 100.0 \\
    \cline{2-22} \vspace{-2.5ex}\\
     & \multirow{2}{*}{LLaVA-1.5} & Acc. [\%] & 36.75 & 33.0 & 34.0 & 34.0 &  & 36.75 & 33.25 & 34.5 & 34.125 &  & 36.375 & 34.125 & 35.0 & 33.625 &  & 35.375 & 0.875 & 34.25 & 5.625 \\
    & & Cov. [\%] & 100.0 & 93.375 & 97.0 & 100.0 & & 100.0 & 96.625 & 99.125 & 100.0 & & 100.0 & 95.0 & 99.625 & 100.0 & & 100.0 & 1.875 & 94.875 & 9.25 \\
    \cline{2-22} \vspace{-2.5ex}\\
     & \multirow{2}{*}{LLaVA-OV} & Acc. [\%] & 65.0 & 65.75 & 63.5 & 34.25 &  & 65.25 & 65.875 & 63.5 & 34.75 &  & 65.125 & 65.25 & 63.25 & 36.0 &  & 65.875 & 42.375 & 63.625 & 35.625 \\
    & & Cov. [\%] & 100.0 & 100.0 & 99.125 & 99.0 & & 100.0 & 100.0 & 99.875 & 99.625 & & 100.0 & 100.0 & 99.375 & 99.75 & & 100.0 & 65.875 & 99.125 & 99.75 \\
    \cline{2-22} \vspace{-2.5ex}\\
     & \multirow{2}{*}{Phi-3.5} & Acc. [\%] & 50.5 & 49.75 & 49.0 & 49.625 &  & 50.625 & 50.25 & 50.25 & 49.125 &  & 50.375 & 49.625 & 50.0 & 48.75 &  & 50.75 & 47.625 & 50.375 & 49.5 \\
    & & Cov. [\%] & 100.0 & 100.0 & 97.875 & 99.75 & & 100.0 & 100.0 & 99.0 & 99.875 & & 100.0 & 100.0 & 99.125 & 99.75 & & 100.0 & 93.125 & 97.875 & 99.875 \\
    \cline{2-22} \vspace{-2.5ex}\\
     & \multirow{2}{*}{Phi-4} & Acc. [\%] & 65.0 & 65.5 & 63.25 & 60.0 &  & 65.0 & 65.125 & 63.25 & 61.375 &  & 64.25 & 65.5 & 63.875 & 56.375 &  & 64.5 & 64.5 & 63.5 & 54.0 \\
    & & Cov. [\%] & 100.0 & 100.0 & 100.0 & 99.125 & & 100.0 & 100.0 & 100.0 & 99.375 & & 100.0 & 100.0 & 100.0 & 96.375 & & 100.0 & 100.0 & 100.0 & 93.125 \\
    \cline{2-22} \vspace{-2.5ex}\\
     & \multirow{2}{*}{Qwen-2-VL} & Acc. [\%] & 67.125 & 67.375 & 68.625 & 65.875 &  & 67.0 & 67.5 & 68.375 & 67.0 &  & 66.625 & 66.875 & 68.375 & 67.5 &  & 64.75 & 25.875 & 68.75 & 68.375 \\
    & & Cov. [\%] & 100.0 & 100.0 & 100.0 & 100.0 & & 100.0 & 100.0 & 100.0 & 100.0 & & 100.0 & 100.0 & 100.0 & 100.0 & & 97.75 & 47.0 & 99.875 & 100.0 \\
    \cline{2-22} \vspace{-2.5ex}\\
     & \multirow{2}{*}{Qwen-2.5-VL} & Acc. [\%] & \bfseries 71.25 & \bfseries 69.875 & \bfseries 69.875 & \bfseries 69.0 &  & \bfseries 71.0 & \bfseries 70.375 & \bfseries 69.625 & \bfseries 68.875 &  & \bfseries 71.0 & \bfseries 70.375 & \bfseries 69.625 & \bfseries 68.625 &  & \bfseries 71.5 & \bfseries 70.5 & \bfseries 69.75 & \bfseries 68.875 \\
    & & Cov. [\%] & 100.0 & 100.0 & 100.0 & 100.0 & & 100.0 & 100.0 & 99.625 & 100.0 & & 100.0 & 100.0 & 100.0 & 100.0 & & 100.0 & 100.0 & 99.875 & 100.0 \\
    \midrule
    \multirow{14}{*}{Semicolon}
     & \multirow{2}{*}{Gemma-3} & Acc. [\%] & 46.375 & 46.875 & 44.875 & 46.625 &  & 46.875 & 46.75 & 45.0 & 47.5 &  & 46.375 & 46.375 & 45.25 & 46.625 &  & 45.375 & 47.0 & 45.625 & 46.625 \\
    & & Cov. [\%] & 100.0 & 100.0 & 100.0 & 100.0 & & 100.0 & 100.0 & 100.0 & 100.0 & & 100.0 & 100.0 & 100.0 & 100.0 & & 100.0 & 100.0 & 100.0 & 100.0 \\
    \cline{2-22} \vspace{-2.5ex}\\
     & \multirow{2}{*}{LLaVA-1.5} & Acc. [\%] & 37.25 & 29.875 & 32.625 & 35.5 &  & 36.625 & 32.875 & 32.5 & 32.75 &  & 36.875 & 33.125 & 34.0 & 33.875 &  & 35.625 & 2.0 & 32.625 & 13.0 \\
    & & Cov. [\%] & 100.0 & 91.875 & 98.75 & 99.875 & & 100.0 & 95.125 & 99.75 & 100.0 & & 100.0 & 94.625 & 100.0 & 100.0 & & 100.0 & 5.375 & 96.75 & 33.875 \\
    \cline{2-22} \vspace{-2.5ex}\\
     & \multirow{2}{*}{LLaVA-OV} & Acc. [\%] & 64.125 & 63.375 & 61.5 & 46.5 &  & 65.5 & 65.125 & 61.625 & 41.75 &  & 64.75 & 65.125 & 61.5 & 43.875 &  & 64.375 & 9.375 & 62.875 & 39.125 \\
    & & Cov. [\%] & 100.0 & 100.0 & 100.0 & 99.25 & & 100.0 & 100.0 & 100.0 & 99.625 & & 100.0 & 100.0 & 99.875 & 99.75 & & 100.0 & 15.5 & 99.875 & 99.5 \\
    \cline{2-22} \vspace{-2.5ex}\\
     & \multirow{2}{*}{Phi-3.5} & Acc. [\%] & 48.125 & 48.375 & 46.5 & 45.875 &  & 48.5 & 48.5 & 48.125 & 46.625 &  & 48.875 & 48.375 & 47.875 & 47.375 &  & 49.875 & 39.125 & 48.375 & 47.125 \\
    & & Cov. [\%] & 100.0 & 100.0 & 99.5 & 99.75 & & 100.0 & 100.0 & 99.875 & 100.0 & & 100.0 & 100.0 & 100.0 & 99.75 & & 100.0 & 79.125 & 97.625 & 100.0 \\
    \cline{2-22} \vspace{-2.5ex}\\
     & \multirow{2}{*}{Phi-4} & Acc. [\%] & 62.125 & 62.0 & 61.375 & 53.875 &  & 63.0 & 62.625 & 62.375 & 55.875 &  & 62.75 & 63.5 & 62.75 & 51.5 &  & 63.625 & 63.125 & 62.75 & 54.875 \\
    & & Cov. [\%] & 100.0 & 100.0 & 100.0 & 93.875 & & 100.0 & 100.0 & 100.0 & 94.0 & & 100.0 & 100.0 & 100.0 & 91.0 & & 100.0 & 100.0 & 100.0 & 95.75 \\
    \cline{2-22} \vspace{-2.5ex}\\
     & \multirow{2}{*}{Qwen-2-VL} & Acc. [\%] & 67.0 & 68.0 & 67.875 & 66.375 &  & 67.75 & 68.125 & 67.25 & 67.25 &  & 67.375 & 68.125 & 68.375 & 68.125 &  & 62.25 & 27.375 & 67.125 & 67.625 \\
    & & Cov. [\%] & 100.0 & 100.0 & 100.0 & 100.0 & & 100.0 & 100.0 & 100.0 & 100.0 & & 100.0 & 100.0 & 100.0 & 100.0 & & 94.25 & 49.125 & 99.875 & 100.0 \\
    \cline{2-22} \vspace{-2.5ex}\\
     & \multirow{2}{*}{Qwen-2.5-VL} & Acc. [\%] & \bfseries 71.75 & \bfseries 69.875 & \bfseries 69.25 & \bfseries 69.0 &  & \bfseries 70.25 & \bfseries 69.875 & \bfseries 69.375 & \bfseries 68.875 &  & \bfseries 69.625 & \bfseries 70.5 & \bfseries 69.625 & \bfseries 69.125 &  & \bfseries 71.25 & \bfseries 69.25 & \bfseries 69.75 & \bfseries 68.375 \\
    & & Cov. [\%] & 100.0 & 100.0 & 100.0 & 100.0 & & 100.0 & 99.875 & 99.875 & 100.0 & & 100.0 & 99.875 & 100.0 & 100.0 & & 100.0 & 99.875 & 100.0 & 100.0 \\
    \bottomrule
    \end{tabular}%
    }
\end{table*}

%% file: tables/full_results_mmbench.tex
\begin{table*}[hbtp]
\centering
 \caption{\textbf{Full results of prompt format variations on MM-Bench dataset.} 
This table presents the performance of seven \acp{MLLM} under different option separators (\textit{Comma}, \textit{Line break}, \textit{Semicolon}) and delimiters (\textit{Dot}, \textit{Colon}, \textit{Bracket}, \textit{Double brackets}).
Option ID sets are varied across four formats: \textit{Uppercase}, \textit{Lowercase}, \textit{Numbers}, and \textit{Roman numbers}.
Metrics include accuracy (\%) and coverage (\%), with coverage consistently near 100\% and accuracy varying by format.
Qwen-2.5-VL is consistently the top-performing \ac{MLLM}, demonstrating strong robustness.
The accuracy gap to the second-best \ac{MLLM} ranges from $0.10$ \ac{pp} to $3.84$ \ac{pp}.
The highest accuracy per prompt format variation is highlighted in bold.}
 \label{tab:full_results_mmbench}
 \resizebox{\textwidth}{!}{%
 \begin{tabular}{l l l S S S S @{} c S S S S @{} c S S S S @{} c S S S S}
\toprule
& & & \multicolumn{19}{c}{Option delimiter} \\
\cline{4-22} \vspace{-2.5ex}\\
& & & \multicolumn{4}{c}{\textbf{Dot}} & & \multicolumn{4}{c}{\textbf{Colon}} & & \multicolumn{4}{c}{\textbf{Bracket}} & & \multicolumn{4}{c}{\textbf{Double brackets}} \\
& & & \multicolumn{4}{c}{Option IDs} & & \multicolumn{4}{c}{Option IDs} & & \multicolumn{4}{c}{Option IDs} & & \multicolumn{4}{c}{Option IDs}\\
\cline{4-7} \cline{9-12} \cline{14-17} \cline{19-22} \vspace{-2.5ex} \\
\textbf{Option sep.} & \textbf{Model} & & \textbf{Upp.} & \textbf{Low.} & \textbf{Num.} & \textbf{Rom.}
 & & \textbf{Upp.} & \textbf{Low.} & \textbf{Num.} & \textbf{Rom.}
 & & \textbf{Upp.} & \textbf{Low.} & \textbf{Num.} & \textbf{Rom.}
 & & \textbf{Upp.} & \textbf{Low.} & \textbf{Num.} & \textbf{Rom.} \\
\midrule
\multirow{14}{*}{Comma}
 & \multirow{2}{*}{Gemma-3} & Acc. [\%] & 76.743 & 77.523 & 77.235 & 76.907 &  & 77.03 & 77.235 & 77.666 & 76.6 &  & 76.846 & 77.482 & 77.461 & 77.092 &  & 78.117 & 78.035 & 77.666 & 77.543 \\
& & Cov. [\%] & 100.0 & 100.0 & 100.0 & 100.0 & & 100.0 & 100.0 & 100.0 & 100.0 & & 100.0 & 100.0 & 100.0 & 100.0 & & 100.0 & 100.0 & 100.0 & 100.0 \\
\cline{2-22} \vspace{-2.5ex}\\
 & \multirow{2}{*}{LLaVA-1.5} & Acc. [\%] & 66.632 & 52.687 & 65.689 & 65.422 &  & 66.12 & 55.989 & 64.91 & 64.889 &  & 66.776 & 56.87 & 67.248 & 65.053 &  & 69.073 & 14.52 & 66.591 & 46.452 \\
& & Cov. [\%] & 99.959 & 83.757 & 99.508 & 99.159 & & 99.959 & 90.033 & 99.713 & 99.426 & & 99.938 & 89.151 & 100.0 & 99.241 & & 99.897 & 17.555 & 99.241 & 57.219 \\
\cline{2-22} \vspace{-2.5ex}\\
 & \multirow{2}{*}{LLaVA-OV} & Acc. [\%] & 84.578 & 84.085 & 83.696 & 78.917 &  & 84.762 & 84.331 & 84.393 & 76.272 &  & 84.762 & 84.454 & 84.147 & 77.687 &  & 85.254 & 9.249 & 84.619 & 74.754 \\
& & Cov. [\%] & 99.979 & 99.959 & 99.569 & 99.077 & & 100.0 & 100.0 & 99.795 & 99.61 & & 100.0 & 100.0 & 99.692 & 99.61 & & 100.0 & 11.095 & 99.569 & 99.282 \\
\cline{2-22} \vspace{-2.5ex}\\
 & \multirow{2}{*}{Phi-3.5} & Acc. [\%] & 75.554 & 76.6 & 77.769 & 77.297 &  & 77.851 & 78.097 & 77.482 & 77.01 &  & 77.4 & 77.666 & 78.384 & 77.871 &  & 79.799 & 68.478 & 78.343 & 76.805 \\
& & Cov. [\%] & 99.959 & 99.959 & 99.918 & 99.979 & & 100.0 & 99.979 & 99.959 & 100.0 & & 100.0 & 99.979 & 99.938 & 99.918 & & 99.959 & 82.075 & 99.631 & 96.801 \\
\cline{2-22} \vspace{-2.5ex}\\
 & \multirow{2}{*}{Phi-4} & Acc. [\%] & 84.516 & 84.249 & 83.921 & 81.009 &  & 84.475 & 84.167 & 83.675 & 80.947 &  & 84.68 & 84.639 & 84.331 & 79.861 &  & 84.721 & 84.126 & 84.598 & 82.752 \\
& & Cov. [\%] & 100.0 & 100.0 & 100.0 & 97.99 & & 100.0 & 100.0 & 100.0 & 97.703 & & 100.0 & 100.0 & 100.0 & 96.534 & & 100.0 & 99.159 & 99.897 & 99.159 \\
\cline{2-22} \vspace{-2.5ex}\\
 & \multirow{2}{*}{Qwen-2-VL} & Acc. [\%] & 82.814 & 83.08 & 83.45 & 83.737 &  & 82.875 & 83.347 & 83.798 & 83.942 &  & 83.039 & 83.265 & 84.085 & 83.942 &  & 66.181 & 40.73 & 83.614 & 84.352 \\
& & Cov. [\%] & 100.0 & 100.0 & 99.856 & 99.959 & & 100.0 & 100.0 & 99.918 & 99.979 & & 100.0 & 100.0 & 99.918 & 99.959 & & 79.532 & 50.082 & 99.344 & 99.713 \\
\cline{2-22} \vspace{-2.5ex}\\
 & \multirow{2}{*}{Qwen-2.5-VL} & Acc. [\%] & \bfseries 88.084 & \bfseries 87.859 & \bfseries 86.485 & \bfseries 85.726 &  & \bfseries 87.756 & \bfseries 87.838 & \bfseries 86.751 & \bfseries 85.172 &  & \bfseries 87.982 & \bfseries 87.797 & \bfseries 86.526 & \bfseries 85.336 &  & \bfseries 87.961 & \bfseries 87.961 & \bfseries 87.059 & \bfseries 84.844 \\
& & Cov. [\%] & 100.0 & 100.0 & 99.938 & 100.0 & & 100.0 & 100.0 & 99.979 & 100.0 & & 100.0 & 100.0 & 99.918 & 100.0 & & 100.0 & 100.0 & 100.0 & 100.0 \\
\midrule
\multirow{14}{*}{Line break}
 & \multirow{2}{*}{Gemma-3} & Acc. [\%] & 77.728 & 77.789 & 78.363 & 77.933 &  & 77.974 & 78.199 & 78.445 & 78.445 &  & 77.564 & 78.056 & 78.384 & 77.81 &  & 78.24 & 78.486 & 78.486 & 78.281 \\
& & Cov. [\%] & 100.0 & 100.0 & 100.0 & 100.0 & & 100.0 & 100.0 & 100.0 & 100.0 & & 100.0 & 100.0 & 100.0 & 100.0 & & 100.0 & 100.0 & 100.0 & 100.0 \\
\cline{2-22} \vspace{-2.5ex}\\
 & \multirow{2}{*}{LLaVA-1.5} & Acc. [\%] & 71.862 & 67.966 & 70.468 & 70.324 &  & 72.129 & 69.463 & 70.468 & 69.914 &  & 71.965 & 68.786 & 70.652 & 70.221 &  & 71.473 & 14.171 & 68.724 & 34.454 \\
& & Cov. [\%] & 100.0 & 95.016 & 99.446 & 99.692 & & 100.0 & 97.949 & 99.754 & 99.836 & & 100.0 & 96.37 & 99.651 & 99.774 & & 99.979 & 15.997 & 97.067 & 38.495 \\
\cline{2-22} \vspace{-2.5ex}\\
 & \multirow{2}{*}{LLaVA-OV} & Acc. [\%] & 85.541 & 85.213 & 84.372 & 65.094 &  & 85.541 & 85.172 & 84.701 & 67.002 &  & 85.521 & 85.131 & 84.762 & 68.888 &  & 85.418 & 27.543 & 84.988 & 71.657 \\
& & Cov. [\%] & 100.0 & 99.918 & 99.672 & 99.036 & & 100.0 & 100.0 & 99.795 & 99.815 & & 100.0 & 99.979 & 99.774 & 99.836 & & 100.0 & 34.372 & 99.508 & 99.692 \\
\cline{2-22} \vspace{-2.5ex}\\
 & \multirow{2}{*}{Phi-3.5} & Acc. [\%] & 80.271 & 80.353 & 80.107 & 79.225 &  & 79.655 & 79.737 & 79.84 & 79.635 &  & 80.271 & 80.271 & 80.23 & 79.348 &  & 80.127 & 75.513 & 79.943 & 79.266 \\
& & Cov. [\%] & 100.0 & 100.0 & 99.979 & 99.344 & & 100.0 & 100.0 & 99.979 & 99.959 & & 100.0 & 100.0 & 100.0 & 99.262 & & 100.0 & 91.879 & 100.0 & 99.098 \\
\cline{2-22} \vspace{-2.5ex}\\
 & \multirow{2}{*}{Phi-4} & Acc. [\%] & 85.254 & 84.824 & 85.029 & 83.142 &  & 85.049 & 84.803 & 85.049 & 83.88 &  & 85.275 & 85.029 & 85.316 & 82.568 &  & 85.193 & 84.721 & 85.152 & 81.87 \\
& & Cov. [\%] & 100.0 & 100.0 & 100.0 & 99.118 & & 100.0 & 100.0 & 100.0 & 99.631 & & 100.0 & 100.0 & 100.0 & 98.605 & & 100.0 & 99.877 & 99.815 & 97.785 \\
\cline{2-22} \vspace{-2.5ex}\\
 & \multirow{2}{*}{Qwen-2-VL} & Acc. [\%] & 85.459 & 85.398 & 84.598 & 84.578 &  & 85.562 & 85.48 & 84.475 & 85.131 &  & 85.295 & 85.398 & 84.639 & 84.967 &  & 79.081 & 44.688 & 84.249 & 85.234 \\
& & Cov. [\%] & 100.0 & 100.0 & 99.918 & 100.0 & & 100.0 & 100.0 & 99.918 & 100.0 & & 100.0 & 100.0 & 99.918 & 100.0 & & 93.007 & 54.614 & 99.795 & 99.918 \\
\cline{2-22} \vspace{-2.5ex}\\
 & \multirow{2}{*}{Qwen-2.5-VL} & Acc. [\%] & \bfseries 87.879 & \bfseries 87.879 & \bfseries 87.244 & \bfseries 85.993 &  & \bfseries 88.023 & \bfseries 88.084 & \bfseries 87.326 & \bfseries 86.157 &  & \bfseries 87.838 & \bfseries 88.043 & \bfseries 87.428 & \bfseries 85.993 &  & \bfseries 87.961 & \bfseries 87.9 & \bfseries 87.346 & \bfseries 85.336 \\
& & Cov. [\%] & 100.0 & 100.0 & 99.959 & 100.0 & & 100.0 & 100.0 & 99.938 & 100.0 & & 100.0 & 100.0 & 99.979 & 100.0 & & 100.0 & 100.0 & 100.0 & 100.0 \\
\midrule
\multirow{14}{*}{Semicolon}
 & \multirow{2}{*}{Gemma-3} & Acc. [\%] & 77.194 & 77.482 & 77.584 & 77.03 &  & 77.256 & 77.584 & 77.687 & 77.338 &  & 77.174 & 77.441 & 77.687 & 77.051 &  & 77.83 & 77.912 & 77.605 & 77.441 \\
& & Cov. [\%] & 100.0 & 100.0 & 100.0 & 100.0 & & 100.0 & 100.0 & 100.0 & 100.0 & & 100.0 & 100.0 & 100.0 & 100.0 & & 100.0 & 100.0 & 100.0 & 100.0 \\
\cline{2-22} \vspace{-2.5ex}\\
 & \multirow{2}{*}{LLaVA-1.5} & Acc. [\%] & 69.381 & 62.592 & 67.494 & 67.617 &  & 69.155 & 64.889 & 67.576 & 66.407 &  & 69.011 & 65.094 & 68.048 & 66.94 &  & 69.647 & 17.248 & 67.658 & 50.738 \\
& & Cov. [\%] & 99.918 & 90.976 & 99.528 & 99.467 & & 99.959 & 95.16 & 99.795 & 99.733 & & 99.918 & 94.606 & 99.754 & 99.569 & & 99.897 & 20.427 & 99.405 & 63.208 \\
\cline{2-22} \vspace{-2.5ex}\\
 & \multirow{2}{*}{LLaVA-OV} & Acc. [\%] & 84.844 & 84.516 & 84.208 & 78.158 &  & 85.418 & 85.029 & 84.721 & 75.738 &  & 85.172 & 84.947 & 84.803 & 77.748 &  & 85.5 & 12.387 & 84.885 & 74.569 \\
& & Cov. [\%] & 99.979 & 99.938 & 99.692 & 99.036 & & 100.0 & 100.0 & 99.815 & 99.733 & & 100.0 & 100.0 & 99.795 & 99.754 & & 100.0 & 14.971 & 99.487 & 99.631 \\
\cline{2-22} \vspace{-2.5ex}\\
 & \multirow{2}{*}{Phi-3.5} & Acc. [\%] & 77.4 & 77.687 & 78.507 & 78.589 &  & 78.917 & 78.856 & 78.548 & 77.974 &  & 78.384 & 78.671 & 78.876 & 78.876 &  & 79.43 & 69.299 & 78.63 & 77.625 \\
& & Cov. [\%] & 100.0 & 100.0 & 99.959 & 99.938 & & 100.0 & 100.0 & 99.979 & 99.979 & & 100.0 & 100.0 & 99.979 & 99.897 & & 99.959 & 83.388 & 99.61 & 97.703 \\
\cline{2-22} \vspace{-2.5ex}\\
 & \multirow{2}{*}{Phi-4} & Acc. [\%] & 84.229 & 84.167 & 83.511 & 81.296 &  & 84.516 & 83.778 & 83.655 & 81.317 &  & 84.516 & 84.311 & 84.311 & 80.209 &  & 84.27 & 84.024 & 84.393 & 83.183 \\
& & Cov. [\%] & 100.0 & 100.0 & 100.0 & 98.052 & & 100.0 & 100.0 & 100.0 & 97.58 & & 100.0 & 100.0 & 100.0 & 96.37 & & 100.0 & 98.954 & 99.897 & 99.364 \\
\cline{2-22} \vspace{-2.5ex}\\
 & \multirow{2}{*}{Qwen-2-VL} & Acc. [\%] & 83.511 & 84.619 & 83.962 & 84.475 &  & 84.413 & 84.701 & 84.065 & 84.27 &  & 84.906 & 84.803 & 84.085 & 84.66 &  & 73.4 & 42.35 & 83.573 & 84.598 \\
& & Cov. [\%] & 100.0 & 100.0 & 99.918 & 99.938 & & 100.0 & 100.0 & 99.918 & 99.959 & & 100.0 & 100.0 & 99.897 & 99.938 & & 87.428 & 52.092 & 99.364 & 99.877 \\
\cline{2-22} \vspace{-2.5ex}\\
 & \multirow{2}{*}{Qwen-2.5-VL} & Acc. [\%] & \bfseries 88.084 & \bfseries 87.941 & \bfseries 86.813 & \bfseries 85.993 &  & \bfseries 88.064 & \bfseries 88.023 & \bfseries 86.874 & \bfseries 85.849 &  & \bfseries 88.043 & \bfseries 87.961 & \bfseries 86.916 & \bfseries 85.685 &  & \bfseries 87.982 & \bfseries 87.777 & \bfseries 87.162 & \bfseries 84.947 \\
& & Cov. [\%] & 100.0 & 100.0 & 99.938 & 100.0 & & 100.0 & 100.0 & 100.0 & 100.0 & & 100.0 & 100.0 & 100.0 & 100.0 & & 100.0 & 100.0 & 100.0 & 100.0 \\
\bottomrule
\end{tabular}%
}
 \end{table*}

%% file: tables/full_results_mme_rw_lite.tex
\begin{table*}[hbtp]
\centering
    \caption{\textbf{Full results of prompt format variations on MME-RealWorld-Lite dataset.} 
    This table presents the performance of seven \acp{MLLM} under different option separators (\textit{Comma}, \textit{Line break}, \textit{Semicolon}) and delimiters (\textit{Dot}, \textit{Colon}, \textit{Bracket}, \textit{Double brackets}).
    Option ID sets are varied across four formats: \textit{Uppercase}, \textit{Lowercase}, \textit{Numbers}, and \textit{Roman numbers}.
    Metrics include accuracy (\%) and coverage (\%), with coverage consistently near 100\% and accuracy varying by format.
    The top-performing model changes across prompt format variations, alternating between Qwen-2.5-VL, Qwen-2-VL, and Phi-4.
    The accuracy gap to the second-best \ac{MLLM} ranges from $0.08$ \ac{pp} to $3.00$ \ac{pp}.
    The highest accuracy per prompt format variation is highlighted in bold.}
    \label{tab:full_results_mme}
 \resizebox{\textwidth}{!}{%
 \begin{tabular}{l l l S S S S @{} c S S S S @{} c S S S S @{} c S S S S}
\toprule
& & & \multicolumn{19}{c}{Option delimiter} \\
\cline{4-22} \vspace{-2.5ex}\\
& & & \multicolumn{4}{c}{\textbf{Dot}} & & \multicolumn{4}{c}{\textbf{Colon}} & & \multicolumn{4}{c}{\textbf{Bracket}} & & \multicolumn{4}{c}{\textbf{Double brackets}} \\
& & & \multicolumn{4}{c}{Option IDs} & & \multicolumn{4}{c}{Option IDs} & & \multicolumn{4}{c}{Option IDs} & & \multicolumn{4}{c}{Option IDs}\\
\cline{4-7} \cline{9-12} \cline{14-17} \cline{19-22} \vspace{-2.5ex} \\
\textbf{Option sep.} & \textbf{Model} & & \textbf{Upp.} & \textbf{Low.} & \textbf{Num.} & \textbf{Rom.}
 & & \textbf{Upp.} & \textbf{Low.} & \textbf{Num.} & \textbf{Rom.}
 & & \textbf{Upp.} & \textbf{Low.} & \textbf{Num.} & \textbf{Rom.}
 & & \textbf{Upp.} & \textbf{Low.} & \textbf{Num.} & \textbf{Rom.} \\
\midrule
\multirow{14}{*}{Comma}
 & \multirow{2}{*}{Gemma-3} & Acc. [\%] & 35.967 & 34.237 & 35.279 & 35.779 &  & 36.227 & 34.101 & 35.331 & 35.331 &  & 35.852 & 32.84 & 35.039 & 35.873 &  & 36.113 & 30.703 & 35.696 & 35.706 \\
& & Cov. [\%] & 100.0 & 100.0 & 99.969 & 100.0 & & 100.0 & 100.0 & 99.99 & 100.0 & & 100.0 & 100.0 & 99.937 & 100.0 & & 100.0 & 99.99 & 99.979 & 100.0 \\
\cline{2-22} \vspace{-2.5ex}\\
 & \multirow{2}{*}{LLaVA-1.5} & Acc. [\%] & 30.443 & 26.618 & 28.869 & 29.026 &  & 29.713 & 26.107 & 28.734 & 28.233 &  & 30.787 & 27.139 & 29.713 & 29.88 &  & 30.943 & 8.504 & 29.724 & 10.703 \\
& & Cov. [\%] & 100.0 & 97.978 & 99.531 & 99.739 & & 100.0 & 99.333 & 99.844 & 99.917 & & 100.0 & 98.749 & 99.687 & 99.812 & & 100.0 & 30.391 & 99.416 & 28.254 \\
\cline{2-22} \vspace{-2.5ex}\\
 & \multirow{2}{*}{LLaVA-OV} & Acc. [\%] & 44.773 & 44.721 & 45.138 & 34.768 &  & 46.18 & 45.513 & 45.545 & 33.08 &  & 46.201 & 45.034 & 45.086 & 33.986 &  & 45.305 & 3.929 & 44.659 & 32.298 \\
& & Cov. [\%] & 100.0 & 100.0 & 99.75 & 99.187 & & 100.0 & 100.0 & 99.948 & 99.489 & & 100.0 & 100.0 & 99.885 & 99.604 & & 100.0 & 9.526 & 98.603 & 98.437 \\
\cline{2-22} \vspace{-2.5ex}\\
 & \multirow{2}{*}{Phi-3.5} & Acc. [\%] & 37.342 & 35.143 & 39.145 & 39.396 &  & 39.25 & 36.206 & 39.604 & 39.5 &  & 37.999 & 36.331 & 39.656 & 39.02 &  & 37.665 & 26.764 & 40.25 & 38.666 \\
& & Cov. [\%] & 100.0 & 100.0 & 99.979 & 100.0 & & 100.0 & 100.0 & 99.99 & 100.0 & & 100.0 & 100.0 & 100.0 & 100.0 & & 99.958 & 88.525 & 99.917 & 99.396 \\
\cline{2-22} \vspace{-2.5ex}\\
 & \multirow{2}{*}{Phi-4} & Acc. [\%] & \bfseries 47.514 & \bfseries 47.817 & 45.492 & 44.19 &  & 47.66 & \bfseries 47.483 & 45.659 & 45.294 &  & \bfseries 47.514 & \bfseries 47.671 & 45.607 & 42.824 &  & \bfseries 46.899 & \bfseries 45.805 & 46.128 & 43.231 \\
& & Cov. [\%] & 100.0 & 100.0 & 100.0 & 99.323 & & 100.0 & 100.0 & 100.0 & 99.343 & & 100.0 & 100.0 & 100.0 & 98.781 & & 100.0 & 99.197 & 99.854 & 99.385 \\
\cline{2-22} \vspace{-2.5ex}\\
 & \multirow{2}{*}{Qwen-2-VL} & Acc. [\%] & 44.273 & 41.959 & 45.92 & 45.43 &  & 44.617 & 42.303 & 46.17 & \bfseries 46.347 &  & 44.982 & 42.96 & 46.139 & \bfseries 45.69 &  & 42.637 & 33.236 & 45.701 & \bfseries 46.587 \\
& & Cov. [\%] & 100.0 & 99.99 & 99.729 & 100.0 & & 100.0 & 100.0 & 99.781 & 100.0 & & 100.0 & 100.0 & 99.875 & 100.0 & & 99.364 & 91.683 & 99.427 & 100.0 \\
\cline{2-22} \vspace{-2.5ex}\\
 & \multirow{2}{*}{Qwen-2.5-VL} & Acc. [\%] & 47.212 & 44.815 & \bfseries 47.483 & \bfseries 46.024 &  & \bfseries 47.744 & 45.169 & \bfseries 47.514 & 45.763 &  & 46.837 & 44.523 & \bfseries 47.316 & 45.513 &  & 46.514 & 44.419 & \bfseries 47.285 & 44.304 \\
& & Cov. [\%] & 100.0 & 100.0 & 99.896 & 100.0 & & 100.0 & 100.0 & 99.896 & 100.0 & & 100.0 & 100.0 & 99.948 & 100.0 & & 100.0 & 100.0 & 99.948 & 99.99 \\
\midrule
\multirow{14}{*}{Line break}
 & \multirow{2}{*}{Gemma-3} & Acc. [\%] & 35.915 & 33.226 & 35.279 & 36.123 &  & 35.779 & 32.35 & 35.581 & 35.383 &  & 35.821 & 31.86 & 35.685 & 36.165 &  & 36.092 & 29.651 & 35.498 & 36.217 \\
& & Cov. [\%] & 100.0 & 100.0 & 99.937 & 100.0 & & 100.0 & 100.0 & 99.948 & 100.0 & & 100.0 & 100.0 & 99.948 & 100.0 & & 100.0 & 100.0 & 99.948 & 100.0 \\
\cline{2-22} \vspace{-2.5ex}\\
 & \multirow{2}{*}{LLaVA-1.5} & Acc. [\%] & 32.527 & 29.38 & 33.153 & 33.757 &  & 31.704 & 28.765 & 33.403 & 32.798 &  & 32.944 & 29.203 & 33.674 & 34.466 &  & 33.132 & 6.483 & 31.965 & 4.763 \\
& & Cov. [\%] & 100.0 & 99.521 & 99.458 & 99.979 & & 100.0 & 99.646 & 99.865 & 99.99 & & 100.0 & 99.635 & 99.708 & 99.99 & & 99.99 & 23.439 & 99.031 & 10.464 \\
\cline{2-22} \vspace{-2.5ex}\\
 & \multirow{2}{*}{LLaVA-OV} & Acc. [\%] & 44.534 & 44.158 & 43.856 & 29.588 &  & 45.482 & 44.961 & 44.919 & 30.12 &  & 45.43 & 44.575 & 44.732 & 30.422 &  & 45.232 & 22.929 & 44.773 & 31.61 \\
& & Cov. [\%] & 100.0 & 100.0 & 98.437 & 99.041 & & 100.0 & 100.0 & 99.854 & 99.531 & & 100.0 & 100.0 & 99.802 & 99.521 & & 100.0 & 54.81 & 99.427 & 99.375 \\
\cline{2-22} \vspace{-2.5ex}\\
 & \multirow{2}{*}{Phi-3.5} & Acc. [\%] & 38.947 & 34.643 & 41.115 & 39.239 &  & 39.01 & 34.851 & 41.251 & 39.156 &  & 38.343 & 35.3 & 40.584 & 38.437 &  & 38.061 & 27.827 & 40.855 & 38.562 \\
& & Cov. [\%] & 100.0 & 100.0 & 100.0 & 100.0 & & 100.0 & 100.0 & 100.0 & 100.0 & & 100.0 & 100.0 & 100.0 & 100.0 & & 100.0 & 91.006 & 99.99 & 99.833 \\
\cline{2-22} \vspace{-2.5ex}\\
 & \multirow{2}{*}{Phi-4} & Acc. [\%] & \bfseries 47.431 & \bfseries 46.952 & 46.67 & 44.7 &  & \bfseries 47.702 & \bfseries 47.056 & 46.378 & \bfseries 46.076 &  & \bfseries 47.087 & \bfseries 47.077 & 46.733 & 43.908 &  & \bfseries 47.358 & \bfseries 46.305 & 46.993 & 43.669 \\
& & Cov. [\%] & 100.0 & 100.0 & 100.0 & 99.75 & & 100.0 & 100.0 & 100.0 & 99.844 & & 100.0 & 100.0 & 100.0 & 99.5 & & 100.0 & 99.917 & 99.969 & 99.031 \\
\cline{2-22} \vspace{-2.5ex}\\
 & \multirow{2}{*}{Qwen-2-VL} & Acc. [\%] & 45.18 & 44.304 & 46.555 & 45.117 &  & 45.795 & 44.825 & 46.451 & 45.972 &  & 45.909 & 45.336 & 46.587 & \bfseries 45.711 &  & 45.065 & 32.715 & 46.483 & \bfseries 46.597 \\
& & Cov. [\%] & 100.0 & 100.0 & 99.802 & 99.99 & & 100.0 & 100.0 & 99.896 & 100.0 & & 100.0 & 100.0 & 99.865 & 100.0 & & 98.916 & 86.795 & 99.562 & 99.948 \\
\cline{2-22} \vspace{-2.5ex}\\
 & \multirow{2}{*}{Qwen-2.5-VL} & Acc. [\%] & 46.17 & 44.409 & \bfseries 47.91 & \bfseries 45.461 &  & 46.774 & 44.857 & \bfseries 47.848 & 45.409 &  & 46.024 & 44.419 & \bfseries 47.546 & 45.076 &  & 46.545 & 44.304 & \bfseries 47.223 & 44.283 \\
& & Cov. [\%] & 100.0 & 100.0 & 99.969 & 100.0 & & 100.0 & 100.0 & 99.979 & 100.0 & & 100.0 & 100.0 & 99.969 & 100.0 & & 100.0 & 100.0 & 99.99 & 99.99 \\
\midrule
\multirow{14}{*}{Semicolon}
 & \multirow{2}{*}{Gemma-3} & Acc. [\%] & 36.04 & 34.153 & 35.175 & 35.779 &  & 36.092 & 34.07 & 35.248 & 35.289 &  & 36.081 & 32.903 & 35.539 & 35.664 &  & 36.092 & 30.735 & 35.654 & 36.186 \\
& & Cov. [\%] & 100.0 & 100.0 & 99.99 & 100.0 & & 100.0 & 100.0 & 99.99 & 100.0 & & 100.0 & 100.0 & 99.99 & 100.0 & & 100.0 & 100.0 & 99.979 & 100.0 \\
\cline{2-22} \vspace{-2.5ex}\\
 & \multirow{2}{*}{LLaVA-1.5} & Acc. [\%] & 32.006 & 28.629 & 30.38 & 31.277 &  & 30.891 & 27.608 & 31.068 & 30.568 &  & 32.256 & 28.765 & 31.371 & 31.704 &  & 31.777 & 10.276 & 30.474 & 12.142 \\
& & Cov. [\%] & 100.0 & 98.822 & 99.666 & 99.854 & & 100.0 & 99.5 & 99.927 & 99.937 & & 100.0 & 99.312 & 99.802 & 99.927 & & 100.0 & 36.092 & 99.562 & 32.527 \\
\cline{2-22} \vspace{-2.5ex}\\
 & \multirow{2}{*}{LLaVA-OV} & Acc. [\%] & 43.898 & 44.179 & 44.429 & 34.247 &  & 45.836 & 45.211 & 45.044 & 32.965 &  & 45.722 & 44.836 & 44.784 & 33.57 &  & 44.982 & 8.254 & 44.669 & 31.808 \\
& & Cov. [\%] & 100.0 & 100.0 & 99.896 & 98.729 & & 100.0 & 100.0 & 99.969 & 99.468 & & 100.0 & 100.0 & 99.948 & 99.583 & & 100.0 & 20.604 & 98.916 & 99.145 \\
\cline{2-22} \vspace{-2.5ex}\\
 & \multirow{2}{*}{Phi-3.5} & Acc. [\%] & 37.551 & 34.997 & 40.521 & 40.448 &  & 39.552 & 36.092 & 40.73 & 39.844 &  & 37.968 & 36.123 & 40.667 & 39.75 &  & 37.884 & 26.712 & 40.563 & 39.104 \\
& & Cov. [\%] & 100.0 & 100.0 & 100.0 & 100.0 & & 100.0 & 100.0 & 99.99 & 100.0 & & 100.0 & 100.0 & 99.99 & 100.0 & & 99.969 & 89.057 & 99.948 & 99.698 \\
\cline{2-22} \vspace{-2.5ex}\\
 & \multirow{2}{*}{Phi-4} & Acc. [\%] & 47.285 & \bfseries 47.493 & 46.097 & 44.825 &  & 47.525 & \bfseries 47.181 & 45.816 & 45.336 &  & \bfseries 47.514 & \bfseries 48.067 & 46.357 & 43.418 &  & 47.254 & \bfseries 45.899 & 46.806 & 44.044 \\
& & Cov. [\%] & 100.0 & 100.0 & 100.0 & 99.156 & & 100.0 & 100.0 & 100.0 & 99.031 & & 100.0 & 100.0 & 100.0 & 98.364 & & 100.0 & 99.104 & 99.917 & 99.312 \\
\cline{2-22} \vspace{-2.5ex}\\
 & \multirow{2}{*}{Qwen-2-VL} & Acc. [\%] & 44.648 & 42.96 & 46.139 & 46.045 &  & 45.232 & 43.137 & 46.024 & \bfseries 46.337 &  & 45.753 & 43.908 & 46.378 & \bfseries 46.066 &  & 43.794 & 34.466 & 45.816 & \bfseries 46.399 \\
& & Cov. [\%] & 100.0 & 100.0 & 99.917 & 100.0 & & 100.0 & 100.0 & 99.917 & 100.0 & & 100.0 & 100.0 & 99.958 & 100.0 & & 99.573 & 92.725 & 99.573 & 99.99 \\
\cline{2-22} \vspace{-2.5ex}\\
 & \multirow{2}{*}{Qwen-2.5-VL} & Acc. [\%] & \bfseries 47.514 & 45.586 & \bfseries 47.619 & \bfseries 46.243 &  & \bfseries 48.035 & 45.816 & \bfseries 47.671 & 46.087 &  & 47.316 & 45.461 & \bfseries 47.577 & 45.618 &  & \bfseries 47.4 & 45.67 & \bfseries 47.077 & 44.669 \\
& & Cov. [\%] & 100.0 & 100.0 & 99.927 & 100.0 & & 100.0 & 100.0 & 99.906 & 100.0 & & 100.0 & 100.0 & 99.958 & 100.0 & & 100.0 & 100.0 & 99.948 & 99.99 \\
\bottomrule
\end{tabular}%
}
\end{table*}

%% file: tables/full_results_vstar.tex
\begin{table*}[hbtp]
\centering
    \caption{\textbf{Full results of prompt format variations on V*Bench dataset.} 
    This table summarizes the performance of seven \acp{MLLM} under different option separators (\textit{Comma}, \textit{Line break}, \textit{Semicolon}) and delimiters (\textit{Dot}, \textit{Colon}, \textit{Bracket}, \textit{Double brackets}).
    Option ID sets are varied across four formats: \textit{Uppercase}, \textit{Lowercase}, \textit{Numbers}, and \textit{Roman numbers}.
    Metrics include accuracy (\%) and coverage (\%), with coverage consistently near 100\% and accuracy varying by format.
    Qwen-2.5-VL is consistently the top-performing \ac{MLLM}, demonstrating strong robustness.
    The accuracy gap to the second-best \ac{MLLM} ranges from $1.01$ \ac{pp} to $9.56$ \ac{pp}.
    The highest accuracy per prompt format variation is highlighted in bold.}
    \label{tab:full_results_vstar}
    \resizebox{\textwidth}{!}{%
    \begin{tabular}{l l l S S S S @{} c S S S S @{} c S S S S @{} c S S S S}
    \toprule
    & & & \multicolumn{19}{c}{Option delimiter} \\
    \cline{4-22} \vspace{-2.5ex}\\
    & & & \multicolumn{4}{c}{\textbf{Dot}} & & \multicolumn{4}{c}{\textbf{Colon}} & & \multicolumn{4}{c}{\textbf{Bracket}} & & \multicolumn{4}{c}{\textbf{Double brackets}} \\
    & & & \multicolumn{4}{c}{Option IDs} & & \multicolumn{4}{c}{Option IDs} & & \multicolumn{4}{c}{Option IDs} & & \multicolumn{4}{c}{Option IDs}\\
    \cline{4-7} \cline{9-12} \cline{14-17} \cline{19-22} \vspace{-2.5ex} \\
    \textbf{Option sep.} & \textbf{Model} & & \textbf{Upp.} & \textbf{Low.} & \textbf{Num.} & \textbf{Rom.}
     & & \textbf{Upp.} & \textbf{Low.} & \textbf{Num.} & \textbf{Rom.}
     & & \textbf{Upp.} & \textbf{Low.} & \textbf{Num.} & \textbf{Rom.}
     & & \textbf{Upp.} & \textbf{Low.} & \textbf{Num.} & \textbf{Rom.} \\
    \midrule
    \multirow{14}{*}{Comma}
     & \multirow{2}{*}{Gemma-3} & Acc. [\%] & 35.403 & 34.564 & 35.738 & 32.215 &  & 34.228 & 34.899 & 34.396 & 35.235 &  & 33.893 & 33.389 & 35.067 & 33.389 &  & 34.228 & 34.899 & 34.06 & 33.557 \\
    & & Cov. [\%] & 100.0 & 100.0 & 100.0 & 100.0 & & 100.0 & 100.0 & 100.0 & 100.0 & & 100.0 & 100.0 & 100.0 & 100.0 & & 100.0 & 100.0 & 100.0 & 100.0 \\
    \cline{2-22} \vspace{-2.5ex}\\
     & \multirow{2}{*}{LLaVA-1.5} & Acc. [\%] & 37.584 & 15.268 & 35.067 & 35.067 &  & 39.43 & 20.134 & 36.745 & 35.403 &  & 37.248 & 23.658 & 36.913 & 35.235 &  & 38.591 & 0.336 & 37.081 & 16.443 \\
    & & Cov. [\%] & 100.0 & 50.671 & 100.0 & 100.0 & & 100.0 & 63.087 & 100.0 & 100.0 & & 100.0 & 68.96 & 100.0 & 100.0 & & 100.0 & 0.503 & 100.0 & 32.383 \\
    \cline{2-22} \vspace{-2.5ex}\\
     & \multirow{2}{*}{LLaVA-OV} & Acc. [\%] & 72.651 & 69.799 & 70.302 & 52.013 &  & 71.812 & 70.805 & 73.49 & 45.805 &  & 72.987 & 71.141 & 73.826 & 48.49 &  & 73.658 & 0.671 & 72.987 & 44.966 \\
    & & Cov. [\%] & 100.0 & 100.0 & 100.0 & 97.315 & & 100.0 & 100.0 & 100.0 & 98.993 & & 100.0 & 100.0 & 100.0 & 99.497 & & 100.0 & 1.174 & 100.0 & 97.651 \\
    \cline{2-22} \vspace{-2.5ex}\\
     & \multirow{2}{*}{Phi-3.5} & Acc. [\%] & 46.309 & 47.483 & 46.141 & 45.973 &  & 49.664 & 48.993 & 47.987 & 46.812 &  & 48.322 & 48.49 & 46.644 & 45.638 &  & 49.497 & 30.034 & 46.644 & 46.309 \\
    & & Cov. [\%] & 100.0 & 100.0 & 100.0 & 100.0 & & 100.0 & 100.0 & 100.0 & 100.0 & & 100.0 & 100.0 & 100.0 & 100.0 & & 99.832 & 59.899 & 99.664 & 98.993 \\
    \cline{2-22} \vspace{-2.5ex}\\
     & \multirow{2}{*}{Phi-4} & Acc. [\%] & 72.315 & 72.315 & 70.302 & 60.067 &  & 73.322 & 72.315 & 71.309 & 62.416 &  & 72.819 & 73.154 & 70.805 & 59.228 &  & 72.651 & 72.651 & 71.477 & 58.221 \\
    & & Cov. [\%] & 100.0 & 100.0 & 100.0 & 93.624 & & 100.0 & 100.0 & 100.0 & 92.114 & & 100.0 & 100.0 & 100.0 & 92.785 & & 100.0 & 100.0 & 100.0 & 91.275 \\
    \cline{2-22} \vspace{-2.5ex}\\
     & \multirow{2}{*}{Qwen-2-VL} & Acc. [\%] & 70.134 & 70.805 & 71.141 & 69.295 &  & 69.799 & 70.302 & 70.638 & 69.966 &  & 69.631 & 70.47 & 70.638 & 68.96 &  & 67.617 & 35.235 & 70.47 & 70.973 \\
    & & Cov. [\%] & 100.0 & 100.0 & 100.0 & 100.0 & & 100.0 & 100.0 & 100.0 & 100.0 & & 100.0 & 100.0 & 100.0 & 100.0 & & 94.966 & 57.886 & 100.0 & 100.0 \\
    \cline{2-22} \vspace{-2.5ex}\\
     & \multirow{2}{*}{Qwen-2.5-VL} & Acc. [\%] & \bfseries 78.356 & \bfseries 78.356 & \bfseries 74.497 & \bfseries 74.329 &  & \bfseries 78.523 & \bfseries 78.188 & \bfseries 75.168 & \bfseries 73.49 &  & \bfseries 77.181 & \bfseries 78.02 & \bfseries 75.671 & \bfseries 74.161 &  & \bfseries 77.852 & \bfseries 79.53 & \bfseries 73.993 & \bfseries 72.651 \\
    & & Cov. [\%] & 100.0 & 100.0 & 98.154 & 100.0 & & 100.0 & 100.0 & 98.49 & 100.0 & & 100.0 & 100.0 & 98.322 & 100.0 & & 100.0 & 100.0 & 97.987 & 100.0 \\
    \midrule
    \multirow{14}{*}{Line break}
     & \multirow{2}{*}{Gemma-3} & Acc. [\%] & 34.899 & 33.725 & 33.893 & 32.55 &  & 35.235 & 35.738 & 34.228 & 34.228 &  & 34.228 & 33.389 & 34.564 & 32.215 &  & 33.221 & 33.054 & 34.06 & 33.893 \\
    & & Cov. [\%] & 100.0 & 100.0 & 100.0 & 100.0 & & 100.0 & 100.0 & 100.0 & 100.0 & & 100.0 & 100.0 & 100.0 & 100.0 & & 100.0 & 100.0 & 100.0 & 100.0 \\
    \cline{2-22} \vspace{-2.5ex}\\
     & \multirow{2}{*}{LLaVA-1.5} & Acc. [\%] & 38.423 & 22.819 & 37.081 & 37.416 &  & 37.248 & 33.221 & 36.577 & 37.248 &  & 38.758 & 30.201 & 36.409 & 37.248 &  & 37.752 & 0.168 & 36.074 & 11.074 \\
    & & Cov. [\%] & 100.0 & 69.799 & 99.664 & 100.0 & & 100.0 & 91.611 & 100.0 & 100.0 & & 100.0 & 85.403 & 99.832 & 100.0 & & 100.0 & 0.168 & 99.664 & 21.98 \\
    \cline{2-22} \vspace{-2.5ex}\\
     & \multirow{2}{*}{LLaVA-OV} & Acc. [\%] & 73.993 & 73.993 & 73.658 & 37.584 &  & 74.832 & 73.993 & 74.161 & 38.087 &  & 74.161 & 74.329 & 74.329 & 39.43 &  & 73.993 & 42.953 & 73.322 & 39.43 \\
    & & Cov. [\%] & 100.0 & 100.0 & 100.0 & 95.47 & & 100.0 & 100.0 & 100.0 & 98.49 & & 100.0 & 100.0 & 100.0 & 98.322 & & 100.0 & 61.913 & 99.832 & 98.826 \\
    \cline{2-22} \vspace{-2.5ex}\\
     & \multirow{2}{*}{Phi-3.5} & Acc. [\%] & 51.007 & 49.497 & 49.497 & 48.322 &  & 50.839 & 48.826 & 48.993 & 47.148 &  & 50.336 & 48.658 & 50.168 & 46.98 &  & 50.0 & 44.966 & 48.49 & 48.49 \\
    & & Cov. [\%] & 100.0 & 100.0 & 100.0 & 100.0 & & 100.0 & 100.0 & 100.0 & 100.0 & & 100.0 & 100.0 & 100.0 & 100.0 & & 100.0 & 91.275 & 100.0 & 99.832 \\
    \cline{2-22} \vspace{-2.5ex}\\
     & \multirow{2}{*}{Phi-4} & Acc. [\%] & 74.329 & 74.161 & 73.154 & 68.624 &  & 73.826 & 72.987 & 72.819 & 70.973 &  & 73.993 & 73.826 & 72.651 & 66.275 &  & 73.658 & 72.987 & 71.98 & 59.899 \\
    & & Cov. [\%] & 100.0 & 100.0 & 100.0 & 98.826 & & 100.0 & 100.0 & 100.0 & 99.329 & & 100.0 & 100.0 & 100.0 & 98.154 & & 100.0 & 100.0 & 99.664 & 91.107 \\
    \cline{2-22} \vspace{-2.5ex}\\
     & \multirow{2}{*}{Qwen-2-VL} & Acc. [\%] & 73.826 & 73.826 & 72.651 & 72.483 &  & 73.322 & 73.826 & 72.315 & 73.993 &  & 73.993 & 73.49 & 72.483 & 72.483 &  & 73.322 & 42.953 & 71.141 & 72.819 \\
    & & Cov. [\%] & 100.0 & 100.0 & 100.0 & 100.0 & & 100.0 & 100.0 & 100.0 & 100.0 & & 100.0 & 100.0 & 100.0 & 100.0 & & 99.497 & 67.617 & 100.0 & 100.0 \\
    \cline{2-22} \vspace{-2.5ex}\\
     & \multirow{2}{*}{Qwen-2.5-VL} & Acc. [\%] & \bfseries 81.711 & \bfseries 80.872 & \bfseries 75.839 & \bfseries 73.826 &  & \bfseries 81.04 & \bfseries 81.376 & \bfseries 77.181 & \bfseries 75.671 &  & \bfseries 81.04 & \bfseries 80.872 & \bfseries 75.503 & \bfseries 73.826 &  & \bfseries 80.705 & \bfseries 80.705 & \bfseries 75.671 & \bfseries 74.832 \\
    & & Cov. [\%] & 100.0 & 100.0 & 98.658 & 100.0 & & 100.0 & 100.0 & 98.154 & 100.0 & & 100.0 & 100.0 & 98.658 & 100.0 & & 100.0 & 100.0 & 97.819 & 100.0 \\
    \midrule
    \multirow{14}{*}{Semicolon}
     & \multirow{2}{*}{Gemma-3} & Acc. [\%] & 35.235 & 35.067 & 36.242 & 34.564 &  & 35.403 & 35.403 & 35.57 & 34.732 &  & 35.403 & 34.564 & 35.067 & 35.067 &  & 34.228 & 34.564 & 34.899 & 33.557 \\
    & & Cov. [\%] & 100.0 & 100.0 & 100.0 & 100.0 & & 100.0 & 100.0 & 100.0 & 100.0 & & 100.0 & 100.0 & 100.0 & 100.0 & & 100.0 & 100.0 & 100.0 & 100.0 \\
    \cline{2-22} \vspace{-2.5ex}\\
     & \multirow{2}{*}{LLaVA-1.5} & Acc. [\%] & 37.416 & 21.141 & 35.738 & 37.752 &  & 37.919 & 30.872 & 36.745 & 36.074 &  & 38.255 & 31.879 & 35.906 & 36.745 &  & 37.584 & 0.336 & 36.074 & 16.611 \\
    & & Cov. [\%] & 100.0 & 64.597 & 100.0 & 100.0 & & 100.0 & 87.248 & 100.0 & 100.0 & & 100.0 & 85.57 & 100.0 & 100.0 & & 100.0 & 0.336 & 99.832 & 36.242 \\
    \cline{2-22} \vspace{-2.5ex}\\
     & \multirow{2}{*}{LLaVA-OV} & Acc. [\%] & 72.483 & 72.148 & 71.812 & 50.168 &  & 73.826 & 73.322 & 74.161 & 42.617 &  & 73.826 & 72.987 & 73.322 & 47.987 &  & 73.826 & 5.872 & 73.658 & 43.792 \\
    & & Cov. [\%] & 100.0 & 100.0 & 100.0 & 94.966 & & 100.0 & 100.0 & 100.0 & 98.658 & & 100.0 & 100.0 & 100.0 & 98.658 & & 100.0 & 8.893 & 100.0 & 98.154 \\
    \cline{2-22} \vspace{-2.5ex}\\
     & \multirow{2}{*}{Phi-3.5} & Acc. [\%] & 45.302 & 48.826 & 47.483 & 47.315 &  & 48.993 & 48.49 & 48.154 & 47.315 &  & 49.329 & 48.993 & 48.826 & 46.812 &  & 49.664 & 34.732 & 47.315 & 47.483 \\
    & & Cov. [\%] & 100.0 & 100.0 & 100.0 & 100.0 & & 100.0 & 100.0 & 100.0 & 100.0 & & 100.0 & 100.0 & 100.0 & 100.0 & & 100.0 & 68.624 & 99.664 & 99.832 \\
    \cline{2-22} \vspace{-2.5ex}\\
     & \multirow{2}{*}{Phi-4} & Acc. [\%] & 72.651 & 71.477 & 69.966 & 59.06 &  & 73.154 & 71.644 & 70.638 & 61.913 &  & 72.651 & 71.98 & 71.98 & 57.047 &  & 73.154 & 71.477 & 71.477 & 60.235 \\
    & & Cov. [\%] & 100.0 & 100.0 & 100.0 & 90.94 & & 100.0 & 100.0 & 100.0 & 91.107 & & 100.0 & 100.0 & 100.0 & 89.094 & & 100.0 & 100.0 & 100.0 & 92.282 \\
    \cline{2-22} \vspace{-2.5ex}\\
     & \multirow{2}{*}{Qwen-2-VL} & Acc. [\%] & 71.309 & 73.658 & 71.141 & 71.812 &  & 72.651 & 73.154 & 71.477 & 71.141 &  & 72.651 & 72.819 & 71.812 & 70.973 &  & 70.805 & 36.409 & 71.477 & 71.644 \\
    & & Cov. [\%] & 100.0 & 100.0 & 100.0 & 100.0 & & 100.0 & 100.0 & 100.0 & 100.0 & & 100.0 & 100.0 & 100.0 & 100.0 & & 97.987 & 60.067 & 100.0 & 100.0 \\
    \cline{2-22} \vspace{-2.5ex}\\
     & \multirow{2}{*}{Qwen-2.5-VL} & Acc. [\%] & \bfseries 80.201 & \bfseries 80.872 & \bfseries 75.0 & \bfseries 74.832 &  & \bfseries 80.201 & \bfseries 80.369 & \bfseries 75.503 & \bfseries 75.0 &  & \bfseries 80.369 & \bfseries 81.04 & \bfseries 75.336 & \bfseries 75.336 &  & \bfseries 81.04 & \bfseries 81.04 & \bfseries 76.007 & \bfseries 74.161 \\
    & & Cov. [\%] & 100.0 & 100.0 & 98.826 & 100.0 & & 100.0 & 100.0 & 98.322 & 100.0 & & 100.0 & 100.0 & 98.826 & 100.0 & & 100.0 & 99.832 & 98.154 & 100.0 \\
    \bottomrule
    \end{tabular}%
    }
\end{table*}

%% file: tables/tokenizer_overview.tex
\begin{table}[hbtp]
    \centering
    \caption{\textbf{Overview tokenizers.} We specify the tokenizer used in each \ac{MLLM} by its HuggingFace tokenizer class name.}
    \label{tab:overview_tokenizers}
    \resizebox{0.85\linewidth}{!}{
    \begin{tabular}{l c}
        \toprule
        \textbf{Model} & \textbf{HF tokenizer} \\
        \midrule
        Gemma-3 & \texttt{GemmaTokenizerFast} \\
        LLaVA-1.5 & \texttt{LlamaTokenizerFast} \\
        LLaVA-OV & \texttt{Qwen2TokenizerFast} \\
        Phi-3.5 & \texttt{LlamaTokenizerFast} \\
        Phi-4 & \texttt{GPT2TokenizerFast} \\
        Qwen2-VL & \texttt{Qwen2TokenizerFast} \\
        Qwen2.5-VL & \texttt{Qwen2TokenizerFast} \\
        \bottomrule
    \end{tabular}%
    }
\end{table}

%% file: tables/tokenization_prompt_formats.tex
\begin{table*}[t]
    \centering
    \caption{\textbf{Option ID–option delimiter tokenization overview}. We report the average token count produced by different tokenizers for each option ID–delimiter combination. Combinations whose average token count across all option IDs exceeds $2$ tokens are highlighted in \textbf{bold}. We additionally mark cases where at least one option ID is not encoded atomically as a single token as a \colorbox{cellgray}{gray cell}.}
    \label{tab:tokenization_encoding}
    \resizebox{\textwidth}{!}{
    \begin{tabular}{l S S S S @{} c S S S S @{} c S S S S @{} c S S S S}
        \toprule
        & \multicolumn{19}{c}{Option delimiter} \\
        \cline{2-20} \vspace{-2.5ex}\\
        & \multicolumn{4}{c}{\textbf{Dot}} & & \multicolumn{4}{c}{\textbf{Colon}} & & \multicolumn{4}{c}{\textbf{Bracket}} & & \multicolumn{4}{c}{\textbf{Double brackets}} \\
        & \multicolumn{4}{c}{Option IDs} & & \multicolumn{4}{c}{Option IDs} & & \multicolumn{4}{c}{Option IDs} & & \multicolumn{4}{c}{Option IDs}\\
        \cline{2-5} \cline{7-10} \cline{12-15} \cline{17-20} \vspace{-2.5ex} \\
        \textbf{Tokenizer} & \textbf{Upp.} & \textbf{Low.} & \textbf{Num.} & \textbf{Rom.} & & \textbf{Upp.} & \textbf{Low.} & \textbf{Num.} & \textbf{Rom.} & & \textbf{Upp.} & \textbf{Low.} & \textbf{Num.} & \textbf{Rom.} & & \textbf{Upp.} & \textbf{Low.} & \textbf{Num.} & \textbf{Rom.} \\
        \midrule
        \texttt{GemmaTokenizerFast} & 2 & 2 & 2 & 2 & & 2 & 2 & 2 & 2 & & 2 & 2 & 2 & 2 & & \bfseries 3 & \bfseries 3 & \bfseries 3 & \bfseries 3\\
        \texttt{GPT2TokenizerFast} & 2 & 2 & 2 & 2 & & 2 & 2 & 2 & 2 & & 2 & 2 & 2 & 2 & & \GrayNum{2} & \GrayNum{2} & \bfseries 3 & \GrayBoldNum{2.75}\\
        \texttt{LlamaTokenizerFast} & 2 & 2 & \bfseries 3 & 2 & & 2 & 2 & \bfseries 3 & 2 & & 2 & 2 & \bfseries 3 & 2 & & \bfseries 3 & \GrayBoldNum{3} & \bfseries 3 & \GrayBoldNum{3}\\
        \texttt{Qwen2TokenizerFast} & 2 & 2 & 2 & 2 & & 2 & 2 & 2 & 2 & & 2 & 2 & 2 & 2 & & \GrayNum{2} & \GrayNum{2} & \bfseries 3 & \GrayBoldNum{2.75}\\
        \bottomrule
    \end{tabular}%
    }
\end{table*}

%% file: tables/decoding_option_IDs.tex
\begin{table*}[b]
    \centering
    \caption{\textbf{Tokenization length for decoding option IDs.}
    The table reports the number of tokens required by each model to generate different option IDs during decoding.
    While most ID sets require a single token, some models (e.g., LLaVA-1.5 and Phi-3.5) require multiple tokens for numeric identifiers.
    This variation affects decoding behavior by increasing generation length and potentially weakening copy-style selection of option IDs.}
    \label{tab:decoding_token_length_option_ids}
    \resizebox{\textwidth}{!}{
    \begin{tabular}{l c c c c c c c c c c c c c c c c}
        \toprule
         & \multicolumn{4}{c}{\textbf{Uppercase}} & \multicolumn{4}{c}{\textbf{Lowercase}} & \multicolumn{4}{c}{\textbf{Numbers}} & \multicolumn{4}{c}{\textbf{Roman numbers}}\\
         \textbf{Model} & \texttt{"A"} & \texttt{"B"} & \texttt{"C"} & \texttt{"D"} & \texttt{"a"} & \texttt{"b"} & \texttt{"c"} & \texttt{"d"} & \texttt{"1"} & \texttt{"2"} & \texttt{"3"} & \texttt{"4"} & \texttt{"I"} & \texttt{"II"} & \texttt{"III"} & \texttt{"IV"}\\
         \midrule
        Gemma-3-4B & 1 & 1 & 1 & 1 & 1 & 1 & 1 & 1 & 1 & 1 & 1 & 1 & 1 & 1 & 1 & 1 \\
        LLaVA-1.5-7B & 1 & 1 & 1 & 1 & 1 & 1 & 1 & 1 & 2 & 2 & 2 & 2 & 1 & 1 & 1 & 1 \\
        LLaVA-OV-7B & 1 & 1 & 1 & 1 & 1 & 1 & 1 & 1 & 1 & 1 & 1 & 1 & 1 & 1 & 1 & 1 \\
        Phi-3.5 & 1 & 1 & 1 & 1 & 1 & 1 & 1 & 1 & 2 & 2 & 2 & 2 & 1 & 1 & 1 & 1 \\
        Phi-4 & 1 & 1 & 1 & 1 & 1 & 1 & 1 & 1 & 1 & 1 & 1 & 1 & 1 & 1 & 1 & 1 \\
        Qwen-2-VL-7B & 1 & 1 & 1 & 1 & 1 & 1 & 1 & 1 & 1 & 1 & 1 & 1 & 1 & 1 & 1 & 1 \\
        Qwen-2.5-VL-7B & 1 & 1 & 1 & 1 & 1 & 1 & 1 & 1 & 1 & 1 & 1 & 1 & 1 & 1 & 1 & 1 \\
         \bottomrule
    \end{tabular}%
    }
\end{table*}

%% file: tables/results_grounded_mc_vqa.tex
\begin{table*}[t]
    \centering
    \caption{\textbf{Full grounded \ac{MC-VQA} results.} We compare grounded \ac{MC-VQA} (indicated by $\dagger$) with standard \ac{MC-VQA} across all models. Grounded \ac{MC-VQA} does not improve evaluation robustness, as indicated by comparable or higher standard deviations.}
    \label{tab:full_results_grounded_mc_vqa}
    \resizebox{\textwidth}{!}{%
    \begin{tabular}{l l | cccc | cccc | cccc}
        \toprule
         & & \multicolumn{4}{c}{\textbf{A-OKVQA} - Acc. $[\%]$} & \multicolumn{4}{c}{\textbf{HRBench-4K} - Acc. $[\%]$} & \multicolumn{4}{c}{\textbf{V*Bench} - Acc. $[\%]$} \\
        \textbf{Model} & \textbf{Option IDs} & Mean & Std. & Min. & Max. & Mean & Std. & Min. & Max. & Mean & Std. & Min. & Max. \\
        \midrule
\multirow{8}{*}{\textbf{Gemma-3}} & Uppercase$^{\dagger}$ & $78.66$ & $0.26$ & $78.21$ & $79.13$ & $48.59$ & $0.71$ & $47.38$ & $50.12$ & $40.98$ & $0.41$ & $40.27$ & $41.61$ \\
 & \basevqa{Uppercase} & \basevqacell{78.15} & \basevqacell{0.38} & \basevqacell{77.49} & \basevqacell{78.62} & \basevqacell{45.80} & \basevqacell{0.74} & \basevqacell{44.25} & \basevqacell{46.88} & \basevqacell{34.63} & \basevqacell{0.72} & \basevqacell{33.22} & \basevqacell{35.40} \\
 & Lowercase$^{\dagger}$ & $78.76$ & $0.21$ & $78.49$ & $79.08$ & $48.42$ & $0.46$ & $47.88$ & $49.25$ & $40.84$ & $0.50$ & $40.10$ & $41.44$ \\
 & \basevqa{Lowercase} & \basevqacell{78.30} & \basevqacell{0.26} & \basevqacell{77.84} & \basevqacell{78.65} & \basevqacell{46.01} & \basevqacell{0.83} & \basevqacell{44.62} & \basevqacell{47.00} & \basevqacell{34.44} & \basevqacell{0.86} & \basevqacell{33.05} & \basevqacell{35.74} \\
 & Numbers$^{\dagger}$ & $78.59$ & $0.16$ & $78.28$ & $78.78$ & $47.77$ & $0.65$ & $46.62$ & $48.88$ & $41.02$ & $0.58$ & $40.10$ & $41.95$ \\
 & \basevqa{Numbers} & \basevqacell{77.89} & \basevqacell{0.30} & \basevqacell{77.40} & \basevqacell{78.30} & \basevqacell{45.15} & \basevqacell{0.76} & \basevqacell{43.50} & \basevqacell{46.38} & \basevqacell{34.82} & \basevqacell{0.75} & \basevqacell{33.89} & \basevqacell{36.24} \\
 & Roman$^{\dagger}$ & $78.32$ & $0.24$ & $77.77$ & $78.65$ & $48.00$ & $0.63$ & $47.00$ & $48.88$ & $40.03$ & $0.30$ & $39.60$ & $40.60$ \\
 & \basevqa{Roman} & \basevqacell{77.63} & \basevqacell{0.26} & \basevqacell{77.14} & \basevqacell{78.06} & \basevqacell{46.68} & \basevqacell{0.36} & \basevqacell{46.12} & \basevqacell{47.50} & \basevqacell{33.77} & \basevqacell{1.05} & \basevqacell{32.21} & \basevqacell{35.23} \\
\midrule
\multirow{8}{*}{\textbf{LLaVA-1.5}} & Uppercase$^{\dagger}$ & $76.35$ & $2.28$ & $72.16$ & $79.08$ & $35.73$ & $1.39$ & $33.00$ & $37.38$ & $40.38$ & $0.37$ & $39.60$ & $41.11$ \\
 & \basevqa{Uppercase} & \basevqacell{76.36} & \basevqacell{2.30} & \basevqacell{72.18} & \basevqacell{79.10} & \basevqacell{35.73} & \basevqacell{1.40} & \basevqacell{33.00} & \basevqacell{37.25} & \basevqacell{38.02} & \basevqacell{0.68} & \basevqacell{37.25} & \basevqacell{39.43} \\
 & Lowercase$^{\dagger}$ & $50.01$ & $25.19$ & $8.34$ & $76.07$ & $22.12$ & $13.57$ & $0.88$ & $33.50$ & $34.12$ & $8.58$ & $18.12$ & $40.77$ \\
 & \basevqa{Lowercase} & \basevqacell{50.32} & \basevqacell{25.43} & \basevqacell{8.28} & \basevqacell{76.05} & \basevqacell{23.15} & \basevqacell{13.40} & \basevqacell{0.88} & \basevqacell{34.12} & \basevqacell{19.17} & \basevqacell{12.58} & \basevqacell{0.17} & \basevqacell{33.22} \\
 & Numbers$^{\dagger}$ & $74.46$ & $2.27$ & $70.35$ & $77.36$ & $32.99$ & $1.50$ & $30.63$ & $35.12$ & $39.26$ & $1.04$ & $37.75$ & $41.11$ \\
 & \basevqa{Numbers} & \basevqacell{74.45} & \basevqacell{2.26} & \basevqacell{70.35} & \basevqacell{77.27} & \basevqacell{33.00} & \basevqacell{1.40} & \basevqacell{30.50} & \basevqacell{35.00} & \basevqacell{36.37} & \basevqacell{0.61} & \basevqacell{35.07} & \basevqacell{37.08} \\
 & Roman$^{\dagger}$ & $66.98$ & $13.10$ & $38.69$ & $77.53$ & $27.23$ & $11.21$ & $5.50$ & $35.38$ & $35.61$ & $7.09$ & $17.79$ & $40.60$ \\
 & \basevqa{Roman} & \basevqacell{67.20} & \basevqacell{13.21} & \basevqacell{38.60} & \basevqacell{77.58} & \basevqacell{27.74} & \basevqacell{10.77} & \basevqacell{5.62} & \basevqacell{35.50} & \basevqacell{31.03} & \basevqacell{9.97} & \basevqacell{11.07} & \basevqacell{37.75} \\
\midrule
\multirow{8}{*}{\textbf{LLaVA-OV}} & Uppercase$^{\dagger}$ & $90.64$ & $0.38$ & $89.85$ & $91.07$ & $64.78$ & $0.62$ & $64.00$ & $65.88$ & $73.13$ & $0.62$ & $71.98$ & $73.83$ \\
 & \basevqa{Uppercase} & \basevqacell{90.64} & \basevqacell{0.38} & \basevqacell{89.85} & \basevqacell{91.07} & \basevqacell{64.76} & \basevqacell{0.58} & \basevqacell{64.12} & \basevqacell{65.88} & \basevqacell{73.50} & \basevqacell{0.85} & \basevqacell{71.81} & \basevqacell{74.83} \\
 & Lowercase$^{\dagger}$ & $72.61$ & $33.31$ & $3.84$ & $90.92$ & $53.08$ & $22.51$ & $4.38$ & $65.88$ & $54.57$ & $32.85$ & $0.00$ & $73.99$ \\
 & \basevqa{Lowercase} & \basevqacell{72.61} & \basevqacell{33.31} & \basevqacell{3.82} & \basevqacell{90.90} & \basevqacell{53.06} & \basevqacell{22.53} & \basevqacell{4.38} & \basevqacell{65.88} & \basevqacell{58.50} & \basevqacell{27.20} & \basevqacell{0.67} & \basevqacell{74.33} \\
 & Numbers$^{\dagger}$ & $90.08$ & $0.26$ & $89.76$ & $90.55$ & $62.17$ & $1.17$ & $60.25$ & $63.62$ & $72.04$ & $0.84$ & $70.64$ & $73.15$ \\
 & \basevqa{Numbers} & \basevqacell{90.03} & \basevqacell{0.26} & \basevqacell{89.72} & \basevqacell{90.50} & \basevqacell{62.11} & \basevqacell{1.21} & \basevqacell{60.25} & \basevqacell{63.62} & \basevqacell{73.25} & \basevqacell{1.14} & \basevqacell{70.30} & \basevqacell{74.33} \\
 & Roman$^{\dagger}$ & $71.85$ & $6.90$ & $59.87$ & $80.76$ & $40.93$ & $4.97$ & $34.25$ & $47.75$ & $54.88$ & $4.46$ & $47.82$ & $62.08$ \\
 & \basevqa{Roman} & \basevqacell{71.86} & \basevqacell{6.92} & \basevqacell{59.85} & \basevqacell{80.72} & \basevqacell{40.92} & \basevqacell{4.97} & \basevqacell{34.25} & \basevqacell{47.75} & \basevqacell{44.20} & \basevqacell{4.88} & \basevqacell{37.58} & \basevqacell{52.01} \\
\midrule
\multirow{8}{*}{\textbf{Phi-3.5}} & Uppercase$^{\dagger}$ & $78.85$ & $1.20$ & $76.99$ & $79.83$ & $49.43$ & $1.40$ & $47.00$ & $51.00$ & $50.78$ & $0.85$ & $48.83$ & $52.01$ \\
 & \basevqa{Uppercase} & \basevqacell{79.69} & \basevqacell{1.25} & \basevqacell{76.92} & \basevqacell{81.07} & \basevqacell{49.15} & \basevqacell{1.42} & \basevqacell{46.88} & \basevqacell{50.75} & \basevqacell{49.11} & \basevqacell{1.72} & \basevqacell{45.30} & \basevqacell{51.01} \\
 & Lowercase$^{\dagger}$ & $77.90$ & $3.14$ & $72.40$ & $79.91$ & $47.12$ & $3.69$ & $39.12$ & $50.25$ & $48.24$ & $7.57$ & $29.53$ & $52.68$ \\
 & \basevqa{Lowercase} & \basevqacell{78.35} & \basevqacell{3.50} & \basevqacell{69.85} & \basevqacell{81.05} & \basevqacell{47.06} & \basevqacell{3.70} & \basevqacell{39.12} & \basevqacell{50.25} & \basevqacell{45.67} & \basevqacell{6.39} & \basevqacell{30.03} & \basevqacell{49.50} \\
 & Numbers$^{\dagger}$ & $79.73$ & $0.45$ & $79.21$ & $80.17$ & $48.22$ & $1.57$ & $45.25$ & $50.38$ & $50.89$ & $1.04$ & $49.50$ & $53.02$ \\
 & \basevqa{Numbers} & \basevqacell{80.04} & \basevqacell{0.65} & \basevqacell{79.21} & \basevqacell{81.03} & \basevqacell{48.22} & \basevqacell{1.57} & \basevqacell{45.25} & \basevqacell{50.38} & \basevqacell{48.03} & \basevqacell{1.23} & \basevqacell{46.14} & \basevqacell{50.17} \\
 & Roman$^{\dagger}$ & $79.35$ & $0.37$ & $78.97$ & $79.85$ & $47.09$ & $1.67$ & $45.12$ & $49.62$ & $50.27$ & $1.57$ & $48.15$ & $52.85$ \\
 & \basevqa{Roman} & \basevqacell{79.46} & \basevqacell{0.48} & \basevqacell{78.56} & \basevqacell{80.13} & \basevqacell{47.16} & \basevqacell{1.70} & \basevqacell{45.12} & \basevqacell{49.62} & \basevqacell{47.05} & \basevqacell{0.84} & \basevqacell{45.64} & \basevqacell{48.49} \\
\midrule
\multirow{8}{*}{\textbf{Phi-4}} & Uppercase$^{\dagger}$ & $82.76$ & $0.13$ & $82.66$ & $82.90$ & $63.19$ & $0.86$ & $62.12$ & $64.50$ & $70.92$ & $0.87$ & $69.63$ & $72.48$ \\
 & \basevqa{Uppercase} & \basevqacell{82.78} & \basevqacell{0.13} & \basevqacell{82.60} & \basevqacell{83.03} & \basevqacell{63.53} & \basevqacell{1.01} & \basevqacell{62.00} & \basevqacell{65.00} & \basevqacell{73.21} & \basevqacell{0.63} & \basevqacell{72.32} & \basevqacell{74.33} \\
 & Lowercase$^{\dagger}$ & $82.48$ & $0.33$ & $82.10$ & $82.69$ & $62.96$ & $1.15$ & $61.25$ & $64.50$ & $69.02$ & $3.92$ & $59.73$ & $72.48$ \\
 & \basevqa{Lowercase} & \basevqacell{82.60} & \basevqacell{0.17} & \basevqacell{82.34} & \basevqacell{82.86} & \basevqacell{63.73} & \basevqacell{1.16} & \basevqacell{62.00} & \basevqacell{65.50} & \basevqacell{72.58} & \basevqacell{0.88} & \basevqacell{71.48} & \basevqacell{74.16} \\
 & Numbers$^{\dagger}$ & $82.35$ & $0.39$ & $82.07$ & $82.62$ & $62.01$ & $0.98$ & $60.50$ & $63.25$ & $69.83$ & $1.34$ & $68.46$ & $72.32$ \\
 & \basevqa{Numbers} & \basevqacell{82.41} & \basevqacell{0.21} & \basevqacell{81.97} & \basevqacell{82.62} & \basevqacell{62.58} & \basevqacell{0.98} & \basevqacell{60.38} & \basevqacell{63.88} & \basevqacell{71.55} & \basevqacell{1.01} & \basevqacell{69.97} & \basevqacell{73.15} \\
 & Roman$^{\dagger}$ & $76.88$ & $1.81$ & $75.63$ & $78.95$ & $55.47$ & $3.08$ & $51.50$ & $61.50$ & $68.65$ & $1.72$ & $65.27$ & $70.64$ \\
 & \basevqa{Roman} & \basevqacell{74.24} & \basevqacell{3.48} & \basevqacell{68.30} & \basevqacell{79.10} & \basevqacell{54.67} & \basevqacell{3.45} & \basevqacell{49.62} & \basevqacell{61.38} & \basevqacell{62.00} & \basevqacell{4.36} & \basevqacell{57.05} & \basevqacell{70.97} \\
    \midrule

    \multirow{8}{*}{\textbf{Qwen2-VL}} & Uppercase$^{\dagger}$ & $82.97$ & $5.72$ & $68.21$ & $86.97$ & $63.51$ & $2.07$ & $58.38$ & $65.75$ & $71.59$ & $2.39$ & $67.95$ & $74.66$ \\
    & \basevqa{Uppercase} & \basevqacell{82.31} & \basevqacell{6.43} & \basevqacell{65.94} & \basevqacell{86.66} & \basevqacell{65.62} & \basevqacell{2.34} & \basevqacell{59.88} & \basevqacell{67.75} & \basevqacell{71.59} & \basevqacell{2.02} & \basevqacell{67.62} & \basevqacell{73.99} \\
    & Lowercase$^{\dagger}$ & $75.57$ & $19.46$ & $38.80$ & $87.01$ & $55.47$ & $17.19$ & $25.50$ & $66.00$ & $70.90$ & $2.19$ & $67.95$ & $73.99$ \\
    & \basevqa{Lowercase} & \basevqacell{74.58} & \basevqacell{20.72} & \basevqacell{35.76} & \basevqacell{86.83} & \basevqacell{57.00} & \basevqacell{18.53} & \basevqacell{25.62} & \basevqacell{68.12} & \basevqacell{63.91} & \basevqacell{15.66} & \basevqacell{35.23} & \basevqacell{73.83} \\
    & Numbers$^{\dagger}$ & $85.48$ & $0.54$ & $84.02$ & $86.03$ & $65.91$ & $0.82$ & $64.88$ & $67.50$ & $67.56$ & $1.83$ & $64.77$ & $69.97$ \\
    & \basevqa{Numbers} & \basevqacell{85.44} & \basevqacell{0.49} & \basevqacell{84.21} & \basevqacell{85.98} & \basevqacell{67.73} & \basevqacell{0.83} & \basevqacell{66.00} & \basevqacell{68.75} & \basevqacell{71.45} & \basevqacell{0.74} & \basevqacell{70.47} & \basevqacell{72.65} \\
    & Roman$^{\dagger}$ & $85.47$ & $0.38$ & $84.63$ & $86.05$ & $64.27$ & $0.85$ & $62.62$ & $65.50$ & $66.95$ & $2.55$ & $63.76$ & $70.81$ \\
    & \basevqa{Roman} & \basevqacell{85.53} & \basevqacell{0.29} & \basevqacell{84.98} & \basevqacell{85.94} & \basevqacell{67.09} & \basevqacell{0.78} & \basevqacell{65.88} & \basevqacell{68.38} & \basevqacell{71.38} & \basevqacell{1.48} & \basevqacell{68.96} & \basevqacell{73.99} \\

    \midrule

    \multirow{8}{*}{\textbf{Qwen2.5-VL}} & Uppercase$^{\dagger}$ & $87.66$ & $0.38$ & $87.07$ & $88.17$ & $70.72$ & $0.67$ & $69.50$ & $71.50$ & $73.90$ & $1.14$ & $72.15$ & $75.00$ \\
    & \basevqa{Uppercase} & \basevqacell{87.64} & \basevqacell{0.37} & \basevqacell{86.97} & \basevqacell{88.17} & \basevqacell{70.74} & \basevqacell{0.68} & \basevqacell{69.62} & \basevqacell{71.75} & \basevqacell{79.85} & \basevqacell{1.48} & \basevqacell{77.18} & \basevqacell{81.71} \\
    & Lowercase$^{\dagger}$ & $87.52$ & $0.36$ & $86.90$ & $87.93$ & $70.19$ & $0.42$ & $69.50$ & $70.88$ & $73.87$ & $1.38$ & $71.64$ & $75.17$ \\
    & \basevqa{Lowercase} & \basevqacell{87.51} & \basevqacell{0.32} & \basevqacell{86.99} & \basevqacell{87.99} & \basevqacell{70.04} & \basevqacell{0.39} & \basevqacell{69.25} & \basevqacell{70.50} & \basevqacell{80.10} & \basevqacell{1.24} & \basevqacell{78.02} & \basevqacell{81.38} \\
    & Numbers$^{\dagger}$ & $86.19$ & $0.38$ & $85.72$ & $86.90$ & $69.35$ & $0.46$ & $68.62$ & $70.00$ & $71.03$ & $0.88$ & $69.97$ & $72.32$ \\
    & \basevqa{Numbers} & \basevqacell{86.19} & \basevqacell{0.38} & \basevqacell{85.68} & \basevqacell{86.83} & \basevqacell{69.31} & \basevqacell{0.59} & \basevqacell{67.75} & \basevqacell{69.88} & \basevqacell{75.45} & \basevqacell{0.79} & \basevqacell{73.99} & \basevqacell{77.18} \\
    & Roman$^{\dagger}$ & $85.72$ & $0.31$ & $85.20$ & $86.07$ & $68.39$ & $0.87$ & $66.38$ & $69.38$ & $69.78$ & $1.10$ & $68.46$ & $71.98$ \\
    & \basevqa{Roman} & \basevqacell{85.73} & \basevqacell{0.34} & \basevqacell{85.13} & \basevqacell{86.16} & \basevqacell{68.36} & \basevqacell{0.81} & \basevqacell{66.50} & \basevqacell{69.12} & \basevqacell{74.34} & \basevqacell{0.84} & \basevqacell{72.65} & \basevqacell{75.67} \\
    
    \bottomrule
    \end{tabular}
    }
\end{table*}